\documentclass[10pt,twocolumn,letterpaper]{article}

\usepackage[preprint]{cvpr}
\usepackage{times}
\usepackage{epsfig}
\usepackage{graphicx}
\usepackage{amsmath}
\usepackage{amssymb}
\usepackage{booktabs}
\usepackage{tabularx}
\usepackage{color}
\usepackage{xcolor,array}

\definecolor{cvprblue}{rgb}{0.21,0.49,0.74}
\usepackage[pagebackref=true,breaklinks=true,letterpaper=true,colorlinks,citecolor=cvprblue,bookmarks=false]{hyperref}
\def\paperID{349} 
\def\confName{3DV\xspace}
\def\confYear{2024\xspace}

\definecolor{mark1}{rgb}{0.2,0.2,0.8}
\definecolor{textcolor}{rgb}{0.42, 0.35, 0.8}
\definecolor{textcolor1}{rgb}{0.45,.42,.75}

\newcommand{\loss}[1]{\mathcal{L}_\text{#1}}

\newcommand{\real}{\mathbb{R}}
\newcommand{\vect}[1]{#1}

\newcommand{\sdf}{f_{\theta_0}}
\newcommand{\colornet}{c_{\theta_1}}

\begin{document}


\title{Few-Shot Multi-Human Neural Rendering Using Geometry Constraints}

\author{Qian li$^{1}$, Victoria Fern\`andez Abrevaya$^{2}$, Franck Multon$^{1}$, Adnane Boukhayma$^{1}$\\
$^{1}$Inria, University Rennes, IRISA, CNRS, France\\
$^{2}$Max Planck Institute for Intelligent Systems, Germany
}


\maketitle

\begin{abstract}

We present a method for recovering the shape and radiance of a scene consisting of multiple people given solely a few images. 
Multi-human scenes are complex due to additional occlusion and clutter. For single-human settings, existing approaches using implicit neural representations have achieved impressive results that deliver accurate geometry and appearance. 
However, it remains challenging to extend these methods for estimating multiple humans from sparse views. 
We propose a neural implicit reconstruction method that addresses the inherent challenges of this task through the following contributions: First, we propose to use geometry constraints by exploiting pre-computed meshes using a human body model (SMPL). Specifically, we regularize the signed distances using the SMPL mesh and leverage bounding boxes for improved rendering. Second, we propose a ray regularization scheme to minimize rendering inconsistencies, and a saturation regularization for robust optimization in variable illumination.  Extensive experiments on both real and synthetic datasets demonstrate the benefits of our approach and show state-of-the-art performance against existing neural reconstruction methods. 

\end{abstract}

\section{Introduction}
\label{sec:intro}

Human reconstruction from single images \cite{choutas2020monocular, kanazawa2018end,liu2022recent}, multiple images \cite{guo2019relightables, collet2015high}, RGB videos \cite{alldieck2018detailed,Kocabas20} or RGB-D data \cite{yu2017bodyfusion,yu2018doublefusion} has received a lot of attention, much less explored is the task of \emph{multiple} human scenario, which is essential for scene understanding, behavior modeling, collaborative augmented reality, and sports analysis.  
The multi-human setting introduces additional challenges, as there is now a higher level of occlusion and clutter 
which hinders matching and reconstruction. 
Although in principle one could approach this by first detecting and then independently processing each person, 
simultaneous reconstruction of multiple humans can help to globally reason about occlusion at the level of the scene~\cite{jiang2020coherent, sun2022putting}, 
and can potentially recover coherent 3D spatial relations among the people.

Several recent works have attempted to recover multiple humans from a single view \cite{choi2022learning, sun2022putting, sun2021monocular, zanfir2018deep, zanfir2018monocular, fieraru2020three, jiang2020coherent, zhang2021body, ugrinovic2021body,mustafa2021multi}. However, the majority of these are based on regressing the parameters of a human body model --typically SMPL \cite{loper2015smpl}--
which provides coarse reconstructions that 
lack hair, clothing, and geometric details. 
Multi-view settings can help resolve some of the occlusions as well as depth ambiguities, but require a dense array of RGB cameras to achieve a detailed reconstruction \cite{collet2015high, joo2015panoptic,vlasic2009dynamic}.
A more convenient capture system 
is the \emph{sparse} multi-view setting, where only a handful of cameras is required.
However, due to the decreased number of views and increased level of occlusion, 
existing methods require segmentation masks and a pre-scanned template mesh \cite{liu2011markerless, wu2013set}, rely on a coarse body model \cite{zhang2021lightweight, huang2021dynamic}, or require temporal information \cite{zheng2021deepmulticap, huang2021dynamic}.

A parallel line of work simultaneously tackles the novel-view-synthesis and geometry-reconstruction problems by combining neural coordinate-based representations, \eg implicit signed distance functions (SDFs) \cite{park2019deepsdf}, with differentiable rendering \cite{yariv2021volume,wang2021neus,yariv2020multiview,mildenhall2020nerf}. 
This approach has the advantage of producing, along with geometry, renderings from novel viewpoints that can capture complex surface/light interactions, increasing the scope of applications. 
NeRF~\cite{mildenhall2020nerf}, for example, uses volumetric rendering to produce impressive images under novel views, albeit at the cost of sub-optimal geometries due to the unconstrained volumetric representation. 
SDF-based methods \cite{yariv2021volume,wang2021neus,yariv2020multiview}, while delivering images of slightly lower quality, have been shown to produce 3D surfaces that are competitive with classical approaches. 
For humans, this has been leveraged to obtain geometry and appearance from monocular video \cite{jiang2022selfrecon,chen2021animatable}, RGB-D video \cite{dong2022pina}, and sparse multi-view video \cite{wang2022arah, liu2021neural, peng2021neural, zheng2021deepmulticap, kwon2021neural, peng2021animatable, xu2021h, weng2022humannerf}. 
However, none of these works, with the exception of \cite{zheng2021deepmulticap,zhang2021editable}, were designed to handle the increased geometric complexity and occlusion of the multi-human case. 
Current works \cite{zheng2021deepmulticap,zhang2021editable} address the multi-human setting, but both require a set of videos, which effectively becomes a dense array of views as long as deformations are modeled correctly.

In this paper, we address the problem of multiple 3D human surfaces and volume rendering from sparse static multi-view images. Our key insight is that human-specific geometric constraints can be leveraged to tackle the challenging sparse-view setting.

Specifically, we first obtain a SMPL body model from the input data and use it to initialize the implicit SDF network, where we define the surface of a multi-human scene as the zero-level set of the SDF. 
Then, the geometry network is optimized with multi-view images by leveraging surface and volume rendering ~\cite{wang2021neus} along with uncertainty estimation methods \cite{deng2022depth,roessle2022dense}, where the SMPL meshes are treated as noisy estimations.  
To achieve higher rendering quality from sparse training views, we additionally propose a patch-based regularization loss that guarantees consistency across different rays and a saturation regularization that ensures consistency for variable image illuminations within the same scene.

We evaluate our method quantitatively and qualitatively on real multi-human (CMU Panoptic~\cite{Simon_2017_CVPR,Joo_2017_TPAMI}) and synthetic (MultiHuman~\cite{zheng2021deepmulticap}) datasets. We demonstrate results on 5,10,15 and 20 training views, where we achieve state-of-the-art performance in terms of surface reconstruction and novel view quality.

\section{Related Work}
\label{sec:relat}


\paragraph{Single-Human Reconstruction.} 
There is a vast amount of work on reconstructing 3D humans from single images \cite{bogo2016keep, choutas2020monocular, kanazawa2018end,muller2021self,liu2022recent}, monocular video \cite{yuan2022glamr, Kocabas20, alldieck2018video}, RGB-D data \cite{yu2017bodyfusion,yu2018doublefusion,burov2021dynamic} and multi-view data \cite{starck2007surface,collet2015high,guo2019relightables,huang2018deep}. We concentrate here on the multi-view setting. 
High-end multi-view capture systems can achieve reconstructions of outstanding quality \cite{leroy2018shape, dou2016fusion4d, guo2019relightables, collet2015high,vlasic2009dynamic,Joo_2017_TPAMI}, but require a complex studio setup that is expensive to build and not easily accessible.
To alleviate this, numerous works have been proposed that use instead a sparse set of RGB cameras (\eg between 2 and 15), 
where the lack of views and presence of wide baselines is compensated by tracking a pre-scanned template \cite{gall2009motion, vlasic2008articulated, carranza2003free, de2008performance, wu2012full},
using a parametric body model  \cite{huang2017towards, balan2007detailed}, 
or more recently, by the use of deep learning \cite{huang2018deep,liang2019shape, wang2022arah, liu2021neural, peng2021neural,  kwon2021neural, peng2021animatable, xu2021h, weng2022humannerf}. 
\paragraph{Multi-Human Reconstruction.} 
In contrast, there has been a limited number of works that address the problem of \emph{multiple} human reconstruction. 
This is a difficult task since the presence of several people increases the geometric complexity of the scene, introduces occlusions,  and amplifies ambiguities such that commonly used features like color, edges, or key points cannot be correctly assigned. 

For single images and video, the problem has been mainly tackled by regressing the parameters of the SMPL \cite{loper2015smpl} body model~\cite{zhang2020perceiving, choi2022learning, sun2022putting, sun2021monocular, zanfir2018deep, dong2021shape, zanfir2018monocular, fieraru2020three, jiang2020coherent, zhang2021body, ugrinovic2021body, guler2019holopose}. Although this can work robustly with as little as one view, the reconstructions are very coarse and cannot explain hair, clothing, and fine geometric details. 
The only exception is the work of Mustafa \etal~\cite{mustafa2021multi}, which performs model-free reconstruction of multiple humans by combining an explicit voxel-based representation with an implicit function refinement. 
However, 
the method requires training on a large synthetic dataset of multiple people which hinders generalization. Our work, on the other hand, performs 3D reconstructions, produces renderings of novel views, and can generalize to arbitrary multi-human scenes.

Multi-view capture setups can help resolve depth ambiguities and some of the occlusions. 
Classic methods for estimating multiple humans rely heavily on segmentation masks and template mesh tracking \cite{liu2011markerless, liu2013markerless, wu2013set}. We avoid the use of segmentation masks by adopting volumetric rendering for implicit surfaces \cite{wang2021neus}. 
More recently, deep learning-based approaches were proposed, but they either require temporal information~\cite{zheng2021deepmulticap,huang2021dynamic,zhang2021lightweight,shuai2022novel}, pre-training on a large dataset~\cite{zheng2021deepmulticap} which cannot work on general scenes, or a coarse body model \cite{zhang2021lightweight, huang2021dynamic, shuai2022novel} which lacks geometric detail. 
Here, we focus on the multi-human setting on static scenes and propose a method that recovers accurate reconstructions and at the same time produces renderings of novel viewpoints. 
%
%
%
%
%
%
%
\paragraph{Neural Surface Reconstruction and Novel-View Synthesis.} 
For generating free-viewpoint video, image-based rendering has been considered as an alternative or complement to 3D reconstruction~\cite{carranza2003free,wu2020multi,liu2021neural,kwon2021neural, liu2021neural, peng2021animatable, weng2022humannerf,xu2021h}. 
When geometry proxies are available, neural rendering \cite{aliev2020neural,thies2019deferred,jena2022neural} can produce competitive novel view synthesis. Recently, NeRF\cite{mildenhall2020nerf} demonstrated impressive rendering results by representing a 3D scene as a neural radiance field, trained only with calibrated multi-view images through the use of volume rendering. However, due to the unconstrained volumetric representation and self-supervised training on RGB values, reconstructed geometries tend to be too noisy to be useful for 3D applications. 
To recover more accurate 3D geometry along with appearance, DVR~\cite{niemeyer2020differentiable}, IDR~\cite{yariv2020multiview}, and NLR~\cite{kellnhofer2021neural} propose to learn an implicit representation directly from multi-view images but require accurate object masks to work. To avoid the need for segmentation masks, recent works propose to combine implicit representations with volume rendering \cite{oechsle2021unisurf, yariv2021volume, wang2021neus}. 
These methods show remarkable reconstruction results but struggle when the number of input views is low. 
Implicit neural representations from sparse input can be obtained by using pre-trained pixel-aligned features or 3D feature volumes for input images ~\cite{saito2019pifu, saito2020pifuhd, alldieck2022photorealistic, he2020geo, huang2020arch, he2021arch, yu2021pixelnerf, li2023learning, jena2024geotransfer} or point clouds ~\citep{boulch2022poco,williams2022neural,huang2023neural,peng2020convolutional,chibane2020implicit,lionar2021dynamic, ouasfi2023mixing, peng2021shape, ouasfi2022few, ouasfi2024robustifying}, 
but this requires ground-truth geometry and is limited by the training data, struggling to generalize to new scenes. Sparse variants that do not require generalizable features were proposed in the image input \eg \cite{niemeyer2022regnerf, kim2022infonerf, long2022sparseneus, younes2025sparsecraft, li2023regularizing} and point cloud input case \eg \cite{NeuralTPS,sparseocc,nap,ouasfi2024robustneuralreconstructionsparse,williams2021neural}. 
InfoNeRF~\cite{kim2022infonerf} regularizes sparse views by adding an entropy constraint on the density of the rays,  RegNeRF~\cite{niemeyer2022regnerf} uses a patch-based regularizer over generated depth maps, and SparseNeuS~\cite{long2022sparseneus} uses a multi-scale approach along with learned features that are fine-tuned on each scene. 
Our approach builds on NeuS \cite{wang2021neus}, and tackles the sparse view challenge by adding human-specific geometric priors and novel regularizations.

 \begin{figure*}[ht]
\vspace{-6mm}
\centering
\includegraphics[width=.65\linewidth]{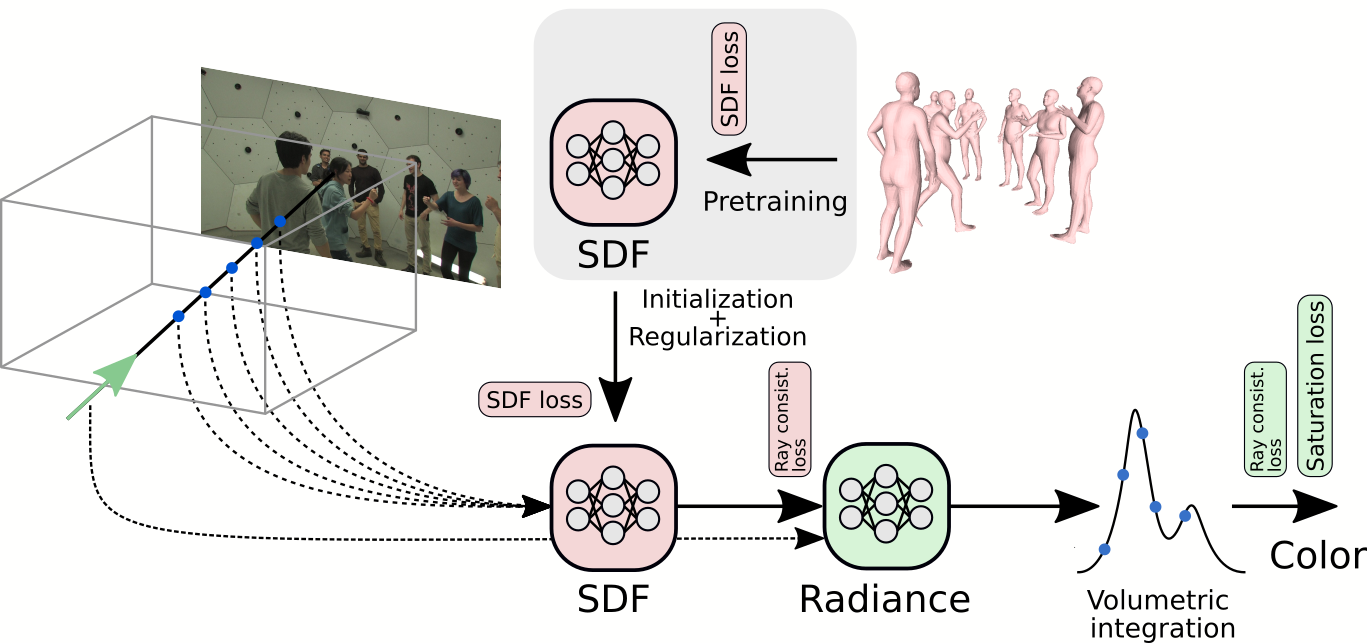}
  \caption{\textbf{Overview.} We address the multi-human implicit shape and appearance learning problem by initializing the geometry using SMPL  (Sec.~\ref{sec:method_geometric_init}), along with uncertainty-based SDF supervision and novel photometric regularizations designed to compensate for the lack of views (Sec.~\ref{sec:method_regularizations}). We also model the foreground (Union of SMLP bounding boxes) and remainder of the scene seperatelty (Sec.~\ref{sec:method_bboxes}).}
  
  \label{fig:pipe}
\vspace{-5mm}
\end{figure*}

\section{Method}
\label{sec:method}
Given a sparse set of views $\{I_i\}_{i=1}^N$ of a multi-human scene with camera intrinsics and extrinsics ${\{ K_i, [R|t]_i \}}$, our goal is to reconstruct geometry and synthesize the appearance of multiple humans from arbitrary viewpoints. The pipeline is illustrated in Fig.~\ref{fig:pipe}. Our approach builds on NeuS  \cite{wang2021neus}, which combines an implicit signed distance representation for geometry with volumetric rendering. 
In order to solve the challenging case of multiple humans occluding each other, we hypothesize that a naive RGB reconstruction loss is insufficient and propose to use a strong geometric prior before training with multi-view images. 
Towards this, we first train the implicit SDF network independently by leveraging off-the-shelf SMPL estimations (Sec.~\ref{sec:method_geometric_init}). 
To handle details and represent appearance, the geometry network is then fine-tuned considering foreground and background objects. Moreover, we propose the use of hybrid bounding box rendering to handle the multi-human setting (Sec.~\ref{sec:method_bboxes}). 
Additionally, we define an explicit SDF constraint based on the uncertainty of the SMPL estimations, together with a ray consistency loss, and a saturation loss to improve image rendering quality for sparse views (Sec.~\ref{sec:method_regularizations}). 

\subsection{Scene Representation and Rendering}
%
We define a multi-human surface $\mathcal{S}$ as the zero-level set of a signed distance function (SDF) 
$\sdf : \real^3 \to \real$, encoded by a Multilayer Perceptron (MLP) $\sdf$ with parameters $\theta_0$:
\begin{equation}
    \mathcal{S} = \left\{p\in \real^3|\sdf(p)=0 \right\}.
  \label{eq:sdf}
\end{equation}
Following NeuS~\cite{wang2021neus}, we train the geometry network $\sdf$ along with a color network 
$\colornet$, with parameters $\theta_1$, mapping a point $\vect{p}$ 
to color values (more details in Sec.~\ref{sec:method_bboxes}). 
Combining the SDF representation with volume rendering, we approximate the color along a ray $r$ by:
\begin{gather} 
    C(r) = \sum_{i=1}^{N} w(\vect{p}_i) \colornet(\vect{p}_i), \\
    w(\vect{p}_i) = T(\vect{p}_i) \alpha(\vect{p}_i), \\
    T(\vect{p}_i) = \prod_{j}^{i-1}(1-\alpha(\vect{p}_j)),
\label{eq:T}
\end{gather}
where $\vect{p}_i = \vect{o} + t_i \vect{v}$ 
is a sampled point along the ray $r$ starting at camera center $\vect{o}$ with direction $\vect{v}$; 
$\colornet(\vect{p}_i)$ is the predicted color at $\vect{p}_i$, $w(\vect{p_i})$ is the weight function,  $T(\vect{p}_i)$ is the accumulated transmittance, and $\alpha(\vect{p_i})$ is the opacity value. 
Following NeuS, $\alpha(\vect{p_i})$ is defined as a function of the signed distance representation:
\begin{gather} 
    \alpha(\vect{p}_i) = max\left(\frac{\Phi(\sdf(\vect{p}_i))-\Phi(\sdf(\vect{p}_{i+1}))} {\Phi(\sdf(\vect{p}_i))},0 \right) 
\label{eq:alpha_p}
\end{gather}
where $\sdf(\vect{p}_i)$ is the signed distance of $\vect{p}_i$, $\Phi(\sdf(\vect{x}))=(1+e^{-s \vect{x}})^{-1}$ is the cumulative distribution function (CDF) of the logistic distribution, and $s$ is a learnable parameter (see \cite{wang2021neus} for more details). 
%
\subsection{Geometric Prior}
\label{sec:method_geometric_init}
Typically, the SDF function $\sdf$ and the color function $\colornet$ are simultaneously optimized by minimizing the difference between the rendered and ground-truth RGB values~\cite{yariv2020multiview,mildenhall2020nerf,wang2021neus}. While this allows to train without the need for geometric supervision, it has been noted that a photometric error alone is insufficient for the challenging sparse-view setting \cite{deng2022depth,roessle2022dense}, since there are not enough images to compensate for the inherent ambiguity in establishing correspondences between views. For the \emph{multi-human} setting this becomes more problematic, as correspondences are even more ambiguous due to clutter. 

To address this, we propose to regularize using geometric information by first independently training $\sdf$ using off-the-shelf SMPL fittings, which can be robustly computed from the input data. 
We train this network in a supervised manner by sampling points with their distance values as in \cite{park2019deepsdf}. 
Given that SMPL can only coarsely represent the real surface, we treat this geometry as a ``noisy'' estimate that will be later improved upon using the multi-view images. Preparing for this, and inspired by \cite{deng2022depth,roessle2022dense}, we model the SMPL ``noise'' as a Gaussian distribution $\mathcal{N}(0,{s_{noise}(\vect{p}_j})^2)$ with standard deviation $s_{noise}(\vect{p}_j)$, and train $\sdf$ to output an estimate of the uncertainty $s_{noise}(\vect{p}_j)$ along with the distance value; that is, $\sdf(\vect{p}_j) = (d_j,s_{noise})$. The geometry network $\sdf$ is then optimized by minimizing the negative log-likelihood of a Gaussian:
%
%
\begin{equation}
    \mathcal{L}_s = \frac{1}{n}\sum_{j=1}^{n} \left( log(s_{noise}(\vect{p}_j)^2) + \frac{{(d_j-d'_j)}^2}{ s_{noise}(\vect{p}_j)^2} \right), 
\end{equation}
where $n$ is the number of sampled points, $d_j$ is the predicted SDF value for point $\vect{p}_j$, and $d_j'$ is the signed distance sampled directly from the SMPL meshes. 

\subsection{Hybrid Rendering with Geometry Constraints}
\label{sec:method_bboxes}

To work with unbounded scenes, NeRF++ \cite{zhang2020NeRF++} proposed to separately model the foreground and background geometries using an inverted sphere parameterization, where the foreground is parameterized within an inner unit sphere, and the rest is represented by an inverted sphere covering the complement of the inner volume. We follow this and train separate models for foreground and background. Specifically, we use a simple NeRF~\cite{mildenhall2020nerf} architecture for the background and train the foreground model using $\sdf$ and the color network $\colornet$, where the output color $C(\vect{p}_i)$ is predicted as:
%
%
\begin{equation} 
C(\vect{p}_i) = \colornet ( \gamma(\vect{p}_i), \gamma(\vect{v}_i), f_0,f_1).\\
\label{eq:colornet}
\end{equation}
Here, $\gamma(\vect{p}_i)$ and $\gamma(\vect{v}_i)$ are the positional encodings \cite{tancik2020fourier,mildenhall2020nerf} of the sampled point $\vect{p}_i$ and its ray direction $\vect{v}_i$, and $f_0$ includes the gradients of predicted SDF and predicted feature from the geometry network $\sdf$ \cite{yariv2020multiview}. Additionally, to inject geometric prior knowledge into the appearance network we condition $\colornet$ on the rasterized depth feature from the corresponding SMPL mesh. 

For reconstructing multiple humans, one difficulty in modeling the foreground as in NeRF++ is that the bounding sphere will contain a large empty space, making it costly to search for the surface during hierarchical sampling and adding non-relevant points to the training. To resolve this, we propose to use instead multiple 3D bounding boxes as the foreground volume. 
Specifically, we define a bounding box $B^j$ for the $j-$th human using the SMPL fittings, with minimum and maximum coordinates $[B_{min}^j - \delta, B_{max}^j + \delta]$, where $B_{min}^j$ and $B_{max}^j$ are the minimum and maximum coordinates of SMPL along the $x,y,z$ axes respectively, and $\delta$ is a spatial margin (here we set to $0.1$). The foreground volume is then defined as $B = \cup_{j=1..M} B^j $, and we define $b(\vect{p}_i)$ as
%
%
\begin{equation} 
    b(\vect{p}_i) =\left\{\begin{matrix}
    1,p_{i} \in B,\\
    0,p_{i} \notin B\\
 \end{matrix}\right.
\label{eq:in_box}
\end{equation}

For points that fall inside the foreground, $\vect{p} \in B$, 
we calculate the opacity value $\alpha^{FG}(\vect{p}_{i})$ using the predictions of $\sdf(\vect{p}_i)$ according to Eq. \ref{eq:alpha_p}, and the color $C(\vect{p_i})^{FG}$ using $\colornet$. 
The points that fall outside the bounding box are modeled as background using a NeRF model, where the opacity is calculated as $\alpha^{BG}(p_{i}) = 1-e^{\sigma(p_i)\delta(p_i)}$, with $\delta$ and $\sigma$ defined as in~\cite{mildenhall2020nerf}, and the color $C^{BG}$ is predicted using $\alpha^{BG}$. 
Given a point $\vect{p}_i$, its color and opacity values are updated as follows:
\begin{gather} 
C(\vect{p}_{i}) = b(\vect{p}_i) C^{FG}(\vect{p}_{i})+(1-b(\vect{p}_i)) C^{BG}(\vect{p}_{i})\\
\alpha(\vect{p}_{i})=b(\vect{p}_i) \alpha^{FG}(\vect{p}_{i})+(1-b(\vect{p}_i)) \alpha^{BG}(\vect{p}_{i}) 
\label{eq:rendering}
\end{gather}

Finally, following \cite{azinovic2022neural}, given a ray $r$ with $n$ sampled points $\{\vect{p}_i = \vect{o} + t_i \vect{v} \}_{i = 1}^{n}$, the color is approximated as:
\begin{gather} 
C(r) =\frac {\sum_{i=1}^{N}W(\vect{p}_{i}) C(\vect{p}_{i})}{ \sum_{i=1}^{N} W(\vect{p}_{i})},
\label{eq:raycolor}
\end{gather}
where $W(\vect{p}_{i}) = T(\vect{p}_{i}) \alpha(\vect{p}_{i})$, $T(\vect{p}_{i}) = \prod_{j}^{i-1}(1-\alpha(\vect{p}_{j}))$. This function allocates higher weights to points near the surface and lower weights to points away from the surface, and is used to improve the rendering quality. 




\subsection{Optimization}
\label{sec:method_regularizations}

Given a set of multi-view images, and a pre-trained SDF network $\sdf'$ (Sec.~\ref{sec:method_geometric_init}), we minimize the following objective:
\begin{align}
    \mathcal{L} = \loss{r}  + \lambda_{eik} \loss{eik} + \lambda_{sdf} \loss{sdf} + \lambda_{r} \loss{r} +\lambda_{s} \loss{s}, 
\end{align}
where $\loss{r}$ is a L1 reconstruction loss between the rendered image $I_r$ and the ground-truth $I_r^{'}$ and $\loss{eik}$ is the Eikonal loss~\cite{gropp2020implicit}. 

%
Additionally, we propose to use an uncertainty-based SDF loss $\loss{sdf}$, a novel ray consistency loss $\loss{r}$ and saturation loss $\loss{s}$ which are explained in the following.

\paragraph{SDF Loss.} 
As detailed in Sec.~\ref{sec:method_geometric_init}, we treat the SMPL mesh as a noisy estimate of the real surface.
When the sampled points are not within the foreground box $B$, or the absolute sdf value predicted by the geometry network $\sdf$ 
is greater than a pre-defined threshold $\xi_0$, or the standard deviation $s_j = s_{noise}(\vect{p})_j$ is bigger than the threshold $\xi_1$, we use the following loss:
\begin{equation} 
    \loss{sdf} =\left\{\begin{matrix}
    \frac{1}{n}\sum_{j=1}^{n}(log(s_{j}^2)+\frac{(d_j^{'}-d_{j})^2}{s_j^2}),\\  s.t.  (\vect{p}_{i} \notin B, |d_j|>\xi_0  \quad or  \quad s_j>\xi_1)\\
        0,otherwise\\
 \end{matrix}\right.
\label{eq:sdfloss}
\end{equation}
where $d_j$ and $d_j^{'}$ are the SDF predictions from the final $\sdf$ and initial network $\tilde{\sdf}$, and $\xi_0$, $\xi_1$ are set to 0.2 and 0.5, respectively. This function encourages the network to maintain geometry consistency during learning while allowing some freedom to learn the details encoded in the images.

\paragraph{Ray Consistency Loss.}
We introduce the following ray consistency loss $\loss{r}$ to ensure photometric consistency across all images under sparse views: 
\begin{align}
\loss{r} = ||C(r_i) - C(r^{*})||_1 + D_{KL}(P(r_i)||P(r^*))
\label{eq:kl}
\end{align}
where $C(r_i)$ is the ground truth color of a randomly sampled ray $r_i$ on a small patch and $C(r_{p}^{*})$ denotes the rendered color of an interpolated ray on a small patch. Inspired by \cite{kim2022infonerf}, we introduce a KL-divergence regularization for the ray density, where $P(r_i)= \frac{\alpha_i}{\sum_{i=1}^{N}a_{i}} $. 
%
The goal of this loss is to ensure consistency and smoothness of unseen rays by constraining the interpolated rays on a small patch to have a similar distribution, both for color and density.
%
\paragraph {Saturation Loss.}
Finally, we observe that real-world images might contain variable illumination or transient occluders among different views (this is the case for example in the CMU Panoptic dataset~\cite{Simon_2017_CVPR,Joo_2017_TPAMI}),  which can degrade the rendering quality due to inconsistency across views. Instead of learning complex transient embeddings as in \cite{martin2021nerf}, we propose to simply convert the RGB image into the HSV space, and calculate the L1 reconstruction loss of the saturation value between the rendered image and the ground truth: $\loss{s} = ||I_s - I_s^{gt}||_1$.

\begin{figure*}[h!]
\flushleft 
\vspace{-3mm}
\def\tabularxcolumn#1{m{#1}}
\setlength{\tabcolsep}{0pt}
\renewcommand{\arraystretch}{0} 
\scalebox{1}{
\begin{tabular}{c cccc ccc}
\rotatebox{90}{\quad 5} 
\includegraphics[width=2.5cm]{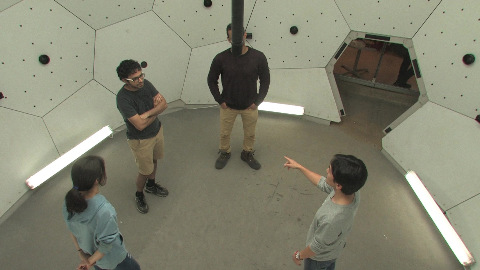} &
\includegraphics[width=2.5cm]{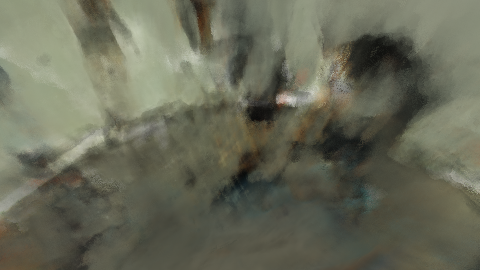} &
\includegraphics[width=2.5cm]{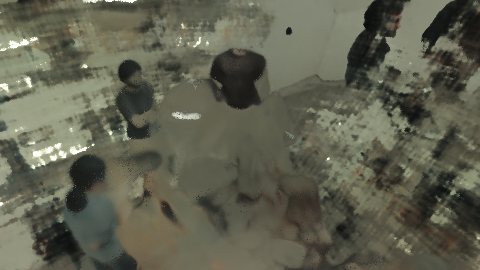} &
\includegraphics[width=2.5cm]{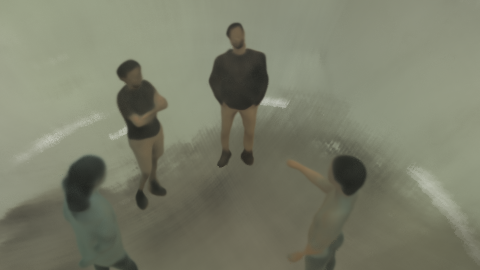} &
\includegraphics[width=2.5cm]{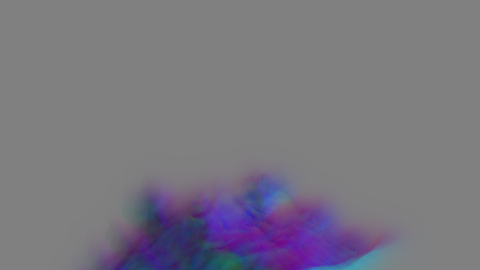} &
\includegraphics[width=2.5cm]{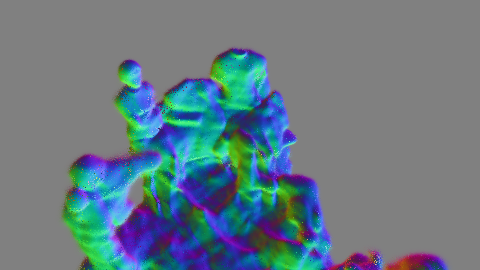} &
\includegraphics[width=2.5cm]{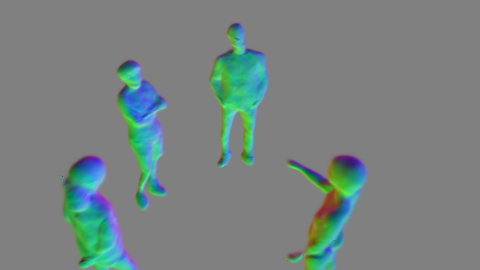} \\
\rotatebox{90}{\quad 5} 
\includegraphics[width=2.5cm]{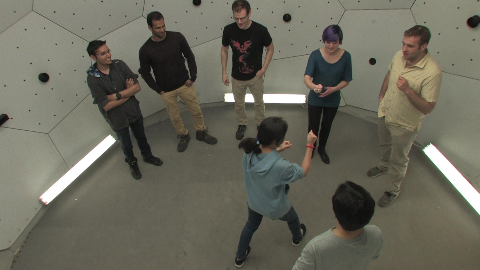} &
\includegraphics[width=2.5cm]{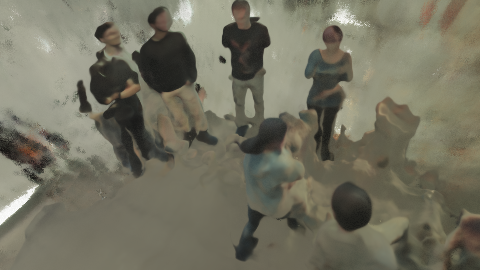} &
\includegraphics[width=2.5cm]{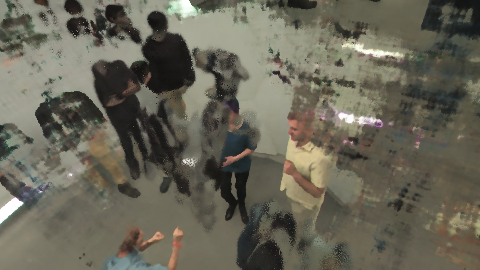} &
\includegraphics[width=2.5cm]{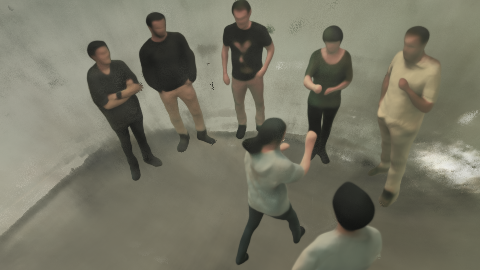} &
\includegraphics[width=2.5cm]{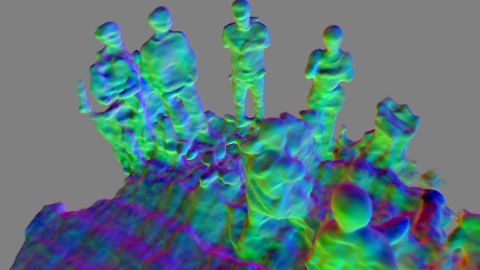} &
\includegraphics[width=2.5cm]{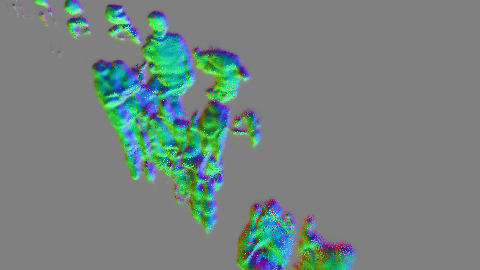} &
\includegraphics[width=2.5cm]{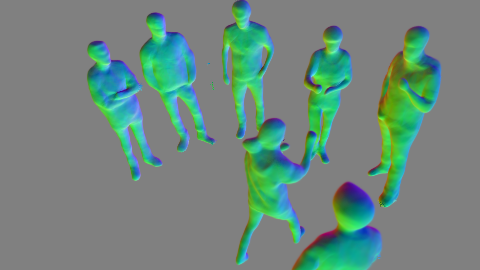} \\
\rotatebox{90}{\quad 10} 
\includegraphics[width=2.5cm]{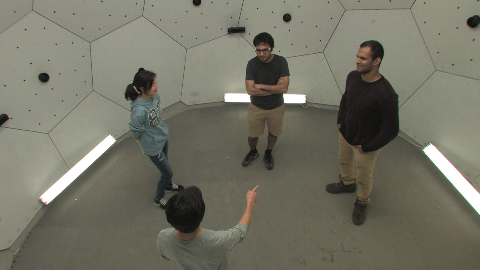} &
\includegraphics[width=2.5cm]{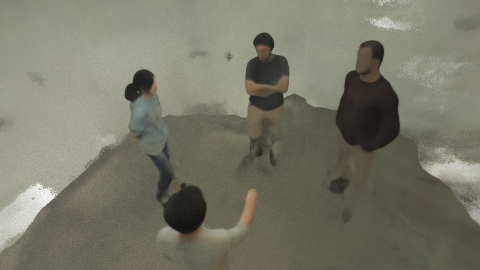} &
\includegraphics[width=2.5cm]{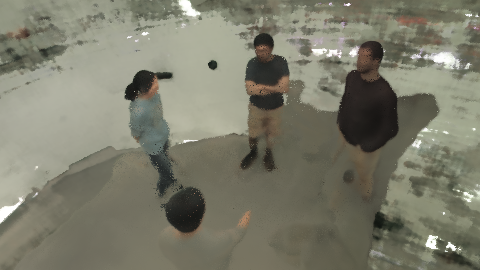} &
\includegraphics[width=2.5cm]{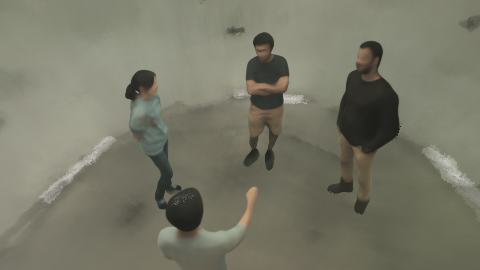} &
\includegraphics[width=2.5cm]{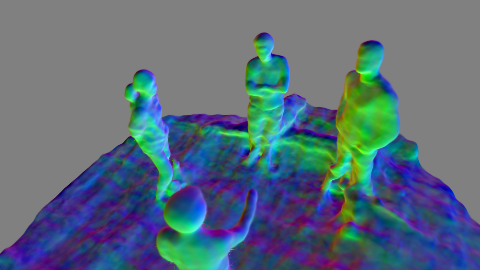} &
\includegraphics[width=2.5cm]{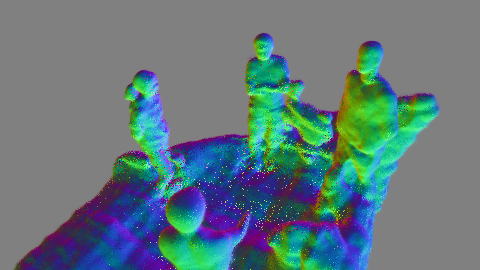} &
\includegraphics[width=2.5cm]{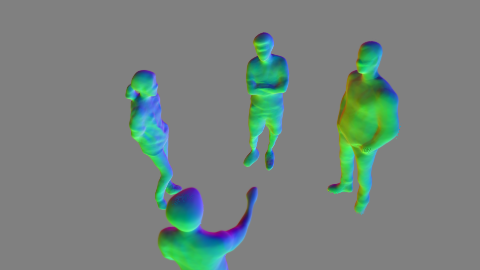} \\
\rotatebox{90}{\quad 10} 
\includegraphics[width=2.5cm]{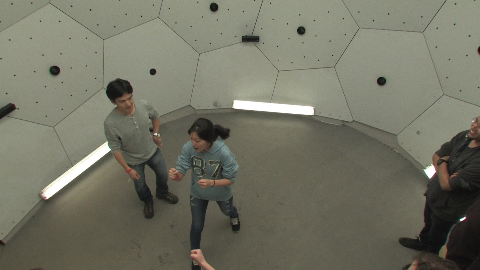} &
\includegraphics[width=2.5cm]{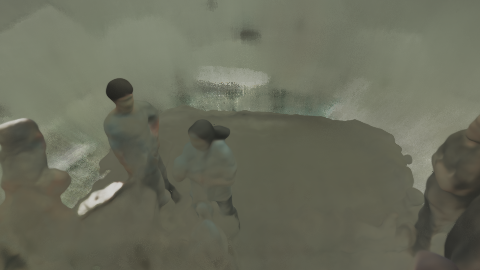} &
\includegraphics[width=2.5cm]{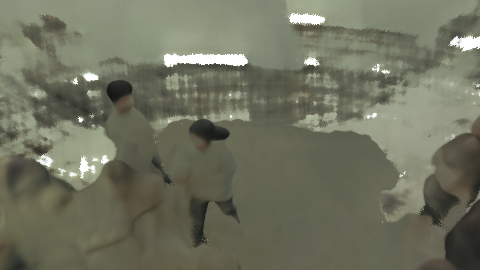} &
\includegraphics[width=2.5cm]{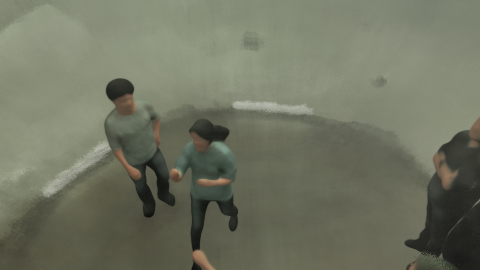} &
\includegraphics[width=2.5cm]{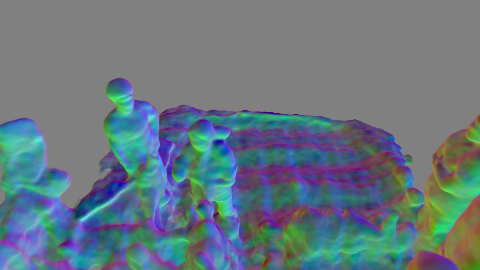} &
\includegraphics[width=2.5cm]{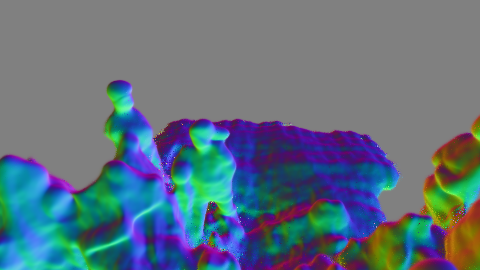} &
\includegraphics[width=2.5cm]{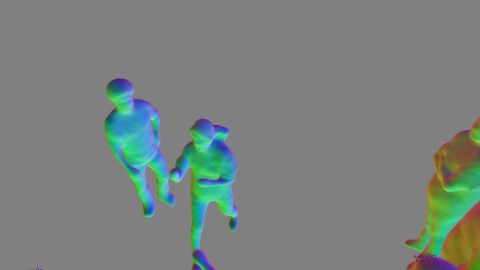} \\
\rotatebox{90}{\quad 15}
\includegraphics[width=2.5cm]{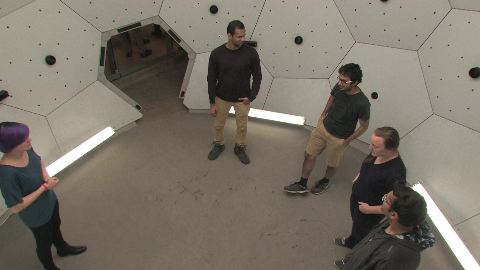} &
\includegraphics[width=2.5cm]{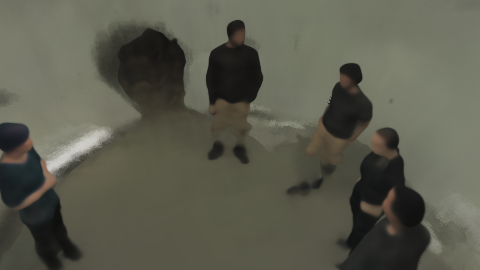} &
\includegraphics[width=2.5cm]{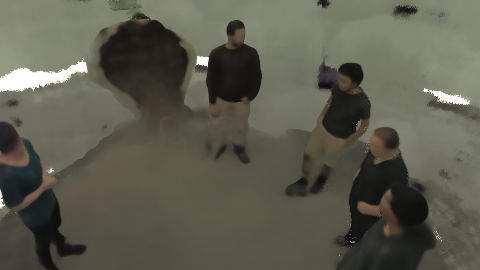} &
\includegraphics[width=2.5cm]{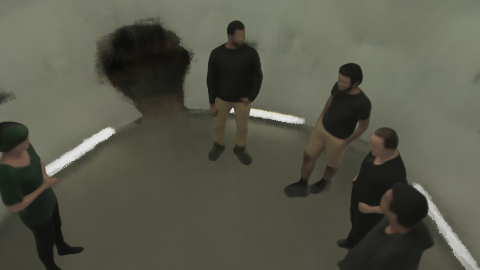} &
\includegraphics[width=2.5cm]{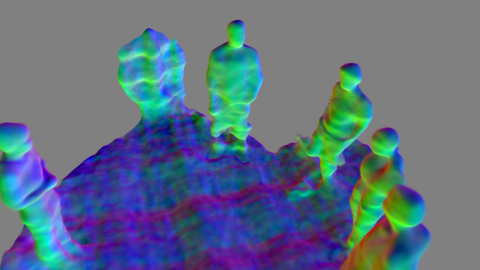} &
\includegraphics[width=2.5cm]{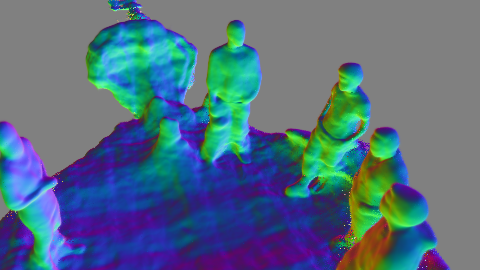} &
\includegraphics[width=2.5cm]{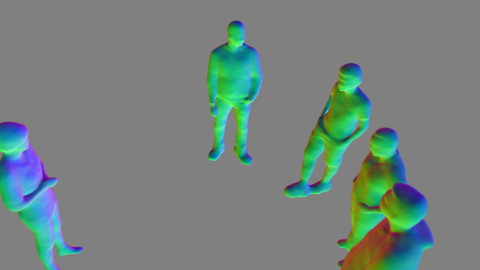} \\
\rotatebox{90}{\quad 15}
\includegraphics[width=2.5cm]{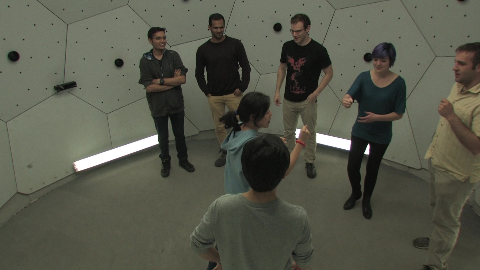} &
\includegraphics[width=2.5cm]{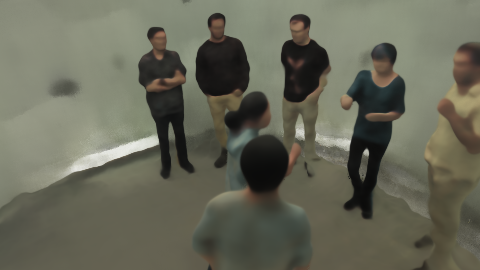} &
\includegraphics[width=2.5cm]{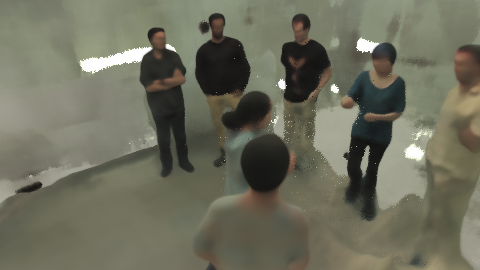} &
\includegraphics[width=2.5cm]{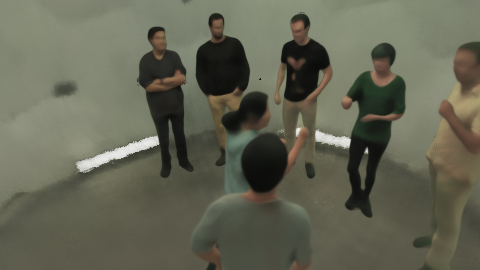} &
\includegraphics[width=2.5cm]{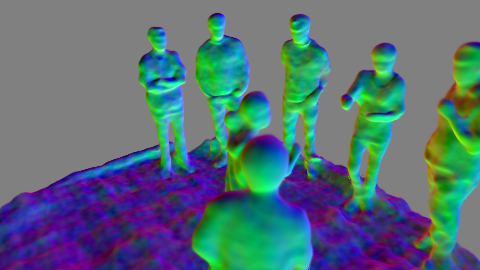} &
\includegraphics[width=2.5cm]{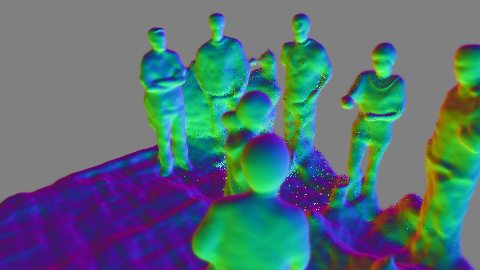} &
\includegraphics[width=2.5cm]{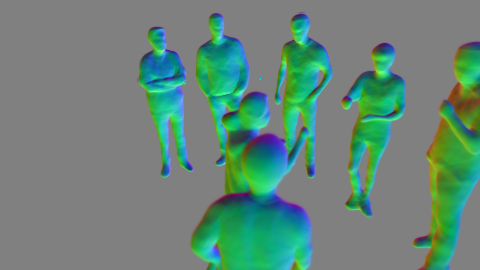} \\
\rotatebox{90}{\quad 20}
\includegraphics[width=2.5cm]{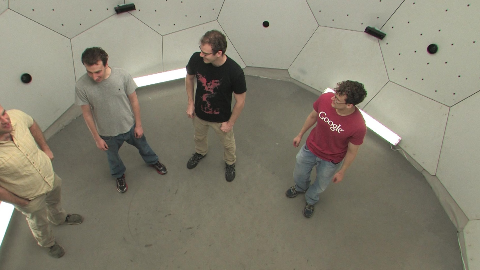} &
\includegraphics[width=2.5cm]{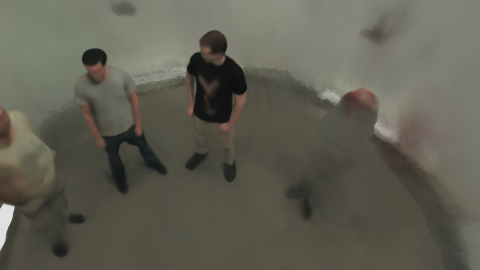} &
\includegraphics[width=2.5cm]{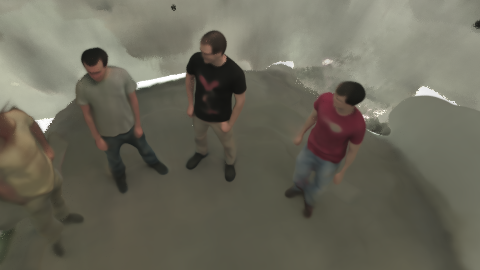} &
\includegraphics[width=2.5cm]{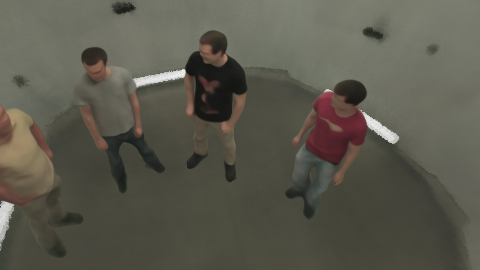} &
\includegraphics[width=2.5cm]{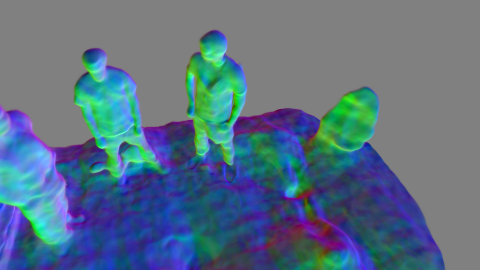} &
\includegraphics[width=2.5cm]{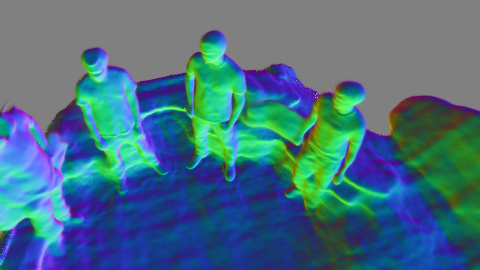} &
\includegraphics[width=2.5cm]{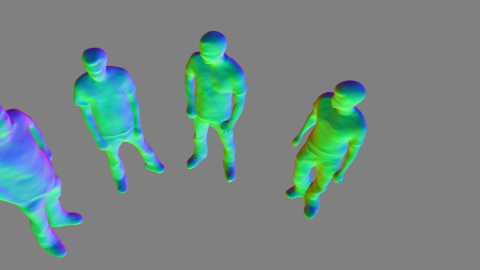} \\
\rotatebox{90}{\quad 20}
\includegraphics[width=2.5cm]{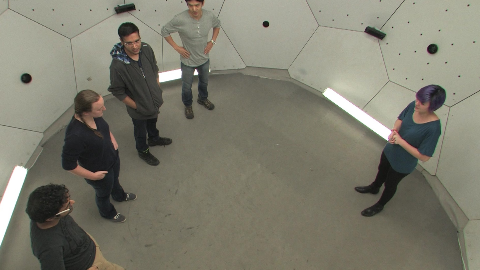} &
\includegraphics[width=2.5cm]{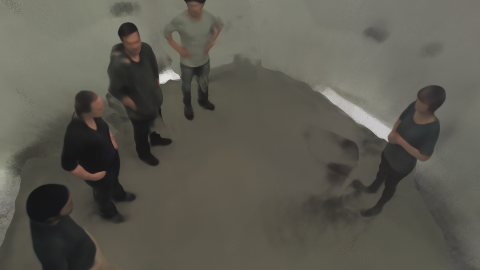} &
\includegraphics[width=2.5cm]{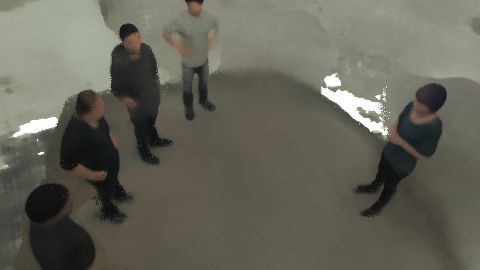} &
\includegraphics[width=2.5cm]{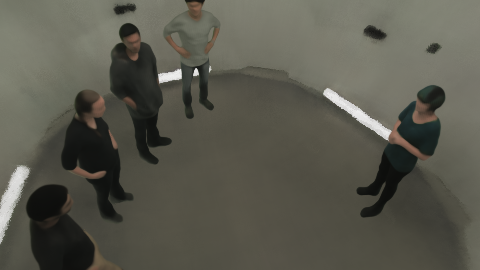} &
\includegraphics[width=2.5cm]{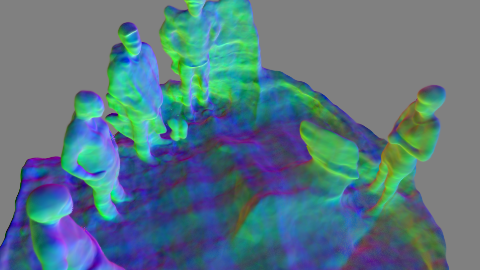} &
\includegraphics[width=2.5cm]{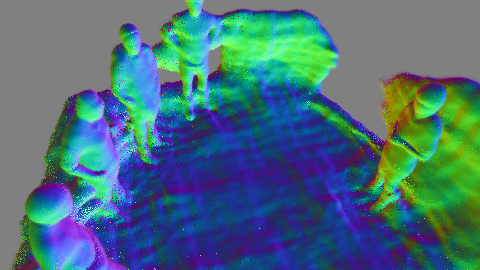} &
\includegraphics[width=2.5cm]{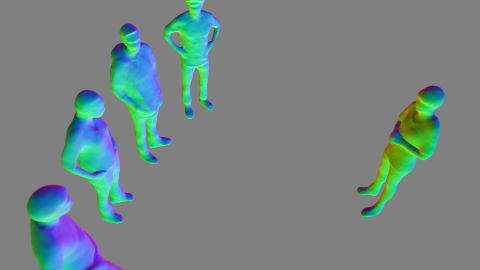} \\
Ground Truth &NeuS &VolSDF & Ours  & Neus &Volsdf & Ours\\
\end{tabular}}
 \caption{Qualitative comparison against NeuS~\cite{wang2021neus} and VolSDF~\cite{yariv2021volume} of synthesised novel views and reconstructed normal images of multiple humans on CMU Panoptic dataset \cite{Simon_2017_CVPR,Joo_2017_TPAMI}, using 5/10/15/20 training views.} 
\label{fig:cmu} 
\end{figure*}

\begin{table*}[h!] 
\normalsize
\begin{center} %
 \scalebox{0.82}{
\begin{tabular}{l|c|cccc|cccc|cccc}
\toprule[1.2pt]
Scene&Method  &\multicolumn{4}{c|} {\bf PSNR↑} & \multicolumn{4}{c|}{\bf SSIM↑}& \multicolumn{4}{c} {\bf LPIPS↓}  \\
 &&5&10   &15  &20 &5  &10  &15  &20&5 &10   &15 &20\\
\hline
 & NeuS &17.83& 18.84&19.39 &21.97&0.62&0.67&0.69&0.55&0.74&0.51&\textcolor{textcolor}{\textbf{0.49}}&\textcolor{textcolor}{\textbf{0.45}}  \\
1&VolSDF &17.50&18.08 &19.51 &22.31&0.64 &0.61&0.67&0.71&0.61&0.54&0.51&0.48\\
 &\textbf{Ours}&\textcolor{textcolor}{\textbf{18.41}}&\textcolor{textcolor}{\textbf{20.32}} &\textcolor{textcolor}{\textbf{21.60}}&\textcolor{textcolor}{\textbf{23.19}}&\textcolor{textcolor}{\textbf{0.67}} &\textcolor{textcolor}{\textbf{0.73}}&\textcolor{textcolor}{\textbf{0.73}}&\textcolor{textcolor}{\textbf{0.74}}&\textcolor{textcolor}{\textbf{0.55}} &\textcolor{textcolor}{\textbf{0.50}}&0.50&0.49\\
\cline{0-13}
 & NeuS&16.87& 18.51&19.40&21.05&0.60 &0.65&0.70&0.71&0.57&0.53&0.51&0.49 \\
2& VolSDF &16.36&17.52 &19.40 &21.60 &0.57&0.59&0.67&0.70&0.62&0.53&0.49&0.47\\
  &\textbf{Ours}&\textcolor{textcolor}{\textbf{19.72}}&\textcolor{textcolor}{\textbf{21.15}} &\textcolor{textcolor}{\textbf{21.40}}&\textcolor{textcolor}{\textbf{23.12}}&\textcolor{textcolor}{\textbf{0.70}} &\textcolor{textcolor}{\textbf{0.73}}&\textcolor{textcolor}{\textbf{0.73}}&\textcolor{textcolor}{\textbf{0.74}}&\textcolor{textcolor}{\textbf{0.50}} &\textcolor{textcolor}{\textbf{0.49}}&\textcolor{textcolor}{\textbf{0.48}}&\textcolor{textcolor}{\textbf{0.47}}\\
\cline{0-13}
 & NeuS &16.03&17.39&19.17 &21.21&0.56&0.61&0.70&0.73&0.62   &0.54&\textcolor{textcolor}{\textbf{0.47}}&\textcolor{textcolor}{\textbf{0.46}}\\
3& VolSDF &16.36&18.21& 19.56&21.06&0.57&0.59&0.64&0.68&0.62&0.52 &0.48&0.47\\
 &\textbf{Ours}&\textcolor{textcolor}{\textbf{18.57}}&\textcolor{textcolor}{\textbf{20.94}} &\textcolor{textcolor}{\textbf{21.86}}&\textcolor{textcolor}{\textbf{23.16}}&\textcolor{textcolor}{\textbf{0.66}} &\textcolor{textcolor}{\textbf{0.73}}&\textcolor{textcolor}{\textbf{0.74}}&\textcolor{textcolor}{\textbf{0.74}}&\textcolor{textcolor}{\textbf{0.52}} &\textcolor{textcolor}{\textbf{0.48}}&\textcolor{textcolor}{\textbf{0.47}}&0.47\\
\cline{0-13}
&NeuS & 14.16&17.14 &19.87&21.37&0.49&0.51&0.70&0.72&0.60&0.57&0.48&0.46\\
4& VolSDF &13.51& 17.07 &18.68&20.89&0.50&0.57&0.65&0.68&0.64&0.54&0.53&0.46\\
 &\textbf{Ours}&\textcolor{textcolor}{\textbf{19.54}}&\textcolor{textcolor}{\textbf{20.94}} &\textcolor{textcolor}{\textbf{21.35}}&\textcolor{textcolor}{\textbf{23.29}}&\textcolor{textcolor}{\textbf{0.69}} &\textcolor{textcolor}{\textbf{0.72}}&\textcolor{textcolor}{\textbf{0.73}}&\textcolor{textcolor}{\textbf{0.75}}&\textcolor{textcolor}{\textbf{0.50}} &\textcolor{textcolor}{\textbf{0.47}}&\textcolor{textcolor}{\textbf{0.47}}&\textcolor{textcolor}{\textbf{0.45}}\\
\cline{0-13}
 & NeuS &17.69& 18.60&20.03 &21.50&0.57 &0.62&0.69&0.70&0.55&0.54&0.50& \textcolor{textcolor}{\textbf{0.47}}\\
5& VolSDF &14.85& 17.32&19.04 &20.91&0.53&0.57&0.66&0.68&0.63&0.58&0.53&0.48\\
 &\textbf{Ours}&\textcolor{textcolor}{\textbf{19.34}}&\textcolor{textcolor}{\textbf{20.55}} &\textcolor{textcolor}{\textbf{21.08}}&\textcolor{textcolor}{\textbf{22.55}}&\textcolor{textcolor}{\textbf{0.67}} &\textcolor{textcolor}{\textbf{0.70}}&\textcolor{textcolor}{\textbf{0.72}}&\textcolor{textcolor}{\textbf{0.72}}&\textcolor{textcolor}{\textbf{0.51}} &\textcolor{textcolor}{\textbf{0.47}}&\textcolor{textcolor}{\textbf{0.47}}&\textcolor{textcolor}{\textbf{0.47}}\\
 \cline{0-13}
 & NeuS &16.52&17.79 &19.57 &21.42&0.57&0.62 &0.69	&0.72 &0.58	&0.54 &	0.49 &	\textcolor{textcolor}{\textbf{0.47}} \\
\textbf{Average}& 
VolSDF &15.81&17.68 &19.23&21.35 &0.56&0.59&	0.66 &	0.69 &0.62&	0.54	&0.50 &\textcolor{textcolor}{\textbf{0.47}}\\
 &\textbf{Ours}&\textcolor{textcolor}{\textbf{19.12}}&\textcolor{textcolor}{\textbf{20.78}} &\textcolor{textcolor}{\textbf{21.46}}&\textcolor{textcolor}{\textbf{23.06}}&\textcolor{textcolor}{\textbf{0.68}} &\textcolor{textcolor}{\textbf{0.72}}&\textcolor{textcolor}{\textbf{0.73}}&\textcolor{textcolor}{\textbf{0.74}}&\textcolor{textcolor}{\textbf{0.52}}
 &\textcolor{textcolor}{\textbf{0.48}}&\textcolor{textcolor}{\textbf{0.48}}&\textcolor{textcolor}{\textbf{0.47}}\\
\bottomrule[1.2pt]
\end{tabular}}
\caption{Comparison against NeuS~\cite{wang2021neus} and VolSDF~\cite{yariv2021volume} on the CMU Panoptic dataset \cite{Simon_2017_CVPR,Joo_2017_TPAMI}, using 5/10/15/20 views for training.}
\label{tab:realhuman}
\end{center}
\vspace{-8mm}
\end{table*}


\section{Results}
\label{sec:results}
In this section we provide implementation details (Sec.~\ref{sec:results_impl_details}), and demonstrate our performance against baselines on real (Sec.~\ref{sec:results_real}) 
and synthetic (Sec.~\ref{sec:results_synth}) 
datasets, in terms of novel-view synthesis, visual reconstructions, and geometry error. Finally, we show ablation studies (Sec.~\ref{sec:results_ablation}) that demonstrate the importance of each of the proposed components. 

\subsection{Implementation Details}
\label{sec:results_impl_details}
Our method was implemented using PyTorch \cite{paszke2019pytorch}, and trained on a Quadro RTX 5000 GPU. We use ADAM optimizer \cite{kingma2014adam} with a learning rate ranging from $5 \times 10^{-4}$ to $2.5 \times 10^{-5}$, controlled by cosine decay schedule. Our network architecture follows \cite{yariv2020multiview, mildenhall2020nerf}. For a fair comparison, we sample 256 rays per batch and follow the coarse and fine sampling strategy of \cite{wang2021neus}. More network structure and training details are shown in the supplementary material. 

\subsection{Real Multi-Human Dataset}
\label{sec:results_real}
We first evaluate our approach on the CMU Panoptic Dataset~\cite{Simon_2017_CVPR,Joo_2017_TPAMI}. 
Our experiments were performed on five different scenes, where each scene originally includes 30 views containing 3/4/5/6/7 people. The training views were randomly extracted from the HD sequences `Ultimatum' and `Haggling'.  We uniformly choose 5/10/15/20 views for training and the rest 25/20/15/10 views for testing. We compare with two major baselines: NeuS \cite{wang2021neus} and VolSDF\cite{yariv2021volume}, both in terms of novel-view synthesis and geometry reconstructions (qualitatively). For quantitative evaluation, we report three commonly used image metrics: peak signal-to-noise ratio (PSNR)~\cite{hore2010image}, structural similarity index (SSIM)~\cite{wang2004image} and learned perceptual image patch similarity (LPIPS)~\cite{zhang2018unreasonable}. For qualitative comparison, both rendered images and rendered normal images are shown. 

\noindent
\textbf{Comparison with baselines.} 
Tab.~\ref{tab:realhuman} demonstrates novel view synthesis results with different training views (5/10/15/20) compared to the baselines. Our proposed method outperforms these in PSNR and SSIM in all the scenes, and consistently performs better or equal in terms of LPIPS. For qualitative comparison, we demonstrate both rendered novel views and normal images in Fig.~\ref{fig:cmu}. As seen here, when given 5/10 training views the baseline methods fail to reconstruct a good geometry or render a realistic appearance. Although the quality of the geometries improves with 15/20 training views, the results exhibit missing body parts or can mix the background with the subjects. On the other hand, our method can reconstruct a complete geometry for all humans in all sparse-view cases.  

\begin{figure}[h!]
\vspace{-3mm}
\centering
\includegraphics[width=0.9\linewidth]{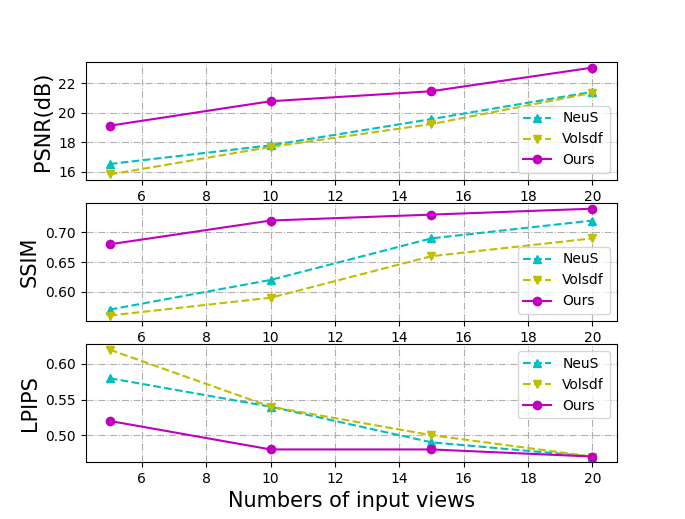}
\vspace{-2mm}

  \caption{Quantitative comparison of average PSNR (↑), SSIM (↑) and LPIPS (↓) with increased number of training views. }
  \label{fig:increase}
\end{figure}

Fig.~\ref{fig:increase} additionally shows the relationship between the number of training views and the quality of the synthesized images. The fewer the number of views, the harder it is for all methods to reconstruct high-quality images, whereas our approach is more robust to fewer training views. For denser inputs~(\eg more than 20 views), our method reaches similar albeit slightly better performance than the baselines, since the proposed work focuses on sparse scenarios. 

\begin{figure}[h]
\centering
\setlength{\tabcolsep}{0pt}
\renewcommand{\arraystretch}{0} 
\begin{tabular}{ccccc}
\includegraphics[width=1.68cm]{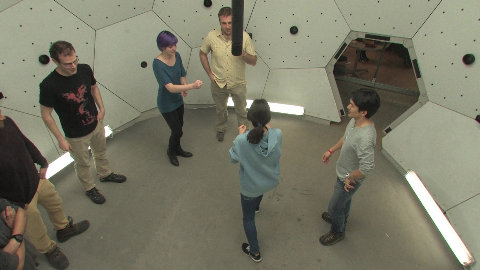} &
\includegraphics[width=1.68cm]{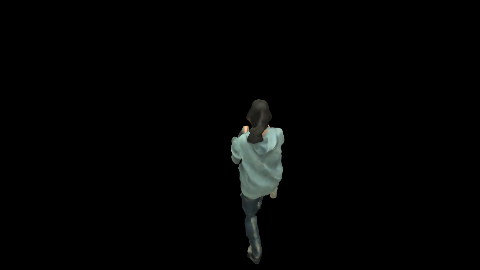} &
\includegraphics[width=1.68cm]{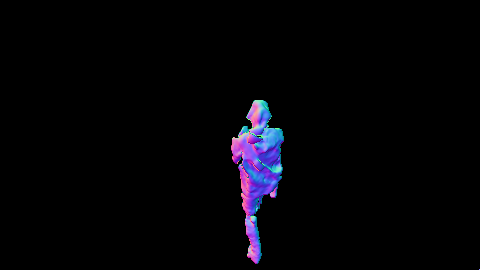}&
\includegraphics[width=1.68cm]{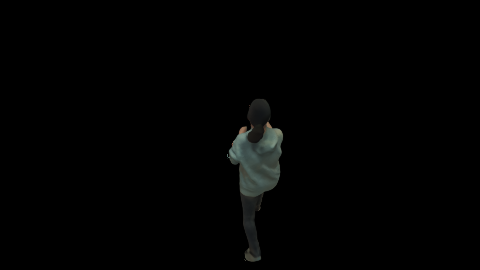} &
\includegraphics[width=1.68cm]{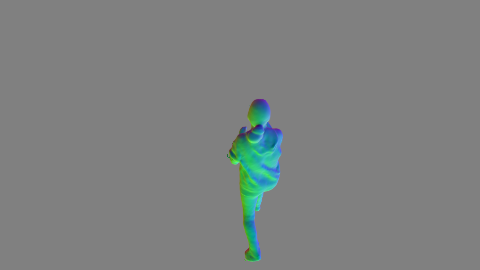} \\
\includegraphics[width=1.68cm]{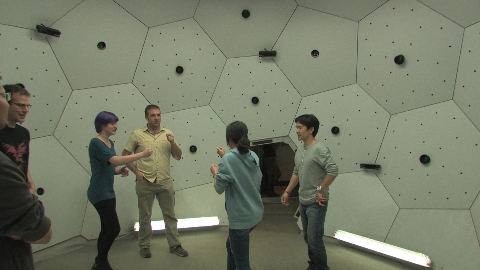} &
\includegraphics[width=1.68cm]{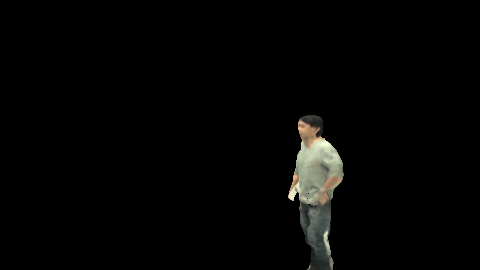} &
\includegraphics[width=1.68cm]{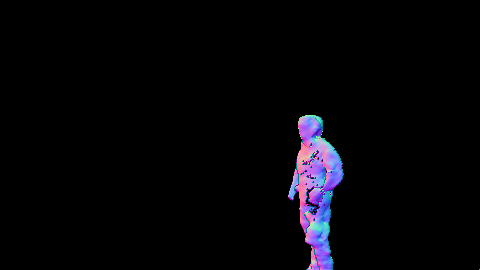}&
\includegraphics[width=1.68cm]{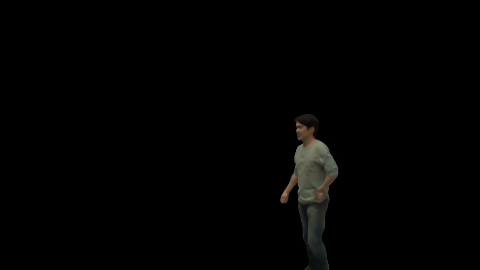} &
\includegraphics[width=1.68cm]{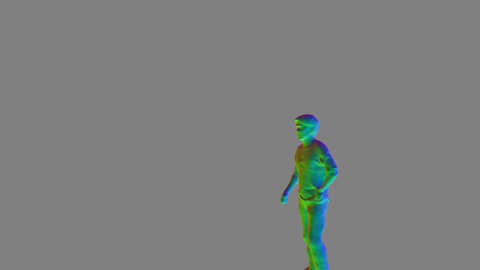} \\
\includegraphics[width=1.68cm]{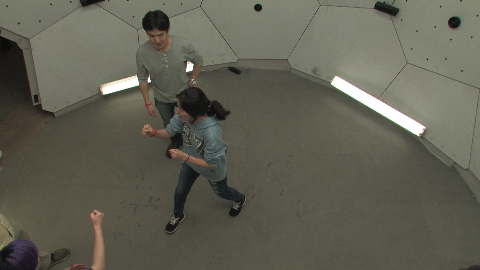} &
\includegraphics[width=1.68cm]{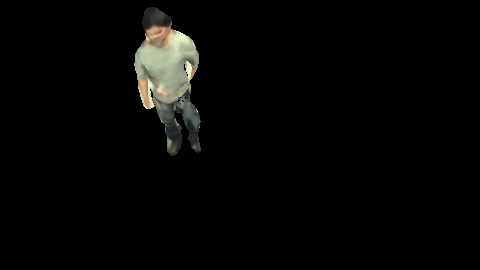} &
\includegraphics[width=1.68cm]{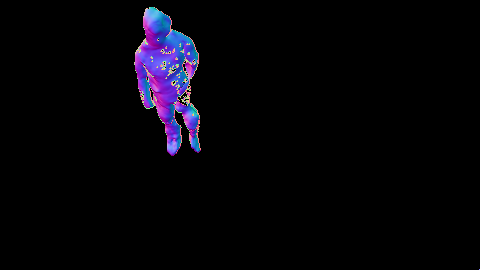}&
\includegraphics[width=1.68cm]{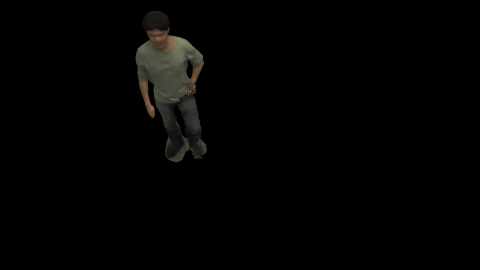} &
\includegraphics[width=1.68cm]{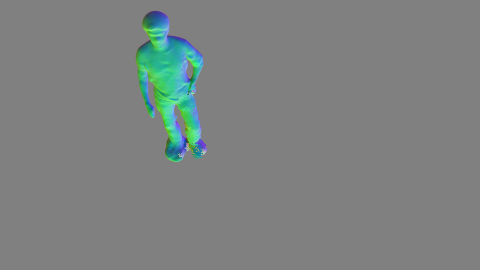} \\
GT& ARAH  & ARAH & Ours &  Ours\\
\end{tabular}
\vspace{-2mm}
\caption{Comparison against ~\cite{wang2022arah} from 5 training views. PSNRs for the 3 examples are respectively: 26.97/\textbf{29.56}, 27.48/\textbf{33.66}, 24.36/\textbf{30.56} (ARAH/\textbf{Ours}).}
\label{fig:arah} 
\end{figure}

\noindent
\textbf{Comparison to single human NeRF.} 
We compare our method to the single human nerf state-of-the-art method ARAH~\cite{wang2022arah}. We note that adapting such methods to our setup requires tedious manual pre-processing (detecting and segmenting people, associating detections across views), which is not required by our approach.
We run a separate ARAH model for each person in the scene using 5 training images (see supp. mat.). Fig.~\ref{fig:arah} shows novel view and reconstruction results. Learning for each person separately implies providing erroneous supervision to the model whenever the person is occluded in the scene or segmentation masks are not accurate. As a result, ARAH's renderings and geometry display many artifacts compared to our results.
Conversely, our method avoids this by learning through rendering the union of SMLP bounding boxes conjointly. We also noticed that ARAH's results are very sensitive to the sparsity and choice of the training views.

\begin{table}[h!]
\begin{center}
\normalsize
\scalebox{0.8}{
\begin{tabular}{c|l|ccc}
\toprule[1.2pt]
Scene& Method & {\bf PSNR↑} & {\bf SSIM↑}& {\bf LPIPS↓}\\
\hline
& InfoNeRF  &  14.64 &0.50 &0.64\\
1&NeuS w/ info &17.98&0.65&0.58\\
&Ours& \bf{18.41}& \bf{0.67} & \bf{0.55}    \\
\hline
& InfoNeRF&14.21&0.49&0.63  \\
2&NeuS w/ info&18.21&0.64&0.57\\
&Ours&\bf{19.72} & \bf{0.70} & \bf{0.50} \\
\hline
&InfoNeRF&13.78&0.45&0.63  \\
3&NeuS w/ info &16.31&0.59&0.60\\
&Ours & \bf{18.57} & \bf{0.66} & \bf{0.52}    \\
\hline
& InfoNeRF  &12.26&0.41&0.68  \\
4&NeuS w/ info&14.42&0.51&0.60\\
&Ours & \bf{19.54} & \bf{0.69} & \bf{0.50} \\
\hline
& InfoNeRF &12.17 &0.45&0.63 \\
5&NeuS w/ info &17.89&0.60&0.61\\
&Ours & \bf{19.34} & \bf{0.67} & \bf{0.51}    \\
\hline
& InfoNeRF  &13.61&0.46&0.64  \\
Ave&NeuS w/ info &16.96&0.60&0.59\\
&Ours & \bf{19.12} & \bf{0.68} & \bf{0.52}    \\

\bottomrule[1.2pt]
\end{tabular}}
 \caption{Comparison against sparse-view NeRF approaches: InfoNeRF~\cite{kim2022infonerf} and NeuS with InfoNeRF's regularizations, on the CMU Panoptic dataset~\cite{Simon_2017_CVPR,Joo_2017_TPAMI} using 5 training views. }
 \label{tab:sparse}
\end{center}
\end{table}

\noindent
\textbf{Comparison with sparse NeRF.} 
We further compare with a recent NeRF method that was specifically designed to handle sparse views, namely  InfoNeRF~\cite{kim2022infonerf}. We compare both against the original InfoNeRF, and a version of NeuS trained with InfoNeRF's regularization. For this experiment, we use again the CMU Panoptic dataset~\cite{Simon_2017_CVPR,Joo_2017_TPAMI} with five training views. 
Tab.~\ref{fig:abl} shows that, compared to InfoNeRF and NeuS with InfoNeRF's regularization, our method improves the rendering quality in all of the scenes.

\begin{figure*}[h!]
\centering
\vspace{-3mm}
\def\tabularxcolumn#1{m{#1}}
\setlength{\tabcolsep}{0pt}
\renewcommand{\arraystretch}{0} 
\begin{tabular}{c cccc ccc}
\rotatebox{90}{\quad 10} 
\includegraphics[width=2.5cm]{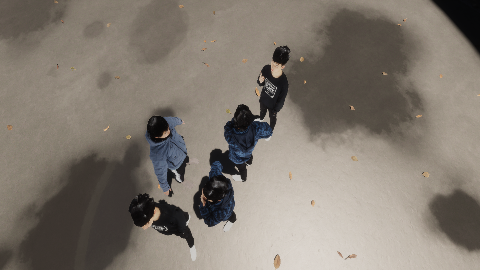} &
\includegraphics[width=2.5cm]{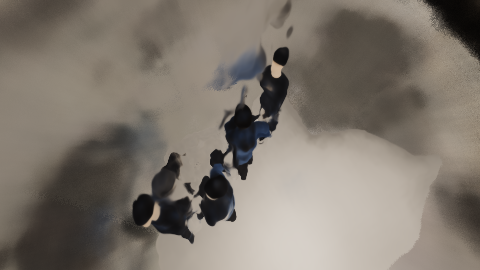} &
\includegraphics[width=2.5cm]{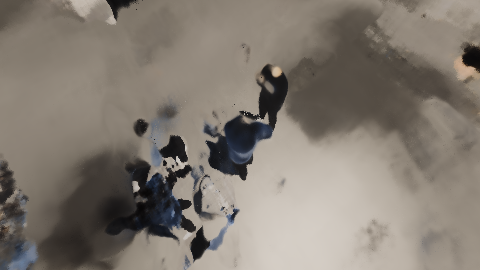} &
\includegraphics[width=2.5cm]{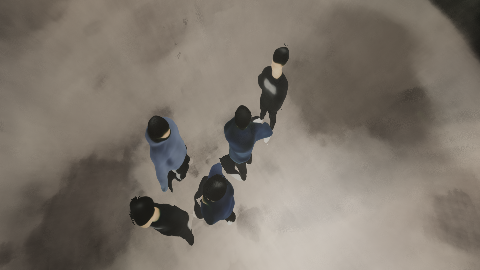} &
\includegraphics[width=2.5cm]{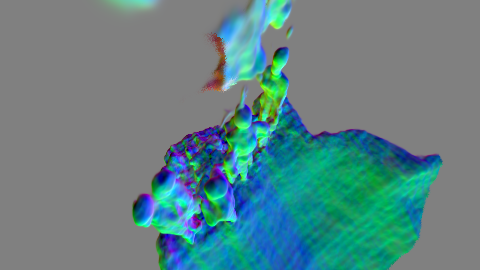} &
\includegraphics[width=2.5cm]{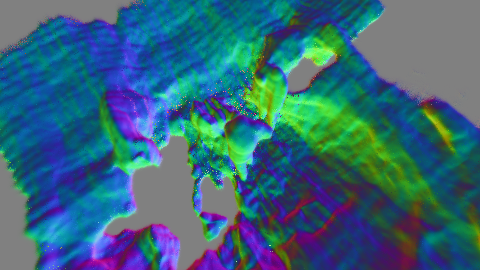} &
\includegraphics[width=2.5cm]{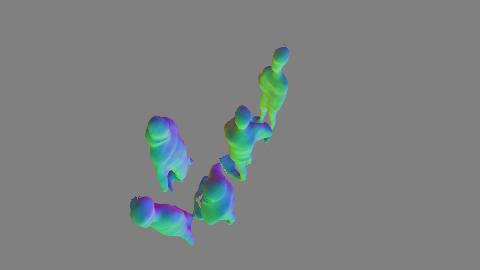} \\
\rotatebox{90}{\quad 15}
\includegraphics[width=2.5cm]{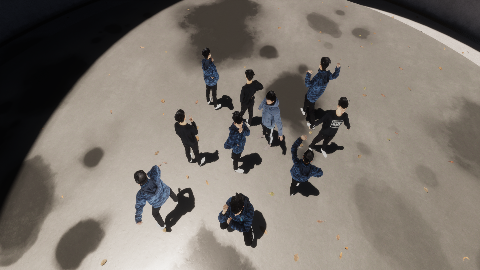} &
\includegraphics[width=2.5cm]{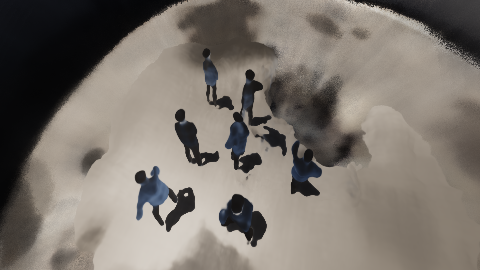} &
\includegraphics[width=2.5cm]{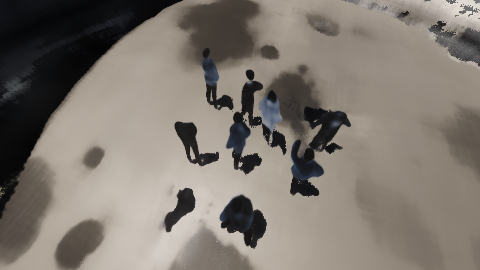} &
\includegraphics[width=2.5cm]{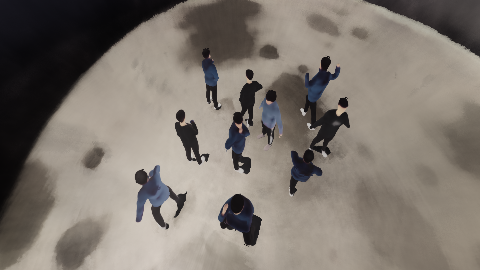} &
\includegraphics[width=2.5cm]{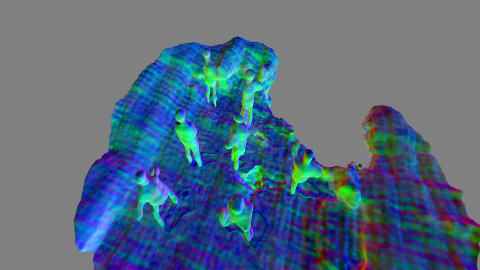} &
\includegraphics[width=2.5cm]{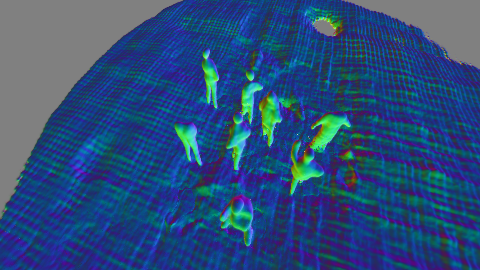} &
\includegraphics[width=2.5cm]{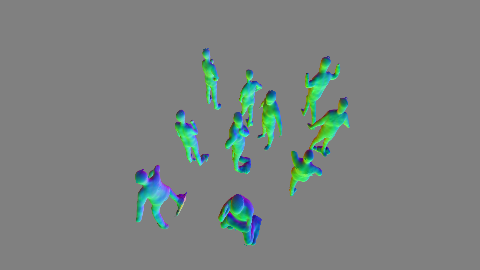} \\
Ground Truth &Neus &Volsdf & Ours  & Neus &Volsdf & Ours\\
\end{tabular}
 \caption{Qualitative comparison of synthesised novel views and reconstructed normal images on the synthetic dataset (MultiHuman-Dataset~\cite{zheng2021deepmulticap}) with 10 and 15 training views respectively.} 
\label{fig:syn} 
\end{figure*}

\begin{figure*}[ht]
\centering
\def\tabularxcolumn#1{m{#1}}
\setlength{\tabcolsep}{0pt}
\renewcommand{\arraystretch}{0} 
\begin{tabular}{ccccccc}
\includegraphics[width=2.5cm]{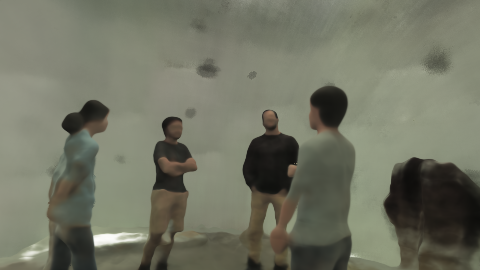} &
\includegraphics[width=2.5cm]{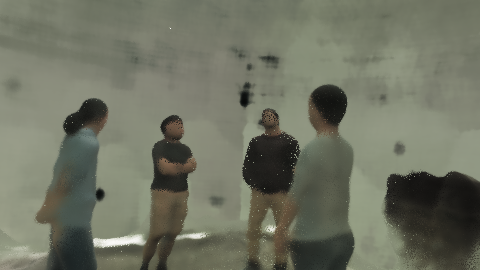} &
\includegraphics[width=2.5cm]{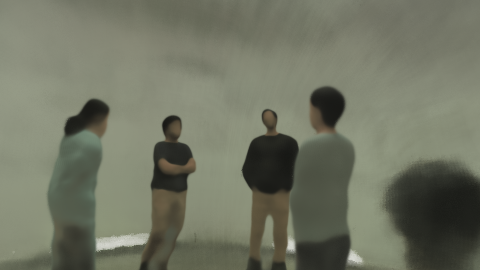} &
\includegraphics[width=2.5cm]{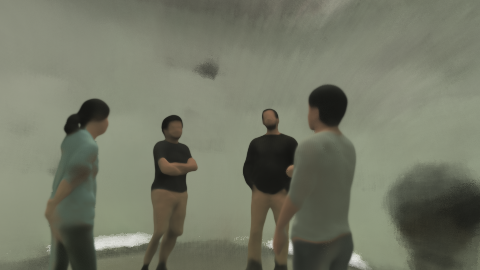} &
\includegraphics[width=2.5cm]{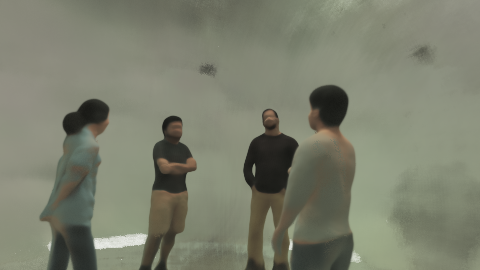}&
\includegraphics[width=2.5cm]{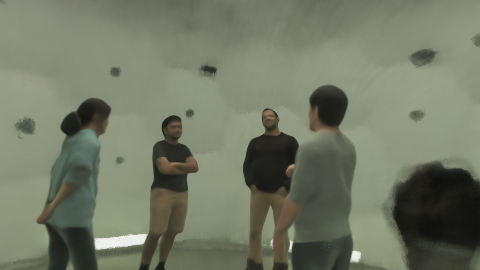} &
\includegraphics[width=2.5cm]{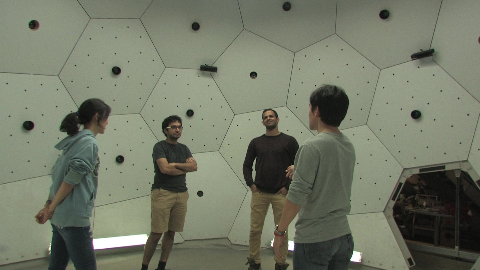}  \\ 
\includegraphics[width=2.5cm]{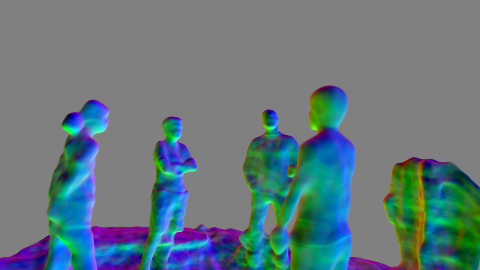} &
\includegraphics[width=2.5cm]{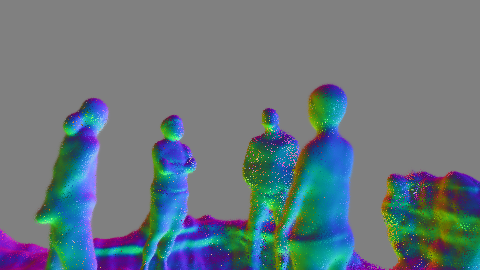} &
\includegraphics[width=2.5cm]{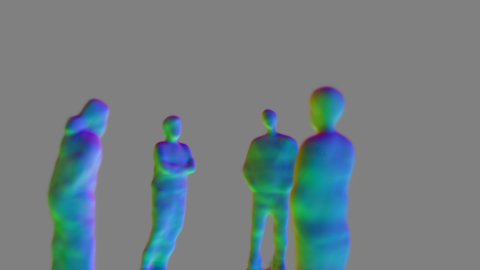} &
\includegraphics[width=2.5cm]{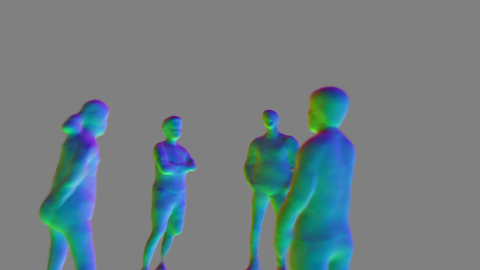} &
\includegraphics[width=2.5cm]{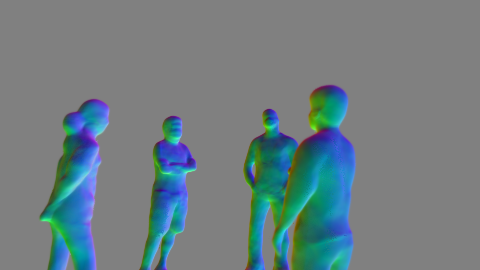}&
\includegraphics[width=2.5cm]{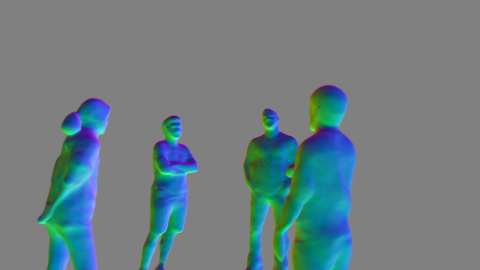} &\\
NeuS & VolSDF &w/o geometry  &w/o ray loss &w/o saturation  & Ours (Full) & Ground Truth\\
\end{tabular}
 \caption{Ablation study on CMU Panoptic dataset \cite{Simon_2017_CVPR,Joo_2017_TPAMI}. Comparison against our method without geometric regularization (w/o geometry), our method without ray consistency regularization (w/o ray loss), and our method without saturation regularization (w/o saturation).}
\label{fig:abl} 
\end{figure*}


\begin{table*}[h!]
\begin{center} %
\normalsize
\scalebox{0.82}{
\begin{tabular}{l|c|ccc|ccc|ccc|c }
\toprule[1.2pt]
$\#$ Humans & Method  &\multicolumn{3}{c|} {\bf PSNR↑} &  \multicolumn{3}{c|}{\bf SSIM↑}& \multicolumn{3}{c|} {\bf LPIPS↓} & {\bf Chamfer ↓} \\ 
 &&5   &10  &15   &5  &10  &15 &5   &10 &15 &15\\
\hline
 & NeuS& 14.04&17.89 &23.25&0.63&0.72&0.84&\textcolor{textcolor}{\textbf{0.55}}& 0.53&0.44&0.308\\
1 & VolSDF & 13.93&21.75 &25.89&0.61&0.81&0.86&\textcolor{textcolor}{\textbf{0.55}}&0.51&0.44&0.019\\
 &  \textbf{Ours}&\textcolor{textcolor}{\textbf{15.36} }&\textcolor{textcolor}{\textbf{23.85}}&\textcolor{textcolor}{\textbf{26.28}} &\textcolor{textcolor}{\textbf{0.65}}&\textcolor{textcolor}{\textbf{0.84}}&\textcolor{textcolor}{\textbf{0.87}}&\textcolor{textcolor}{\textbf{0.55}}&\textcolor{textcolor}{\textbf{0.43}}&\textcolor{textcolor}{\textbf{0.41}}&\textcolor{textcolor}{\textbf{0.018}}\\
\hline
 & NeuS &14.15&18.14 &18.54 &0.61&0.72&0.72&0.54&0.46&0.44&0.321\\
5 & VolSDF &12.97&15.11&18.59&0.58&0.63&0.73&0.56&0.55&0.47&0.151\\
 &  \textbf{Ours}&\textcolor{textcolor}{\textbf{17.63} }&\textcolor{textcolor}{\textbf{20.10}}&\textcolor{textcolor}{\textbf{20.33}} &\textcolor{textcolor}{\textbf{0.71}}&\textcolor{textcolor}{\textbf{0.79}}&\textcolor{textcolor}{\textbf{0.77}}&\textcolor{textcolor}{\textbf{0.47}}&\textcolor{textcolor}{\textbf{0.40}}&\textcolor{textcolor}{\textbf{0.40}}&\textcolor{textcolor}{\textbf{0.020}}\\
\hline
 & NeuS &14.09&15.69&19.27&0.58&0.65&0.75&0.52&0.48&0.42&0.383\\
10 & VolSDF & 12.66&16.99 &19.30&0.56&0.70&0.77&0.56&0.50&0.41&0.248\\
 &  \textbf{Ours}&\textcolor{textcolor}{\textbf{16.52}} &\textcolor{textcolor}{\textbf{18.39}}&\textcolor{textcolor}{\textbf{21.01}} &\textcolor{textcolor}{\textbf{0.65}}&\textcolor{textcolor}{\textbf{0.71}}&\textcolor{textcolor}{\textbf{0.80}}&\textcolor{textcolor}{\textbf{0.50}}&\textcolor{textcolor}{\textbf{0.44}}&\textcolor{textcolor}{\textbf{0.37}}&\textcolor{textcolor}{\textbf{0.043}}\\
 \hline
 & NeuS &14.09&17.24&20.35 &0.60	&0.70	&0.77 &0.54 &0.49 &0.43&0.337\\
\textbf{Average}& VolSDF&13.18 &17.95&	21.26 &0.58 &0.71 &	0.79 &0.56	&0.52	&0.44&0.139\\
 &\textbf{Ours}&\textcolor{textcolor}{\textbf{16.50}} &\textcolor{textcolor}{\textbf{20.78}}&\textcolor{textcolor}{\textbf{22.54}} &\textcolor{textcolor}{\textbf{0.67}}&\textcolor{textcolor}{\textbf{0.78}}&\textcolor{textcolor}{\textbf{0.81}} &\textcolor{textcolor}{\textbf{0.51}}&\textcolor{textcolor}{\textbf{0.42}}&\textcolor{textcolor}{\textbf{0.39}}&\textcolor{textcolor}{\textbf{0.026}}\\
\bottomrule[1.2pt]
\end{tabular}}
\caption{Comparison against NeuS~\cite{wang2021neus} and VolSDF~\cite{yariv2021volume} on the synthetic dataset, for different number of humans in the scene. We measure novel-view synthesis quality in terms of PSNR, SSIM and LIPIS, as well as geometry error in terms of Chamfer distance.}
\label{tab:syn}
\end{center}
\end{table*}

\subsection{Synthetic Dataset}
\label{sec:results_synth}
Based on the MultiHuman-Dataset \cite{tao2021function4d,zheng2021deepmulticap}, we used Unity 3D to create a synthetic dataset with 29 cameras arranged in a great circle. This includes three scenes with similar backgrounds but different camera locations and orientations. Each of the scenes contains 1/5/10 humans respectively. We train with 5/10/15 views on each scene and test with 14 fixed views. Tab.~\ref{tab:syn} reports the average error for all testing views in PSNR, SSIM and LPIPS metrics. Our method reaches state-of-the-art performance on synthesized novel-view results. Fig.~\ref{fig:syn}  shows generated novel views and corresponding normal images using 10/15 training images. Our approach can reconstruct complete geometry of all humans in the scene, while the baseline methods might miss some of the people when they have similar color with the background, \eg the shadow area in Fig.~\ref{fig:syn}. 

In the 5/10 input views case, the baseline methods usually fail to reconstruct the full geometry of humans due to the sparse inputs. Thus, we report Chamfer distance in Tab.~\ref{tab:syn} only for the 15-views case. Since the baseline methods usually contain extra floor, for a fair comparison, we sample points from ground-truth meshes and compute the distance towards the reconstructed mesh for all methods. We report the bi-directional Chamfer distance in the supplementary material. Tab.~\ref{tab:syn} shows that, with an increasing number of humans in the scene, the quality of the reconstructed geometry of all methods decreases. However, compared with the baselines, our method can better handle multiple human scenes, achieving an order of magnitude less error.

\subsection{Ablation Study}
\label{sec:results_ablation}

To prove the effectiveness of our proposed components we performed ablation studies on the CMU Panoptic dataset~\cite{Simon_2017_CVPR,Joo_2017_TPAMI}. We demonstrate quantitative comparisons in Tab.~\ref{tab:abl} and qualitative results in Fig.~\ref{fig:abl}. We test the following settings: 

\noindent
\textbf{Without geometry regularization (``w/o geometry'').} We compare our full model against the model without geometry regularization (Sec.~\ref{sec:method_geometric_init}) and SDF uncertainty regularization (Eq.~\ref{eq:sdfloss}). We can see here that, although the method is still capable of isolating humans thanks to the bounding box rendering, both geometry and novel views are much less accurate, and the rendered images exhibit background artifacts and overly smooth results. 

\noindent
\textbf{Without ray consistency loss (``w/o ray loss'').} Here we remove the proposed ray consistency loss, without which the average rendering quality also degrades.

\noindent
\textbf{Without saturation loss.} Finally, we remove the saturation loss from our methods, which decreases by about 0.5 in PSNR on average.  Fig.~\ref{fig:abl} shows that, without this, the image tone can contain artifacts due to changes in lighting (see for example the back of the rightmost subject).


\begin{table}[h!]
\begin{center}
\normalsize
\scalebox{0.8}{
\begin{tabular}{l|cc |cc |cc}
\toprule[1.2pt]
 Method &\multicolumn{2}{c|} {\bf PSNR↑} &  \multicolumn{2}{c|}{\bf SSIM↑}& \multicolumn{2}{c} {\bf LPIPS↓}\\
 &5   &15    &5  &15   &5  &15 \\
\hline
 Neus \cite{wang2021neus}& 16.87&19.40 &0.60&0.70 &0.51 &0.53 \\
 Volsdf\cite{yariv2021volume}&16.03&19.40 &0.53&0.67 &0.60&0.49\\
\hline
 w/o Geometry&17.54  &20.28&0.60&0.70 &0.53&0.48\\
 w/o Ray loss&19.07&20.95& 0.67  &0.72&0.52&0.47  \\
 w/o Saturation&18.95&20.92&0.65&0.72 &0.54&0.49\\
\hline
{\bf Ours(Full)} &  {\bf 19.72 }& {\bf 21.40 } &  {\bf 0.70} &  {\bf 0.73}& {\bf 0.50 }& {\bf 0.48}\\
\bottomrule[1.2pt]
\end{tabular}}
 \vspace{-1mm}
 \caption{Ablation study on the CMU Panoptic dataset \cite{Simon_2017_CVPR,Joo_2017_TPAMI} with 5/15 training views respectively. Comparison against our method without geometric regularization (w/o
Geometry), without ray consistency regularization (w/o Ray loss), and without saturation regularization (w/o Saturation).}
 \label{tab:abl}
\end{center}
\end{table}

\section{Conclusion}
\label{sec:conclusion}
We presented an approach for novel view synthesis of multiple humans from a sparse set of input views. To achieve this, we proposed geometric regularizations that improve geometry training by leveraging a pre-computed SMPL model, along with a patch-based ray consistency loss and a saturation loss that help with novel-view renderings in the sparse-view setting. Our experiments showed state-of-the-art performance for multiple human geometry and appearance reconstruction on real multi-human dataset (CMU Panoptic~\cite{Simon_2017_CVPR,Joo_2017_TPAMI}) and on synthetic data (MultiHuman-Dataset \cite{zheng2021deepmulticap}). Our method still has several limitations. For instance, our method does not model close human interactions, as this is a much more challenging case. Addressing this is an interesting direction for future work.




{\small
\bibliographystyle{ieee_fullname}
\bibliography{egbib}
}

\twocolumn[\centering \Large{\textbf{Few-Shot Multi-Human Neural Rendering Using Geometry Constraints\\ Supplementary Material}}\vspace{30pt}]


\begin{figure*}[t!]
\setlength{\linewidth}{\textwidth}
\setlength{\hsize}{\textwidth}
\centering
\def\tabularxcolumn#1{m{#1}}
\setlength{\tabcolsep}{0pt}
\renewcommand{\arraystretch}{0} 
\scalebox{1}{
\begin{tabular}{c ccccccc}
\rotatebox{90}{\quad remove} 
\includegraphics[width=2.5cm]{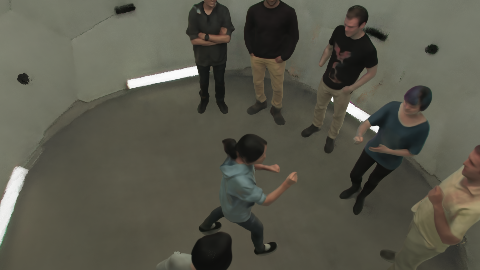} &
\includegraphics[width=2.5cm]{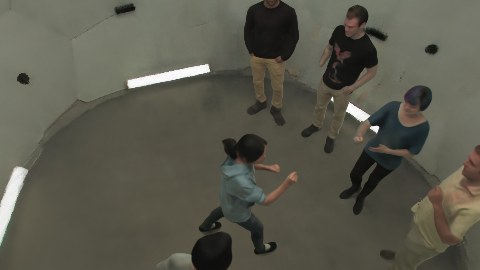} &
\includegraphics[width=2.5cm]{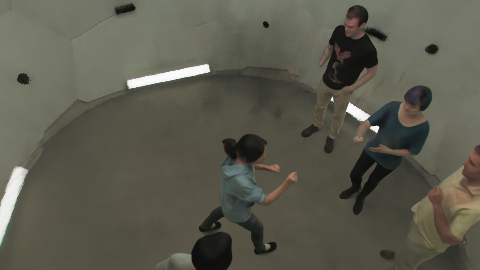} &
\includegraphics[width=2.5cm]{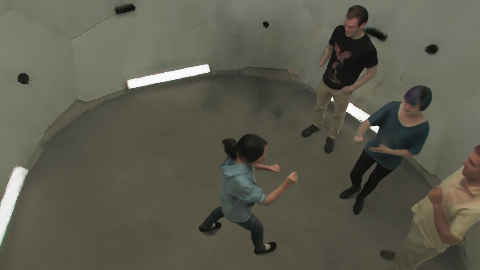} &
\includegraphics[width=2.5cm]{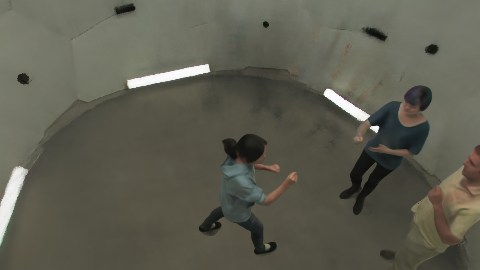} &
\includegraphics[width=2.5cm]{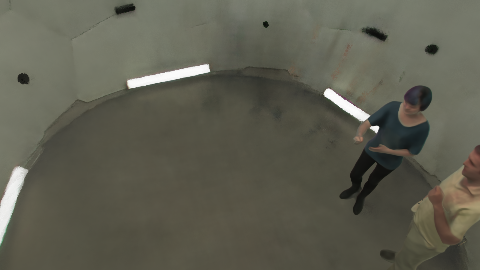} &
\includegraphics[width=2.5cm]{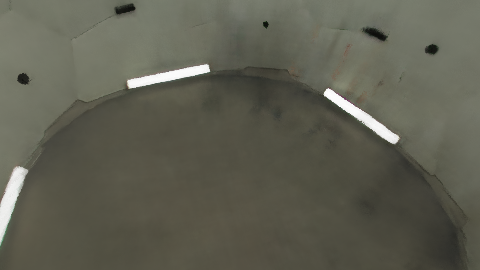}\\
\rotatebox{90}{\quad remove}
\includegraphics[width=2.5cm]{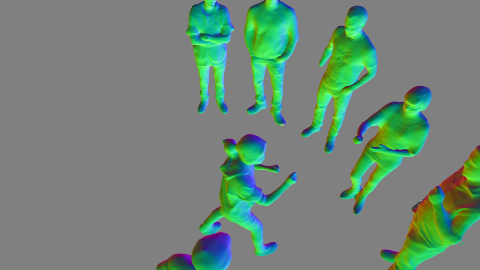} &
\includegraphics[width=2.5cm]{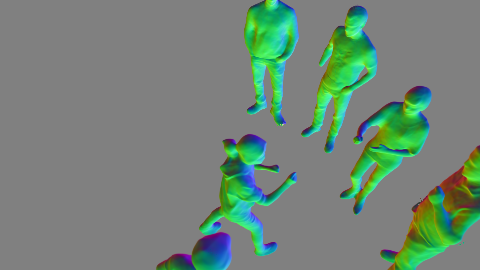} &
\includegraphics[width=2.5cm]{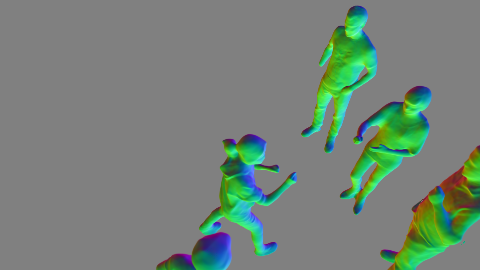} &
\includegraphics[width=2.5cm]{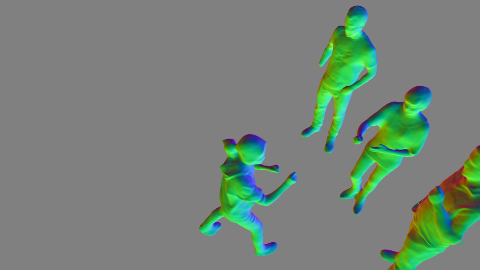} &
\includegraphics[width=2.5cm]{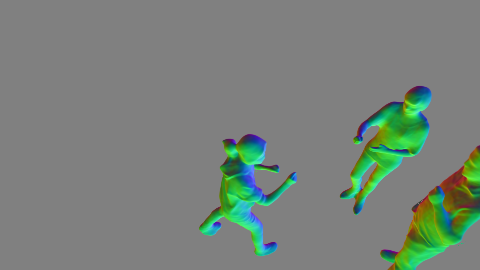} &
\includegraphics[width=2.5cm]{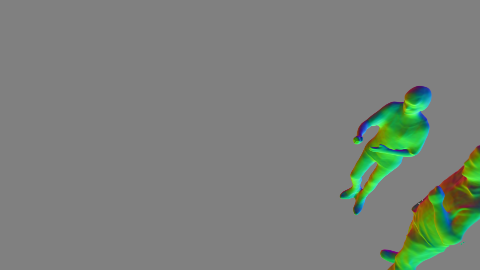} &
\includegraphics[width=2.5cm]{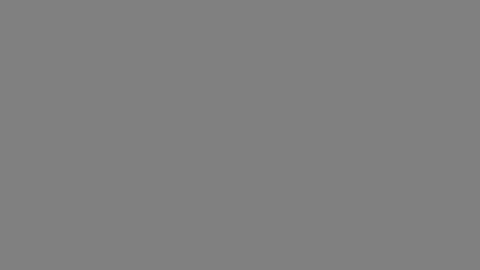} \\
\rotatebox{90}{\quad trans} 
\includegraphics[width=2.5cm]{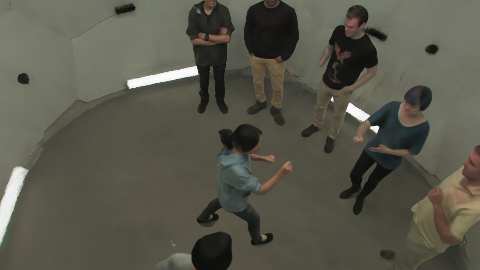} &
\includegraphics[width=2.5cm]{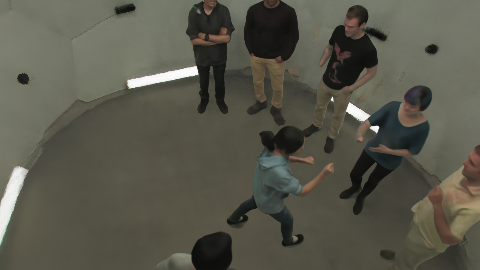} &
\includegraphics[width=2.5cm]{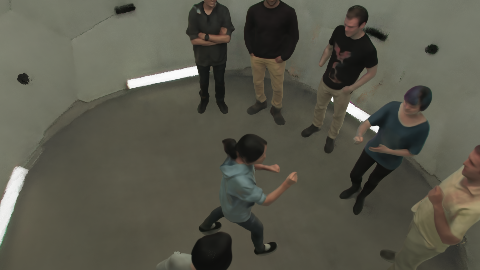} &
\includegraphics[width=2.5cm]{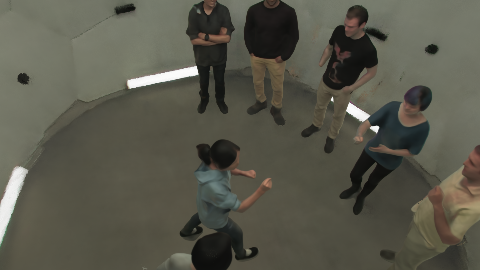} &
\includegraphics[width=2.5cm]{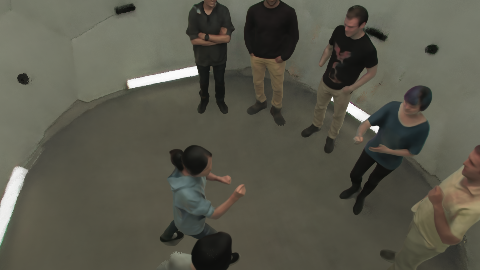} &
\includegraphics[width=2.5cm]{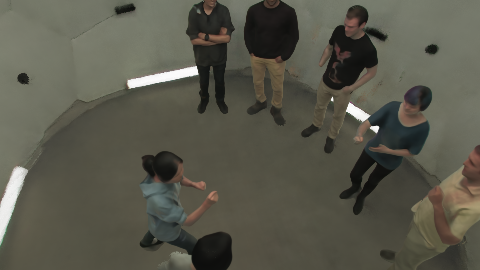} &
\includegraphics[width=2.5cm]{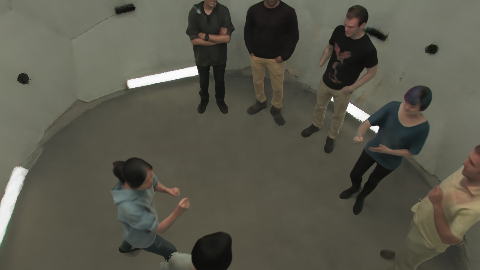} \\
\rotatebox{90}{\quad trans} 
\includegraphics[width=2.5cm]{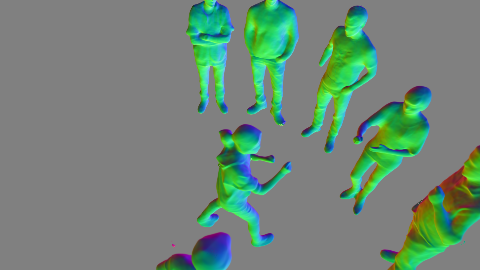} &
\includegraphics[width=2.5cm]{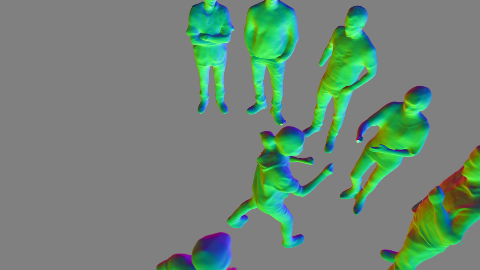} &
\includegraphics[width=2.5cm]{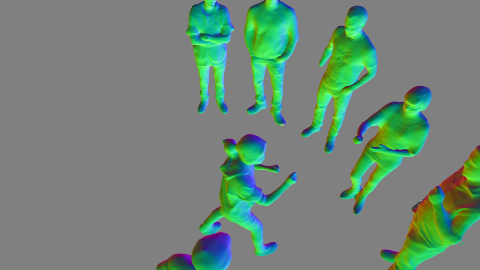} &
\includegraphics[width=2.5cm]{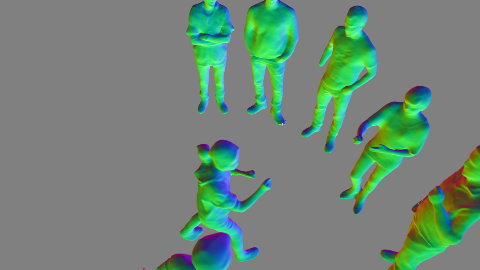} &
\includegraphics[width=2.5cm]{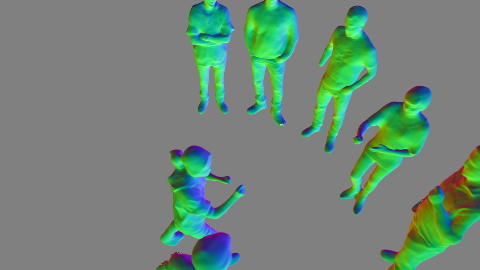} &
\includegraphics[width=2.5cm]{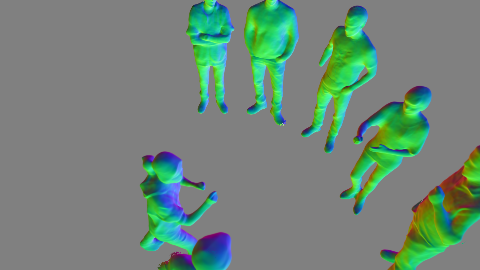} &
\includegraphics[width=2.5cm]{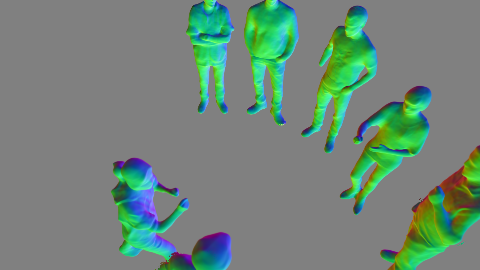} \\
\rotatebox{90}{\quad rotate} 
\includegraphics[width=2.5cm]{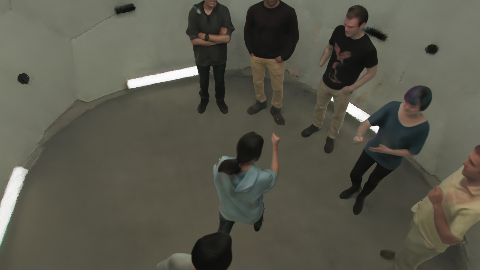} &
\includegraphics[width=2.5cm]{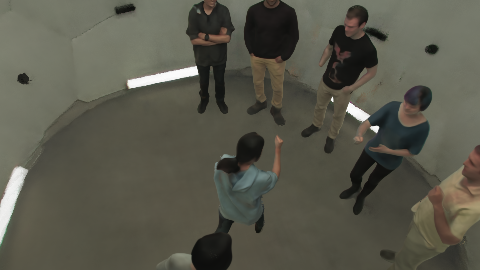} &
\includegraphics[width=2.5cm]{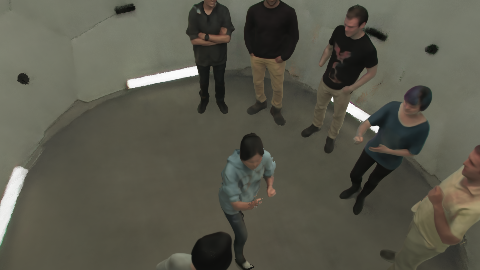} &
\includegraphics[width=2.5cm]{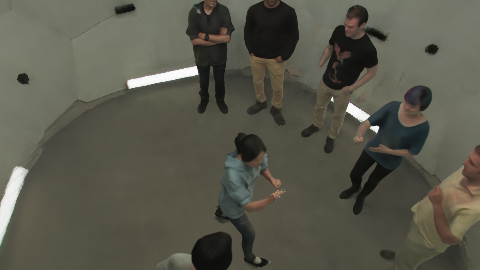} &
\includegraphics[width=2.5cm]{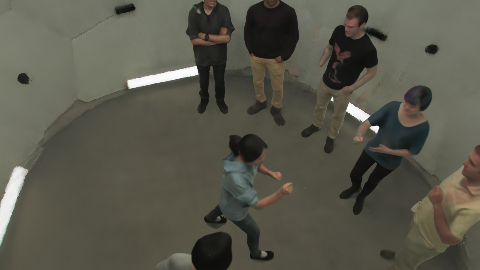} &
\includegraphics[width=2.5cm]{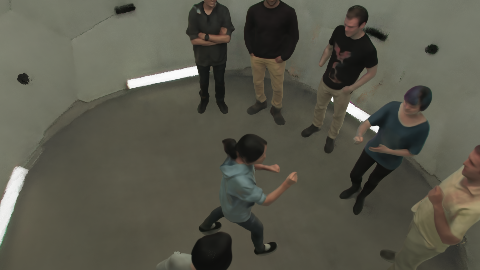} &
\includegraphics[width=2.5cm]{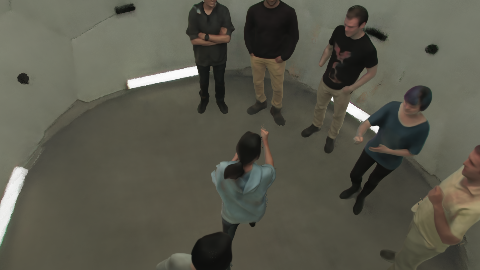} \\
\rotatebox{90}{\quad rotate} 
\includegraphics[width=2.5cm]{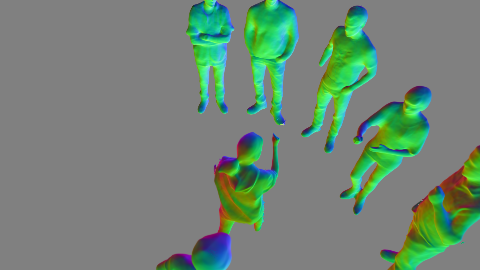} &
\includegraphics[width=2.5cm]{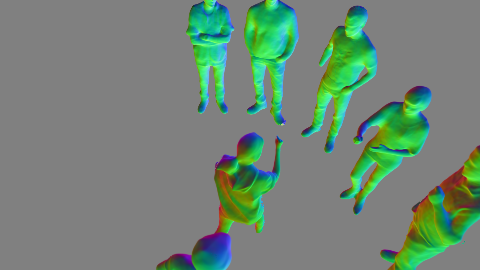} &
\includegraphics[width=2.5cm]{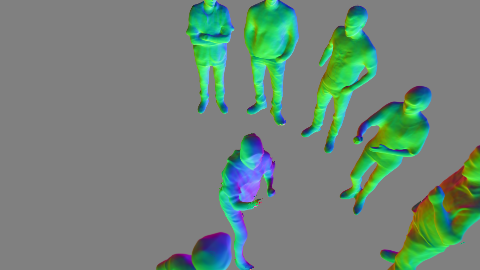} &
\includegraphics[width=2.5cm]{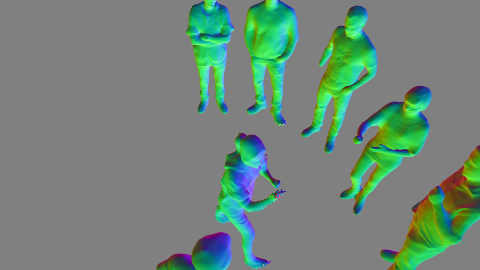} &
\includegraphics[width=2.5cm]{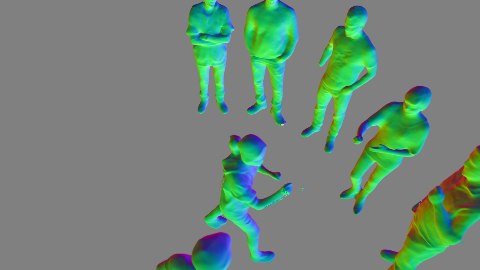} &
\includegraphics[width=2.5cm]{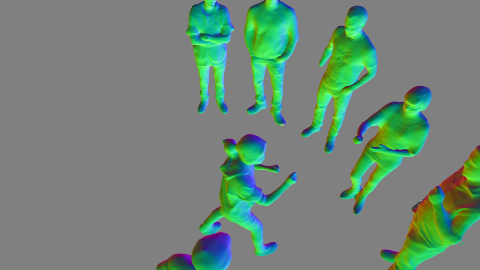} &
\includegraphics[width=2.5cm]{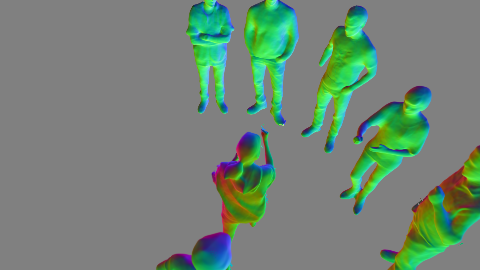} \\
\rotatebox{90}{\quad scale} 
\includegraphics[width=2.5cm]{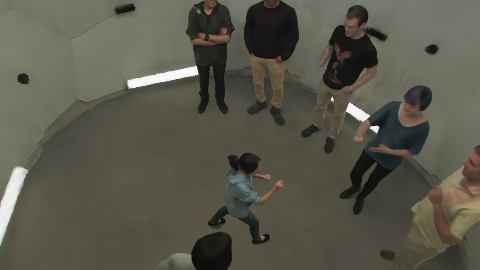} &
\includegraphics[width=2.5cm]{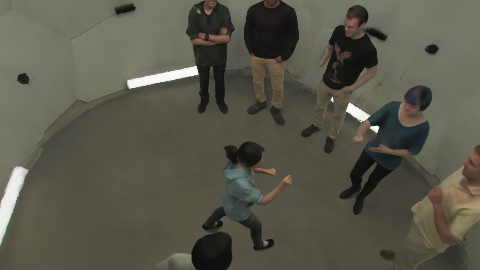} &
\includegraphics[width=2.5cm]{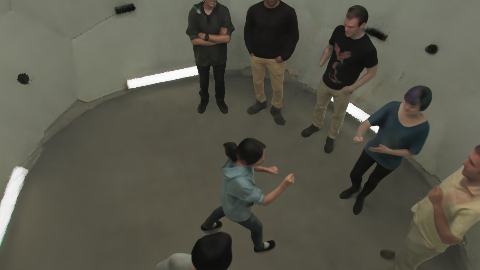} &
\includegraphics[width=2.5cm]{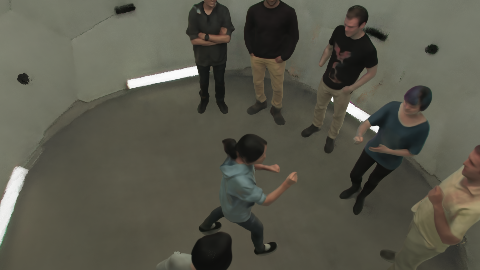} &
\includegraphics[width=2.5cm]{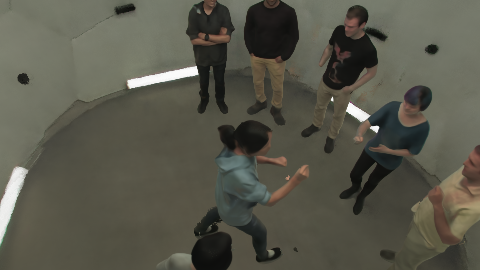} &
\includegraphics[width=2.5cm]{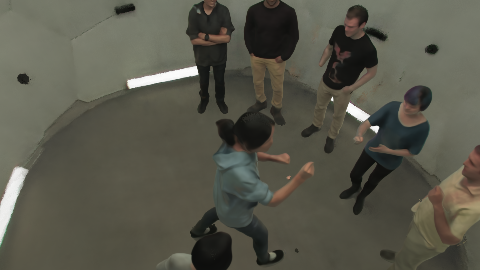} &
\includegraphics[width=2.5cm]{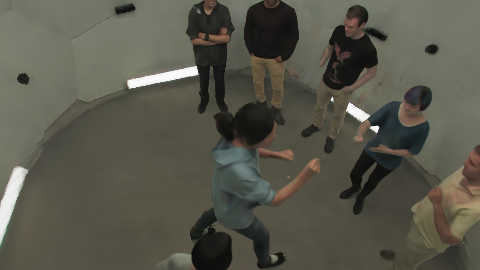} \\
\rotatebox{90}{\quad scale} 
\includegraphics[width=2.5cm]{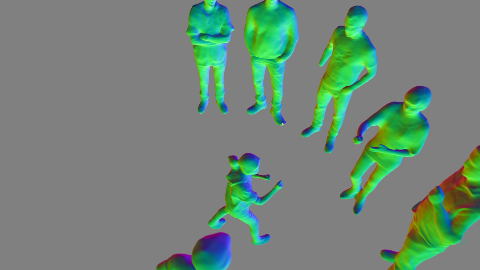} &
\includegraphics[width=2.5cm]{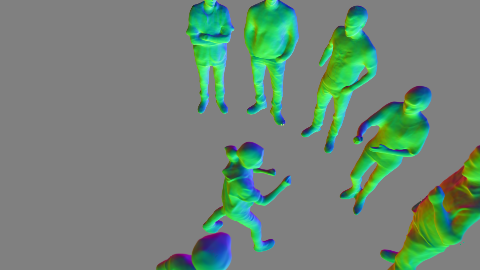} &
\includegraphics[width=2.5cm]{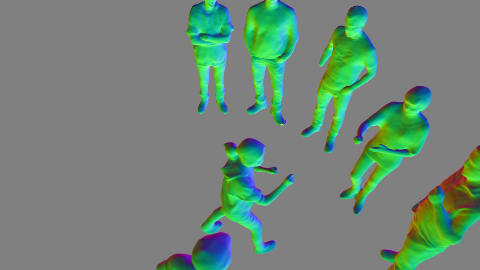} &
\includegraphics[width=2.5cm]{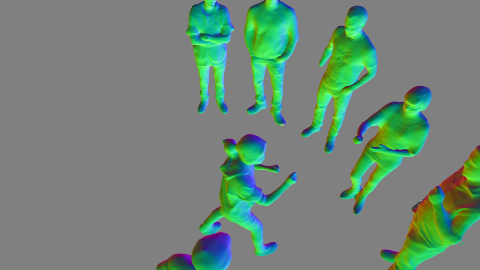} &
\includegraphics[width=2.5cm]{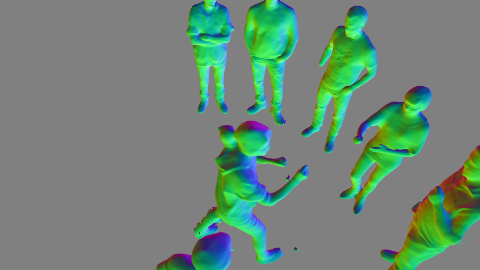} &
\includegraphics[width=2.5cm]{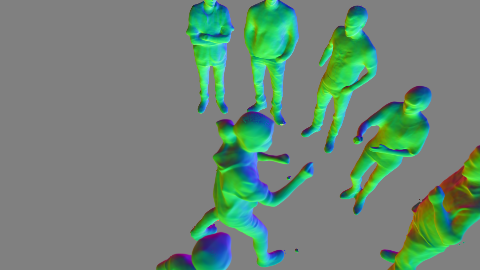} &
\includegraphics[width=2.5cm]{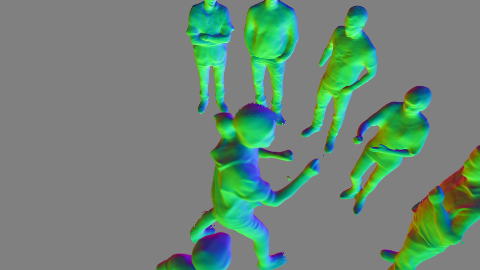} \\
\end{tabular}}
 \caption{Qualitative results for the editing application. We show synthesised novel views and reconstructed normal images of multiple humans when (1) removing, (2) translating, (3) rotating and (4) scaling subjects in the scene. } 
\label{fig:edit} 
\end{figure*}

\section{Additional Results} 
\paragraph{Scene editing.}
We show here how our method can be used to perform post-learning scene editing without any additional training.
Thanks to the human bounding-box-based modeling of the foreground scene, it is straightforward to rigidly transform or omit each person by simply applying, before rendering, the corresponding manipulation to the points sampled inside the defined bounding box. 
Figure~\ref{fig:edit} shows qualitative results of such application, trained on scene $\#5$ from the CMU Panoptic dataset \cite{Simon_2017_CVPR,Joo_2017_TPAMI} using 20 training views. We can see here that our approach can generate realistic new scenes as well as plausible inpaintings of the missing regions.  

\paragraph{Comparisons with varying number of people.} In Fig.~\ref{fig:cmu} we provide additional qualitative comparisons against NeuS~\cite{wang2021neus} and VolSDF~\cite{yariv2021volume}, where we show results on the CMU Panoptic dataset \cite{Simon_2017_CVPR,Joo_2017_TPAMI} with varying number of people in the scene  (Going from 3 to 7 people). Note here how increasing the number of people reduces the quality of our baselines results, \ie mixing the background with humans or generating noisy geometries. Meanwhile, our method performs consistently, independently of the number of people.


\begin{figure*}[ht]
\flushleft 
\def\tabularxcolumn#1{m{#1}}
\setlength{\tabcolsep}{0pt}
\renewcommand{\arraystretch}{0} 
\scalebox{1}{
\begin{tabular}{c cccc ccc}
\rotatebox{90}{\quad 3} 
\includegraphics[width=2.5cm]{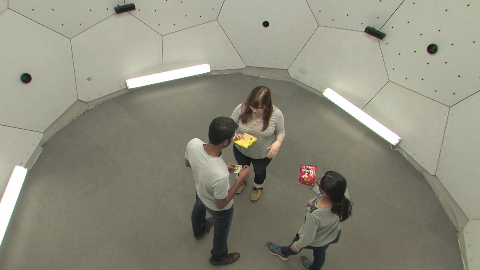} &
\includegraphics[width=2.5cm]{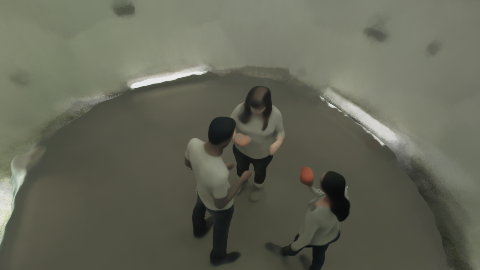} &
\includegraphics[width=2.5cm]{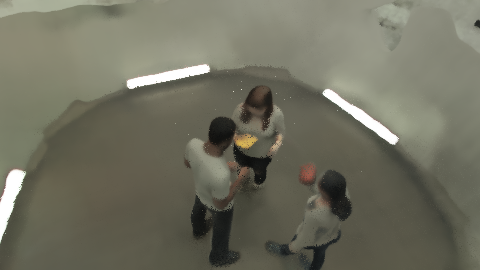} &
\includegraphics[width=2.5cm]{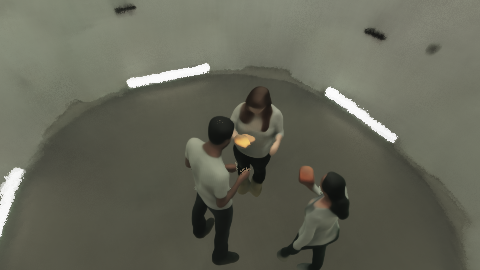} &
\includegraphics[width=2.5cm]{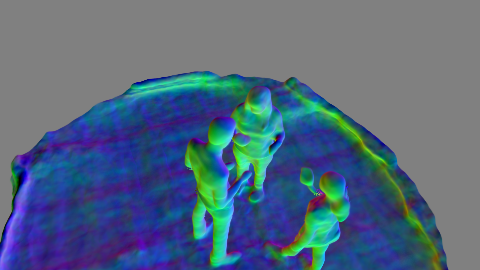} &
\includegraphics[width=2.5cm]{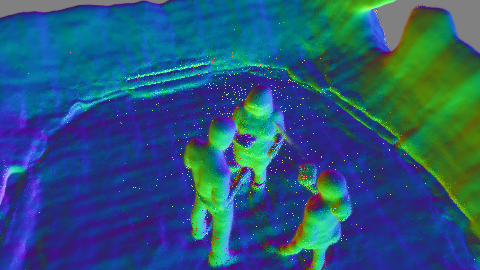} &
\includegraphics[width=2.5cm]{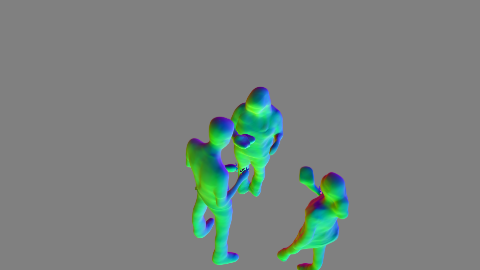} \\
\rotatebox{90}{\quad 4}
\includegraphics[width=2.5cm]{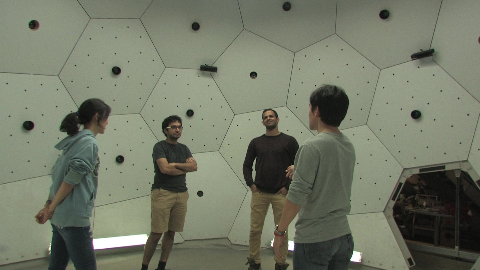} &
\includegraphics[width=2.5cm]{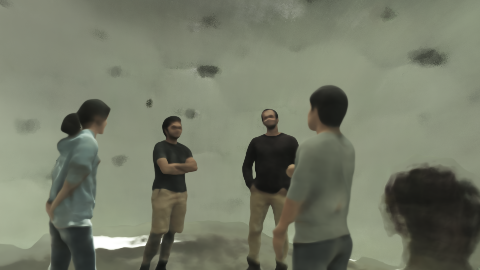} &
\includegraphics[width=2.5cm]{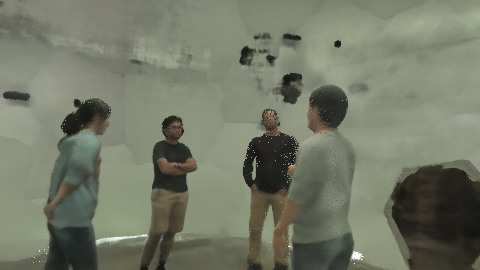} &
\includegraphics[width=2.5cm]{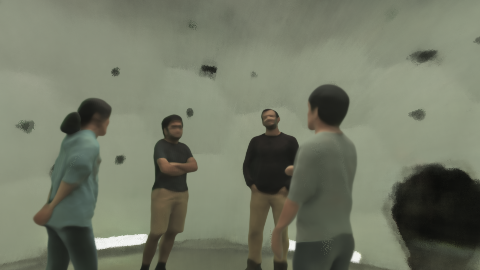} &
\includegraphics[width=2.5cm]{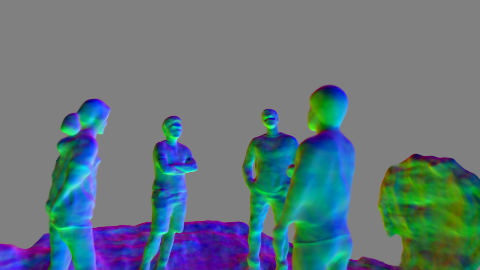} &
\includegraphics[width=2.5cm]{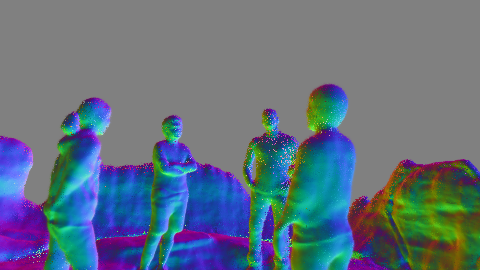} &
\includegraphics[width=2.5cm]{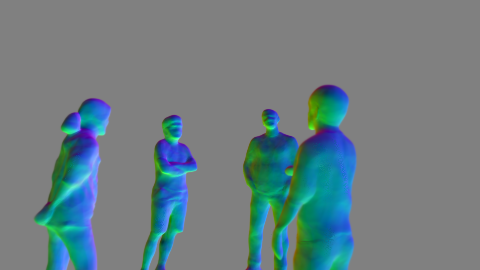} \\
\rotatebox{90}{\quad 5}
\includegraphics[width=2.5cm]{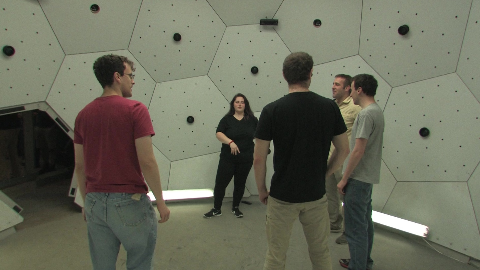} &
\includegraphics[width=2.5cm]{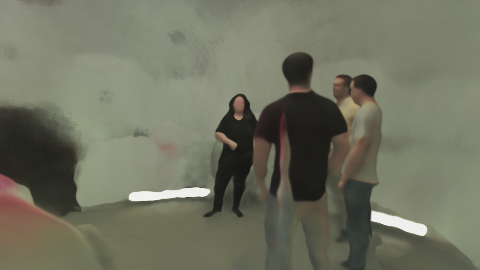} &
\includegraphics[width=2.5cm]{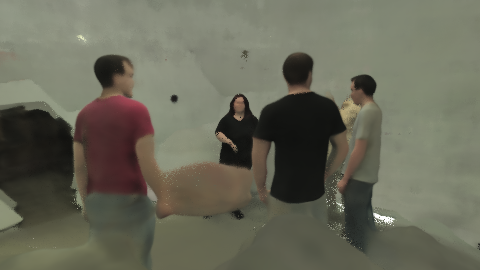} &
\includegraphics[width=2.5cm]{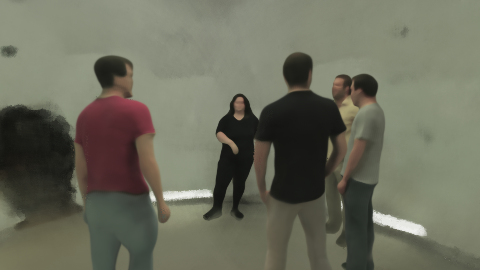} &
\includegraphics[width=2.5cm]{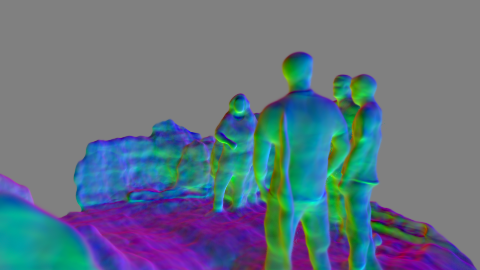} &
\includegraphics[width=2.5cm]{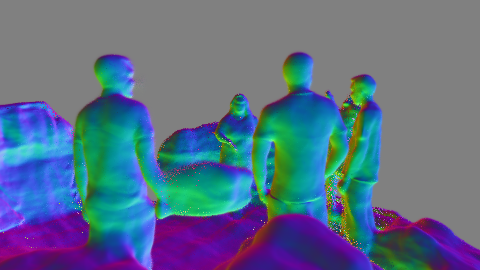} &
\includegraphics[width=2.5cm]{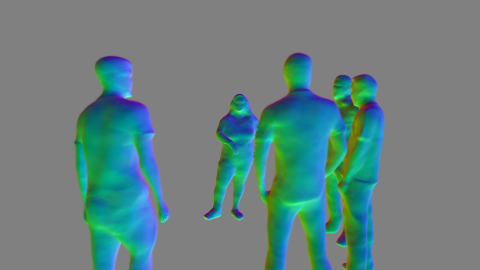} \\
\rotatebox{90}{\quad 6}
\includegraphics[width=2.5cm]{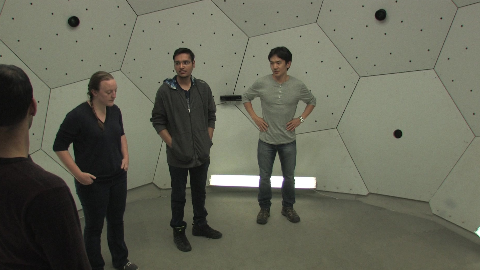} &
\includegraphics[width=2.5cm]{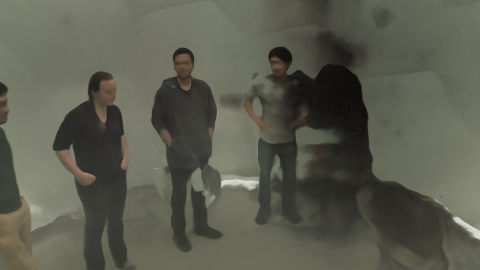} &
\includegraphics[width=2.5cm]{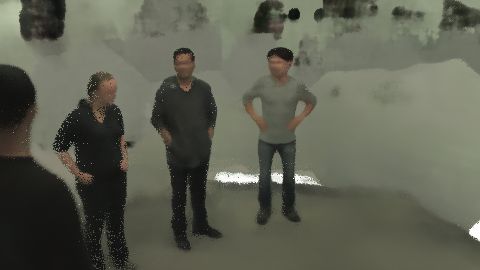} &
\includegraphics[width=2.5cm]{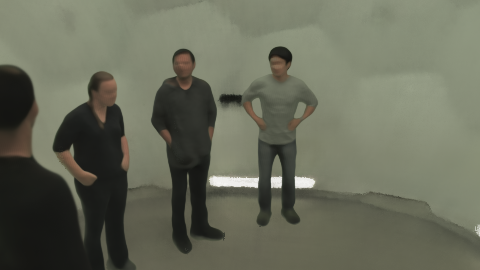} &
\includegraphics[width=2.5cm]{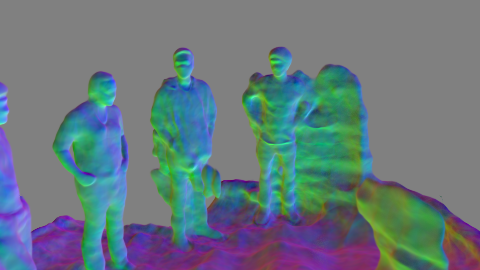} &
\includegraphics[width=2.5cm]{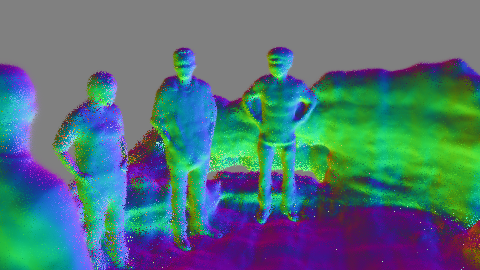} &
\includegraphics[width=2.5cm]{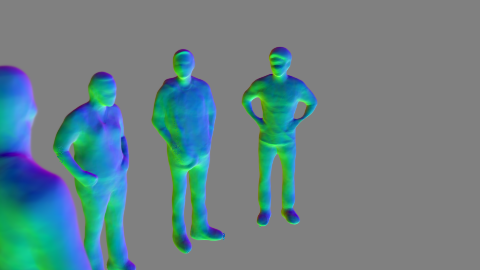} \\
\rotatebox{90}{\quad 7}
\includegraphics[width=2.5cm]{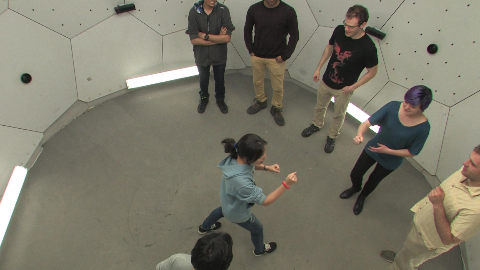} &
\includegraphics[width=2.5cm]{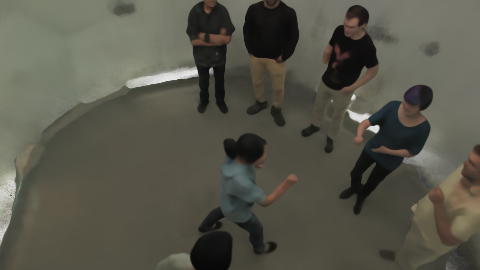} &
\includegraphics[width=2.5cm]{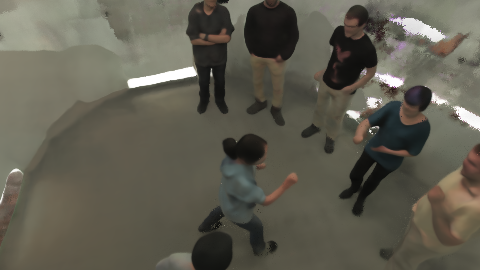} &
\includegraphics[width=2.5cm]{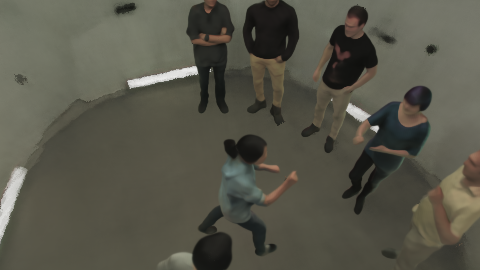} &
\includegraphics[width=2.5cm]{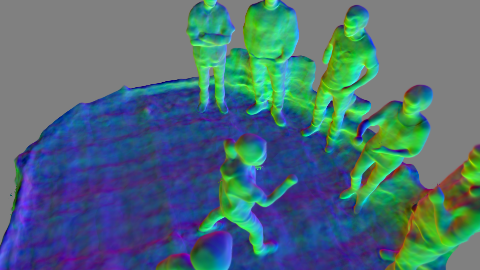} &
\includegraphics[width=2.5cm]{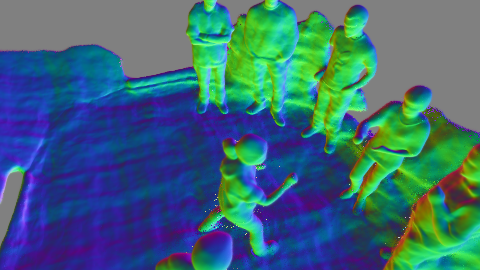} &
\includegraphics[width=2.5cm]{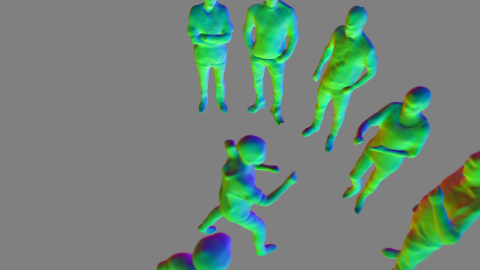} \\
Ground Truth &NeuS &VolSDF & Ours  & Neus &Volsdf & Ours\\
\end{tabular}}
 \caption{Qualitative comparisons against NeuS~\cite{wang2021neus} and VolSDF~\cite{yariv2021volume} for different number of people in the scene. Showing synthesised novel views and reconstructed normal images on 5 scenes from CMU Panoptic dataset \cite{Simon_2017_CVPR,Joo_2017_TPAMI}, using 20 training views.} 
\label{fig:cmu} 
\end{figure*}

\paragraph{Additional Quantitative Results.}
Table~\ref{tab:chamfer} provides a full Chamfer distance comparison in the synthetic data setup as an addition to the results reported in Table 3 of the main submission. Symbol `$-$' represents cases where the baselines fail to reconstruct a meaningful geometry, and hence the error is too large. To favor the baselines NeuS~\cite{wang2021neus} and VolSDF~\cite{yariv2021volume} in the main submission, we computed the uni-directional Chamfer distance from ground-truth to source, as the baselines reconstructed the ground of the scene in addition to the people. For a more standard evaluation, we additionally show here the bi-directional Chamfer distance after removing the floor for the competing methods. 


\begin{table}[h!]
\begin{center}
\scalebox{0.78}{
\begin{tabular}{l|l|ccc | ccc}
\toprule[1.2pt]
\# People& Method &\multicolumn{3}{c|}{one-way Chamfer ↓} &  \multicolumn{3}{c}{bidirectional Chamfer ↓}\\
 & &5  &10 &15    &5 &10 &15   \\
\hline
 &NeuS& -&-&0.308&-&-& 3.026\\
 1&VolSDF &-&0.020&0.019&-& \bf{0.039} & 0.167 \\
& Ours &0.025 &\bf{0.019} & \bf{0.018} & 0.271&0.211& \bf{0.154}\\
\hline
 &NeuS & -&-&0.321&-&-& 3.044\\
 5&VolSDF &-&-&0.151&-&-&1.478\\
& Ours&0.025&0.023 & \bf{0.020} & 0.391 & 0.289 & \bf{0.138} \\
\hline
 &NeuS &-&-&0.383&-&-&4.639 \\
10&VolSDF &-&-&0.248&-&-&1.579\\
&Ours & 0.082 & 0.063 & \bf{0.043} & 0.111 & 0.085 & \bf{0.081}\\
 \bottomrule[1.2pt]
\end{tabular}}
\caption{Geometry reconstruction error under varying number of people, compared to NeuS~\cite{wang2021neus} and VolSDF~\cite{yariv2021volume} using the synthetic dataset, with 5/10/15 views for training. Symbol `$-$' represents cases where the baselines fail to reconstruct a meaningful geometry.}
\label{tab:chamfer}
\end{center}
\end{table}

\begin{figure}[h]
\centering
\vspace{-3mm}
\setlength{\tabcolsep}{0pt}
\renewcommand{\arraystretch}{0} 
\begin{tabular}{ccccc}
\includegraphics[width=1.68cm]{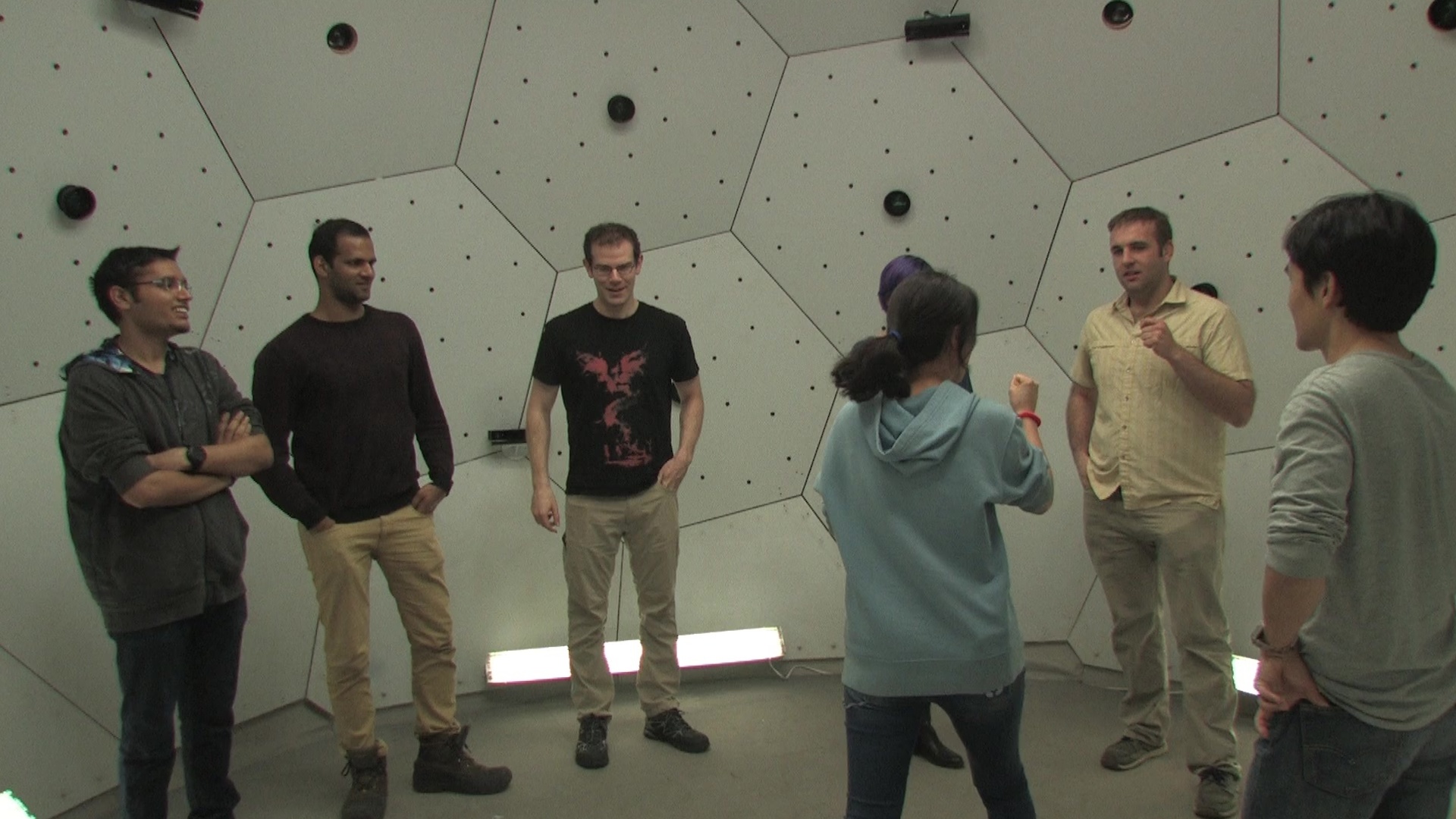} &
\includegraphics[width=1.68cm]{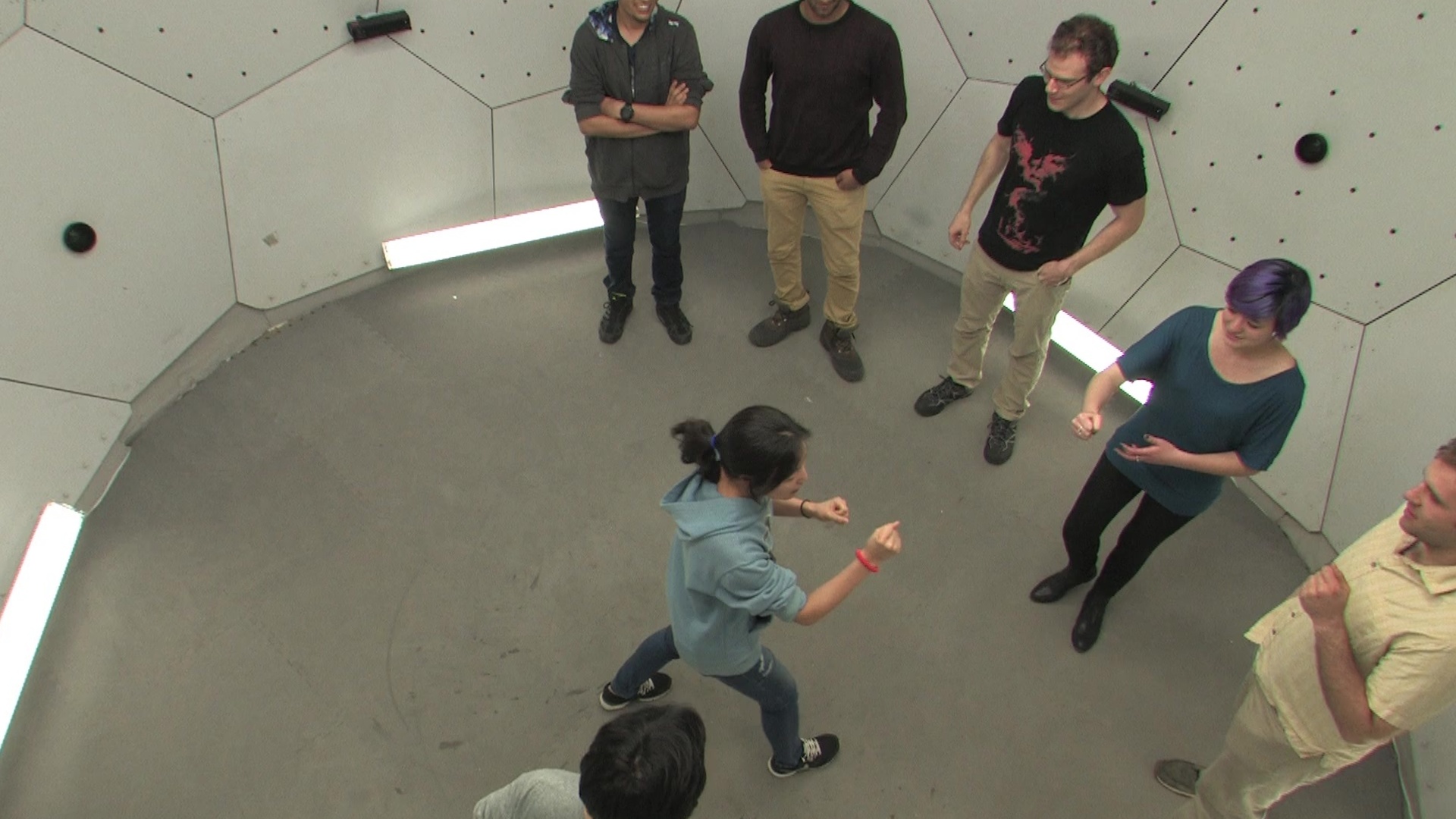} &
\includegraphics[width=1.68cm]{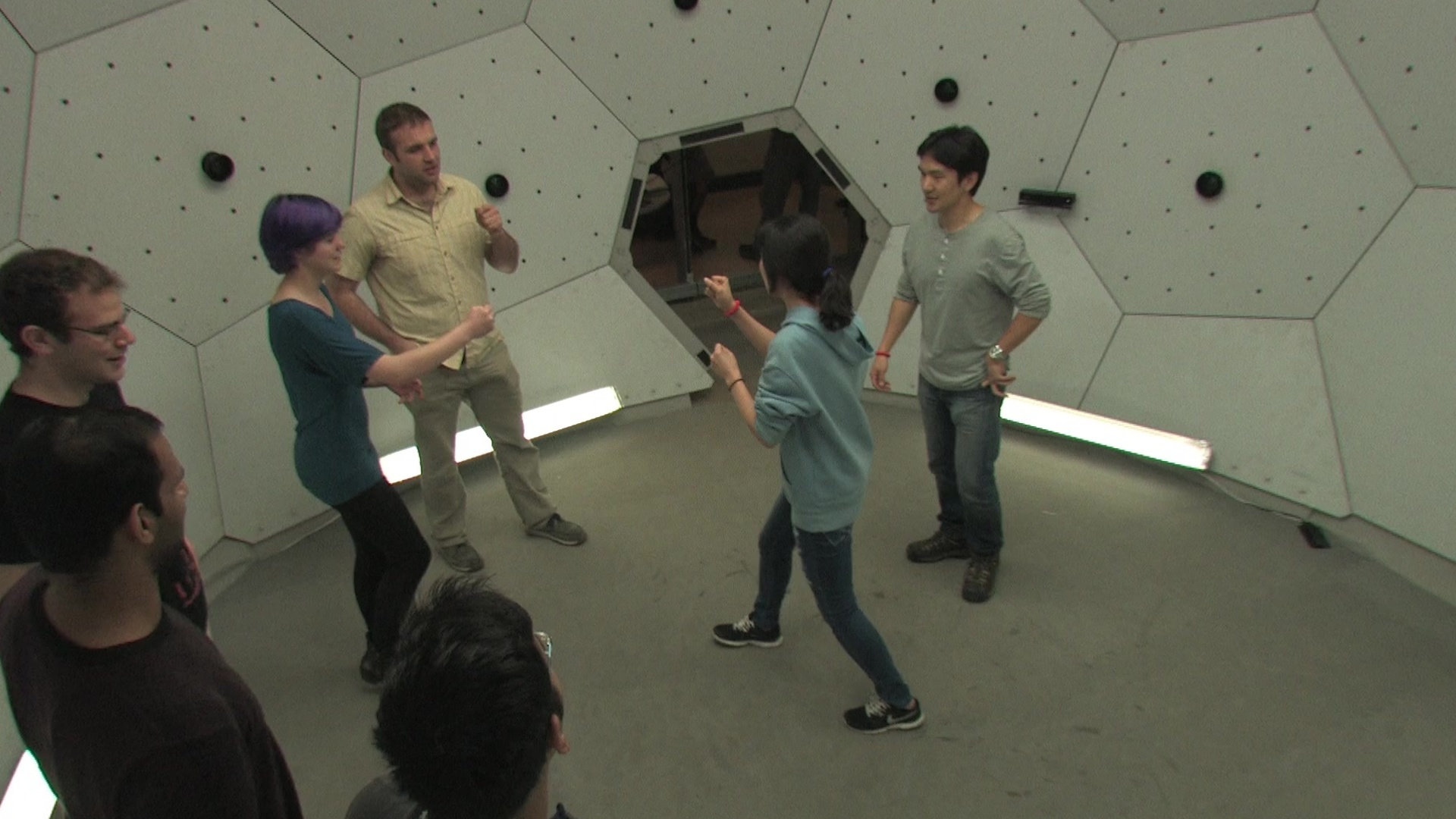} &
\includegraphics[width=1.68cm]{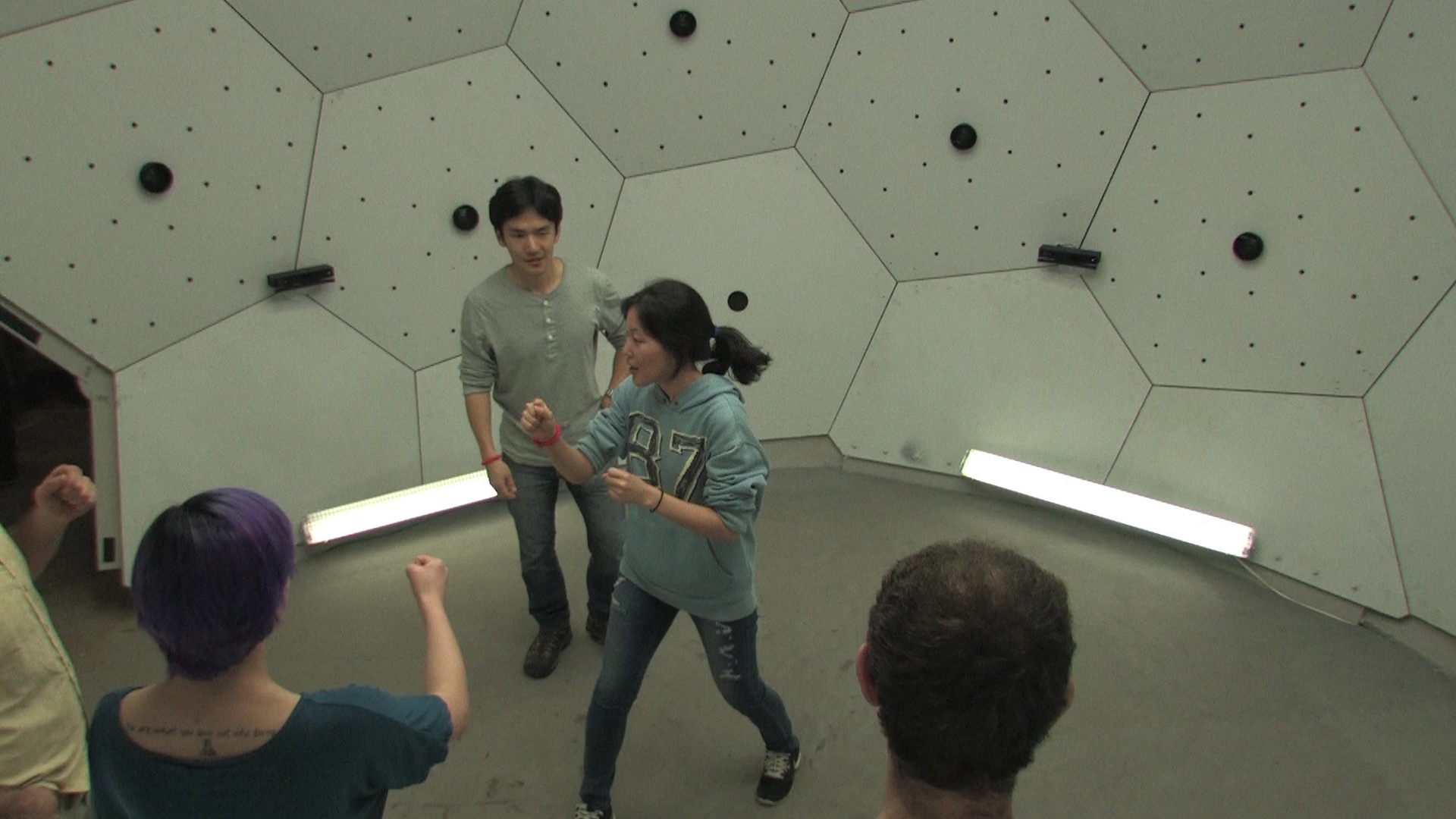} &
\includegraphics[width=1.68cm]{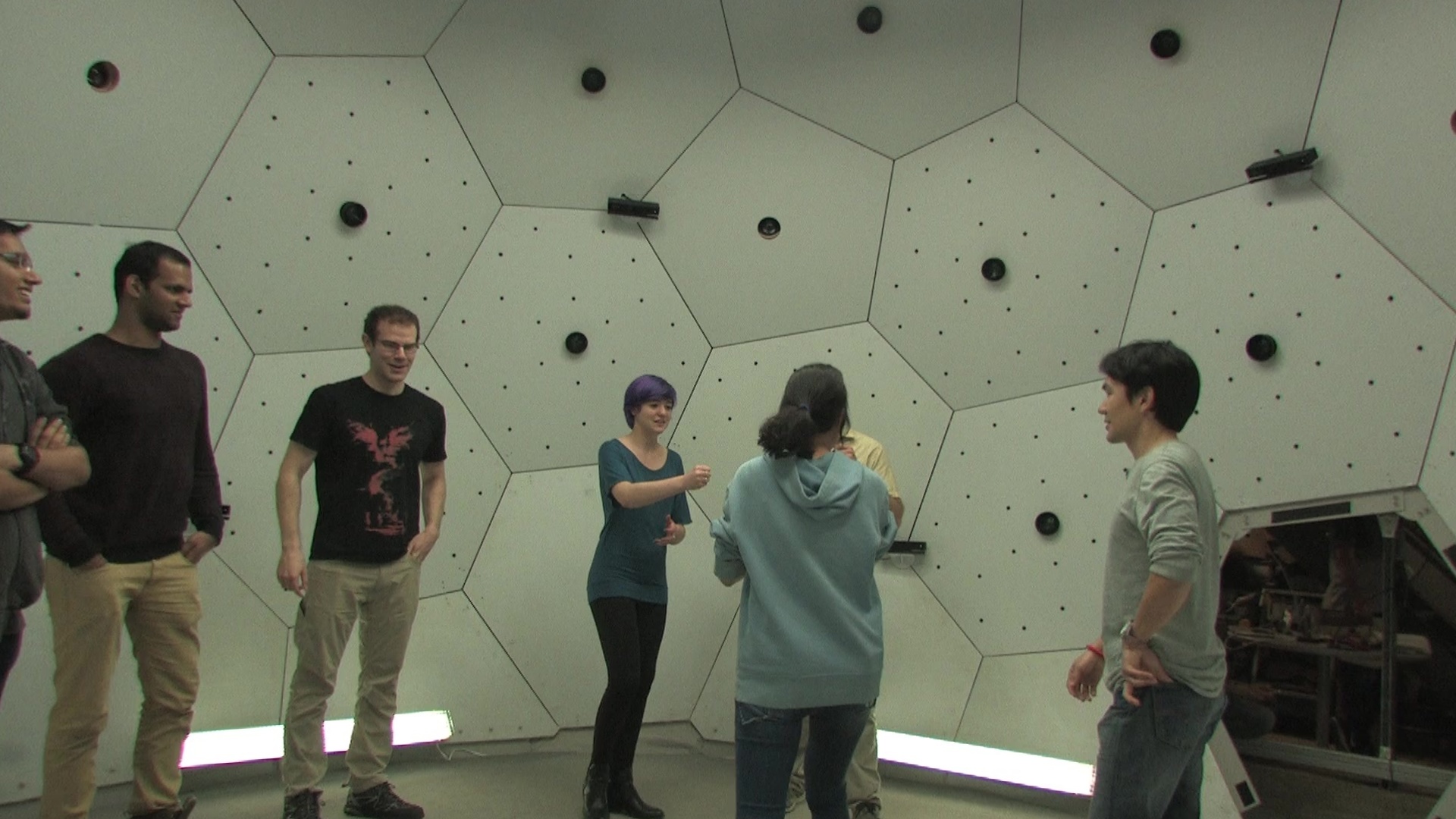} \\
\includegraphics[width=1.68cm]{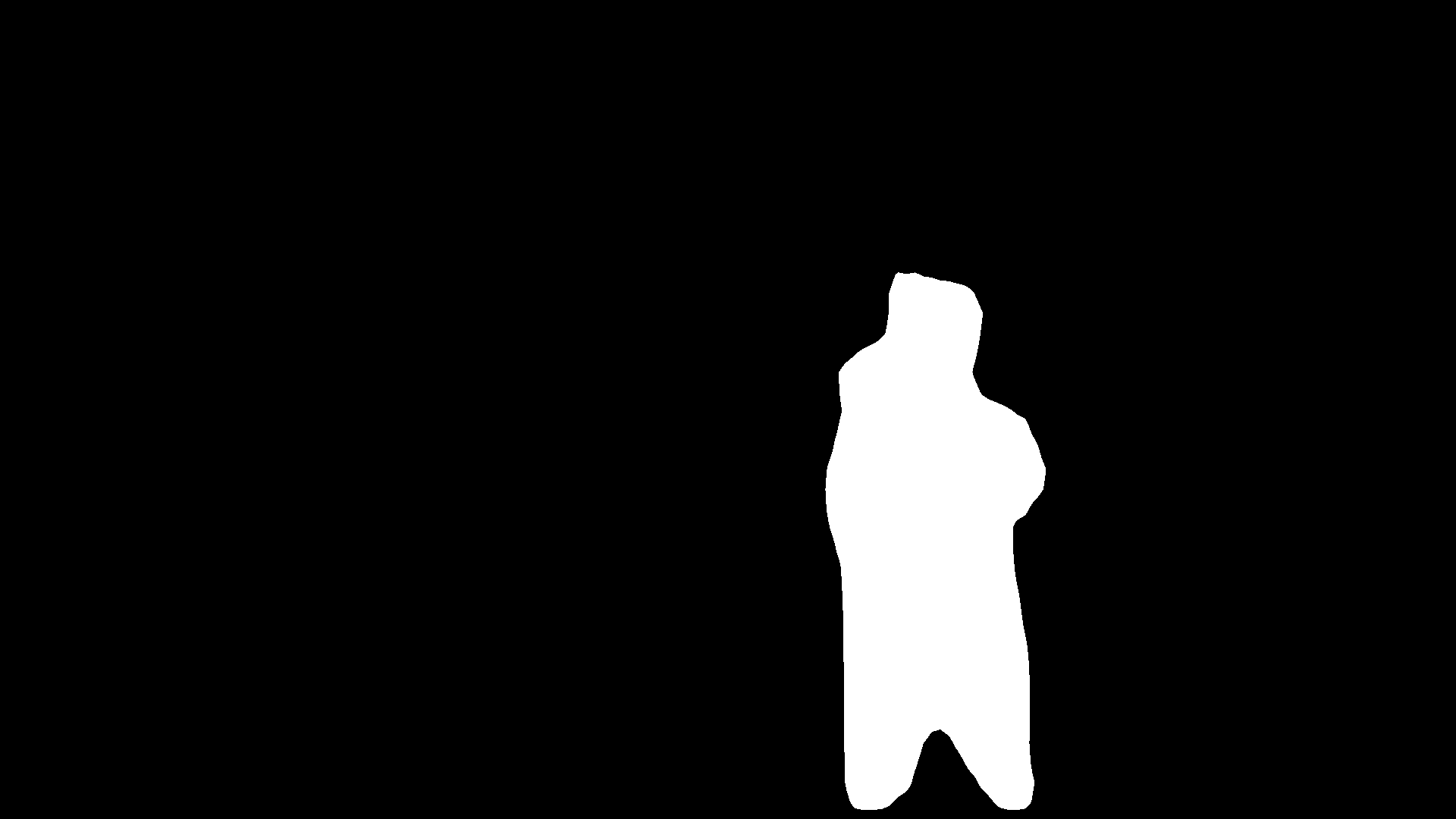} &
\includegraphics[width=1.68cm]{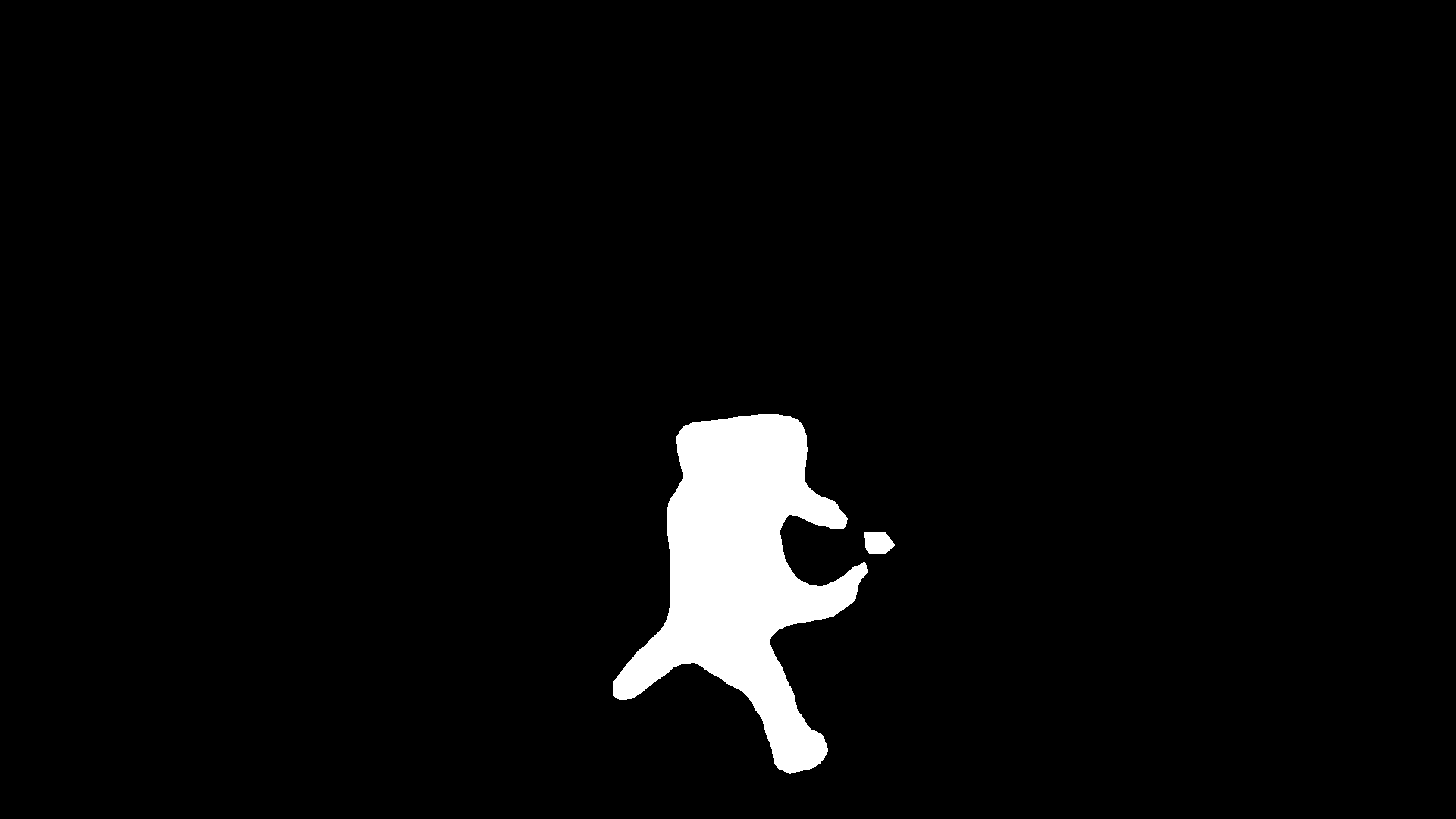} &
\includegraphics[width=1.68cm]{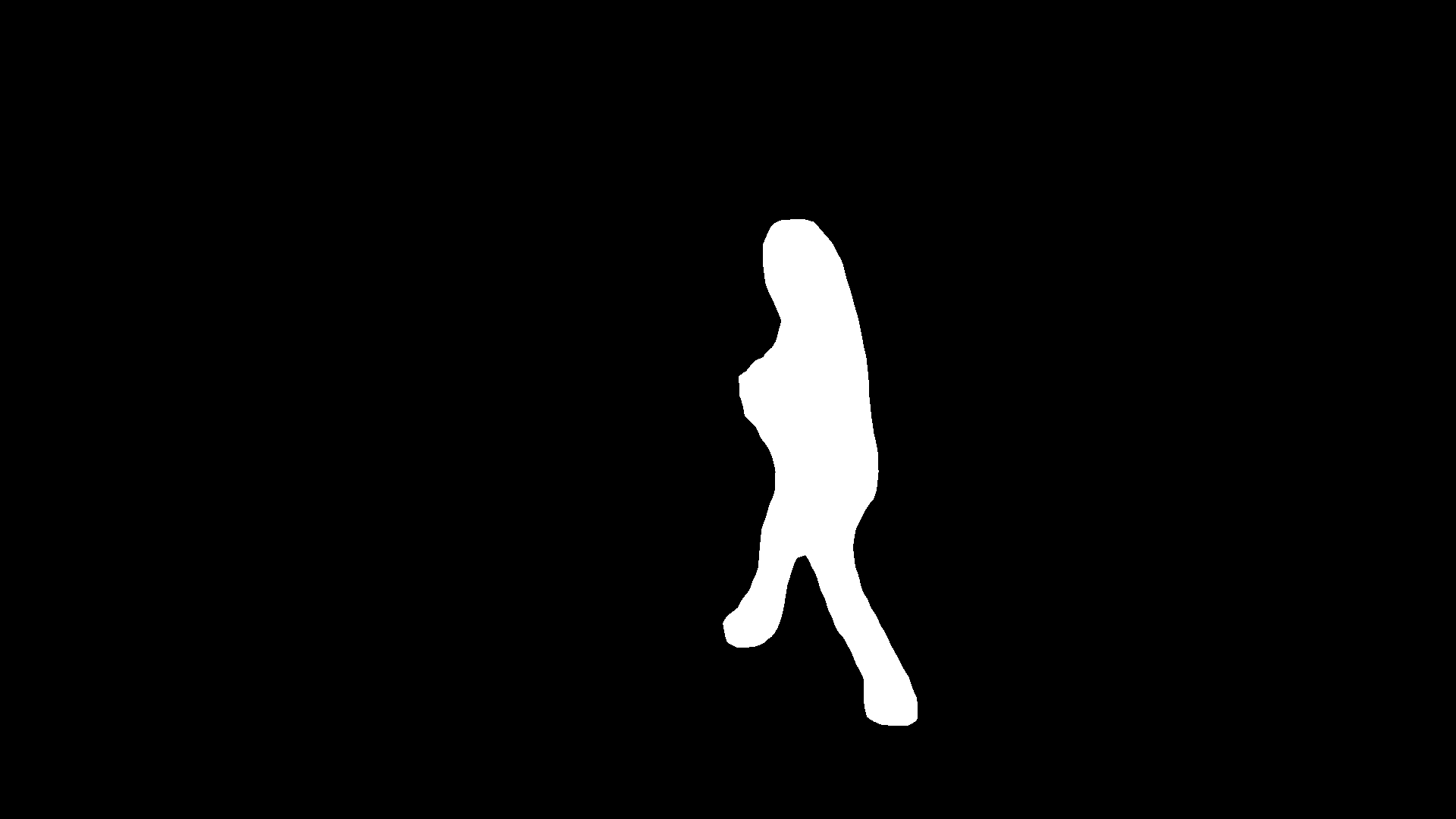} &
\includegraphics[width=1.68cm]{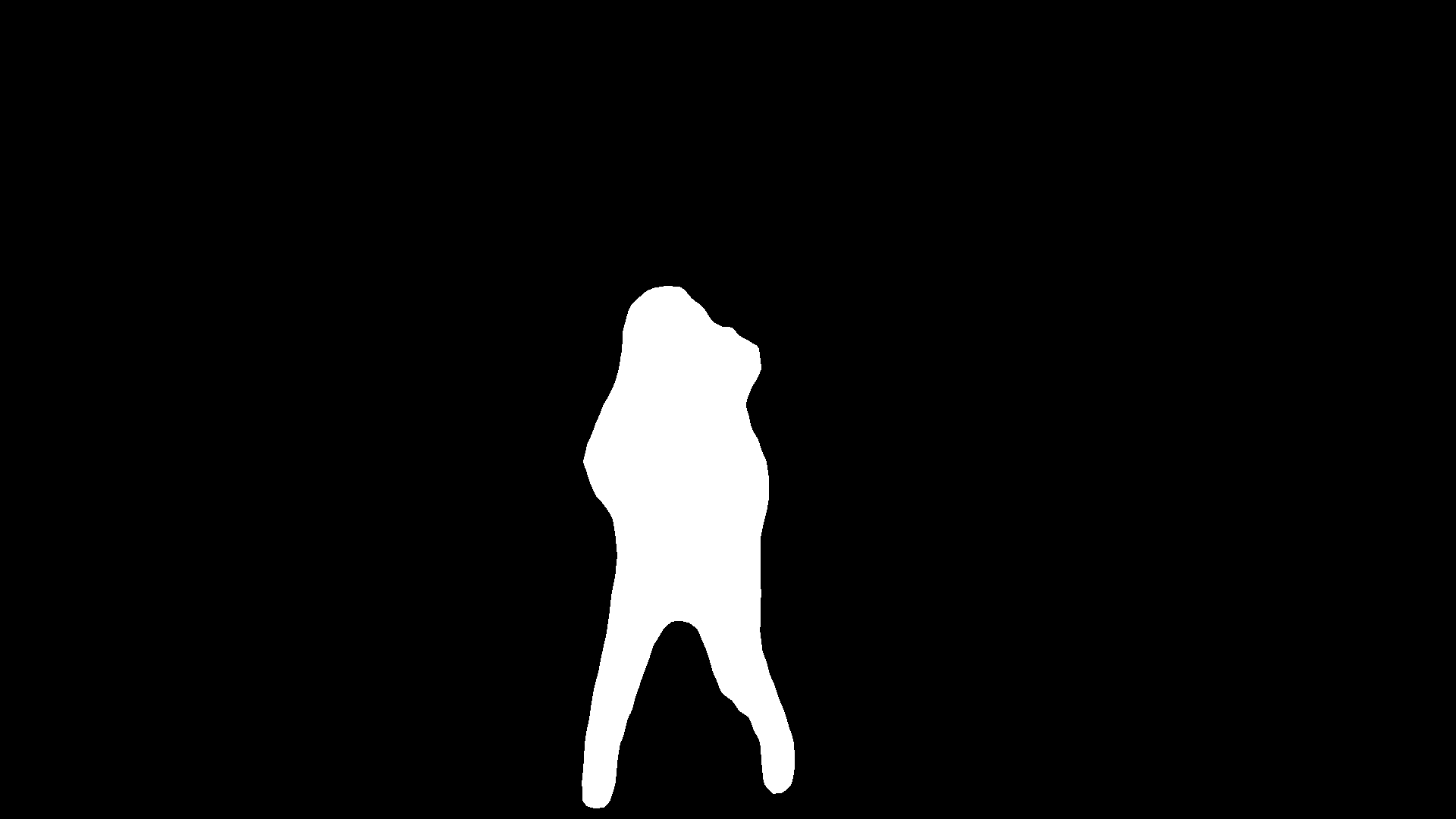} &
\includegraphics[width=1.68cm]{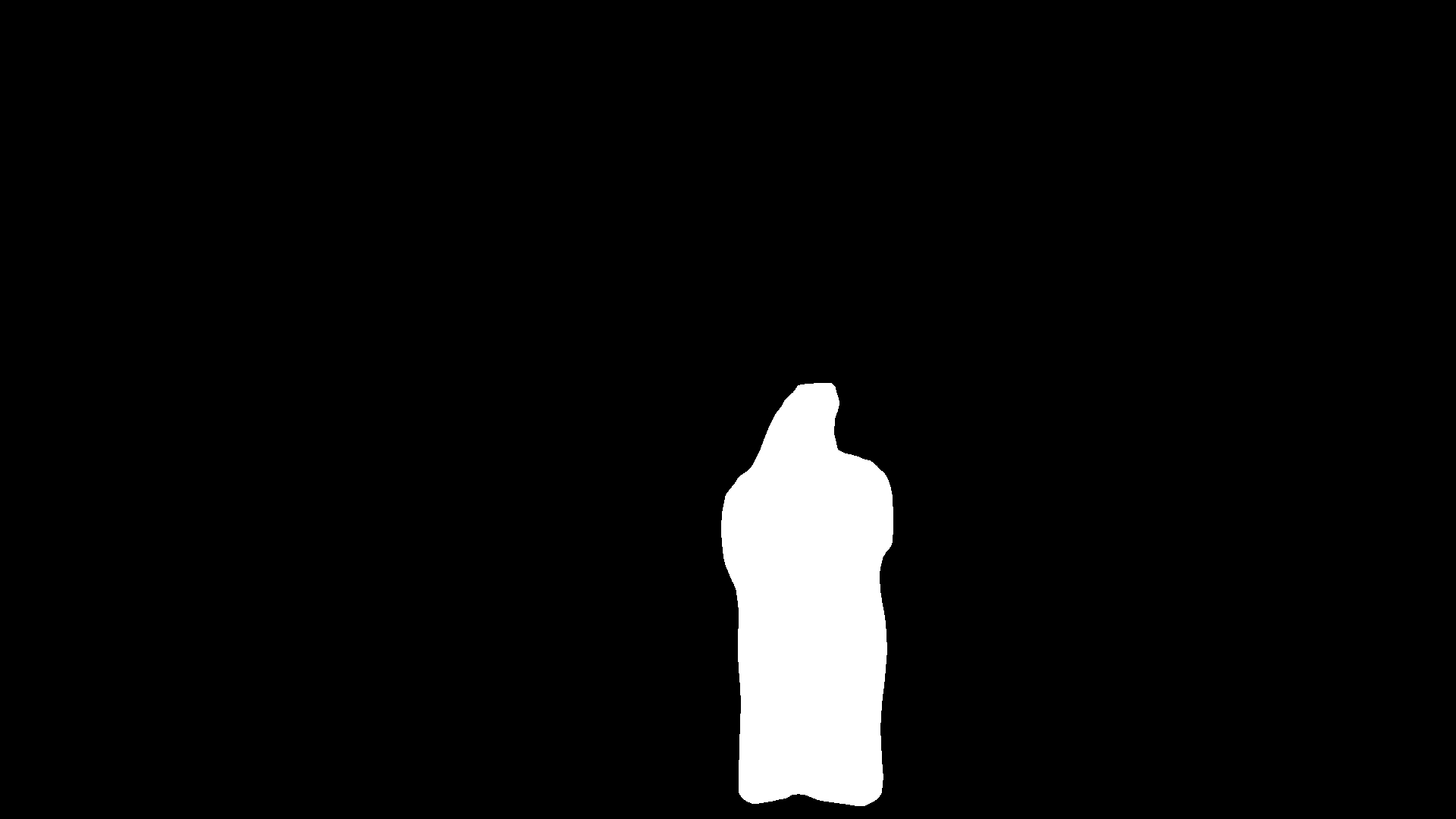} \\
\includegraphics[width=1.68cm]{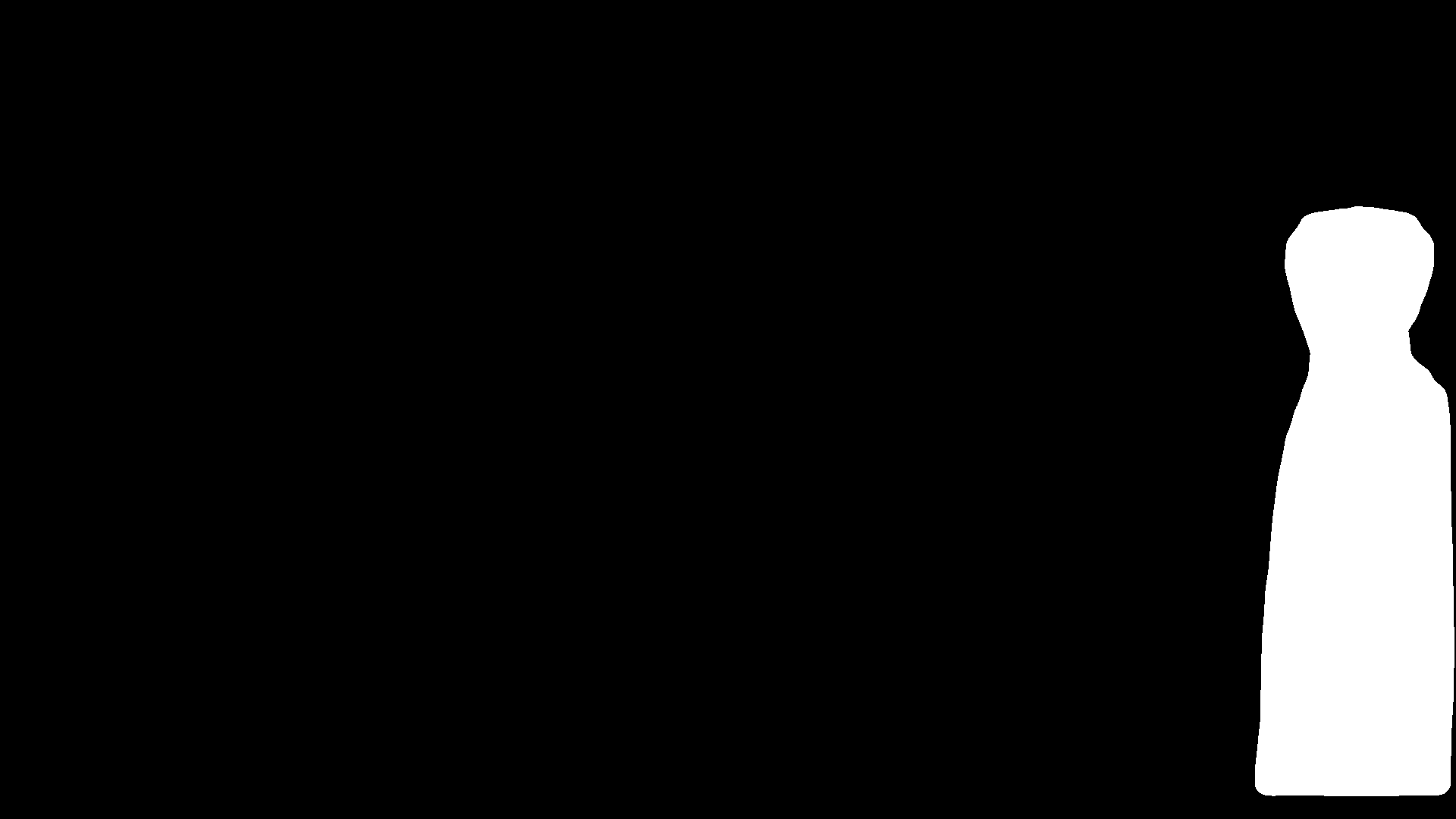} &
\includegraphics[width=1.68cm]{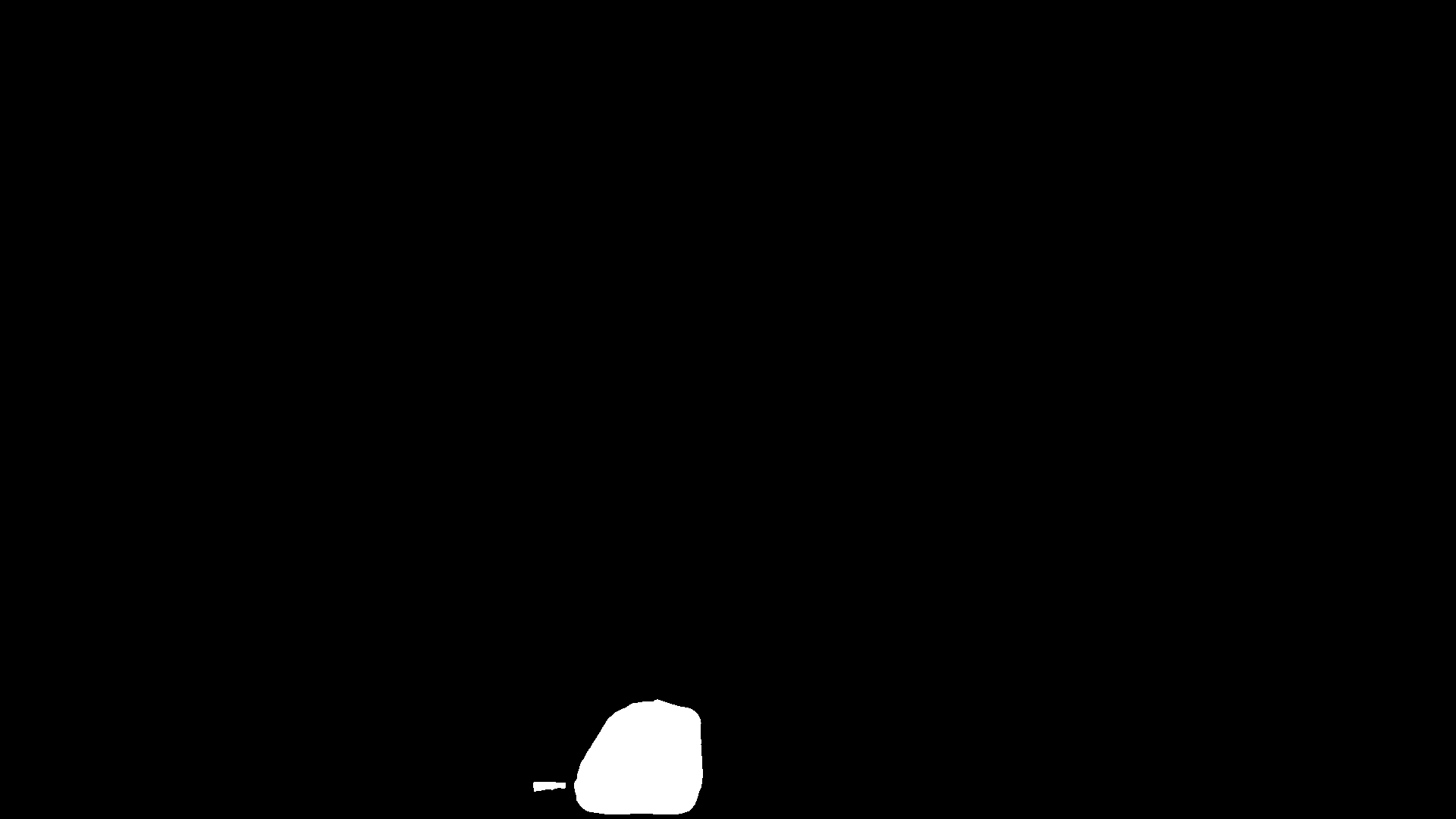} &
\includegraphics[width=1.68cm]{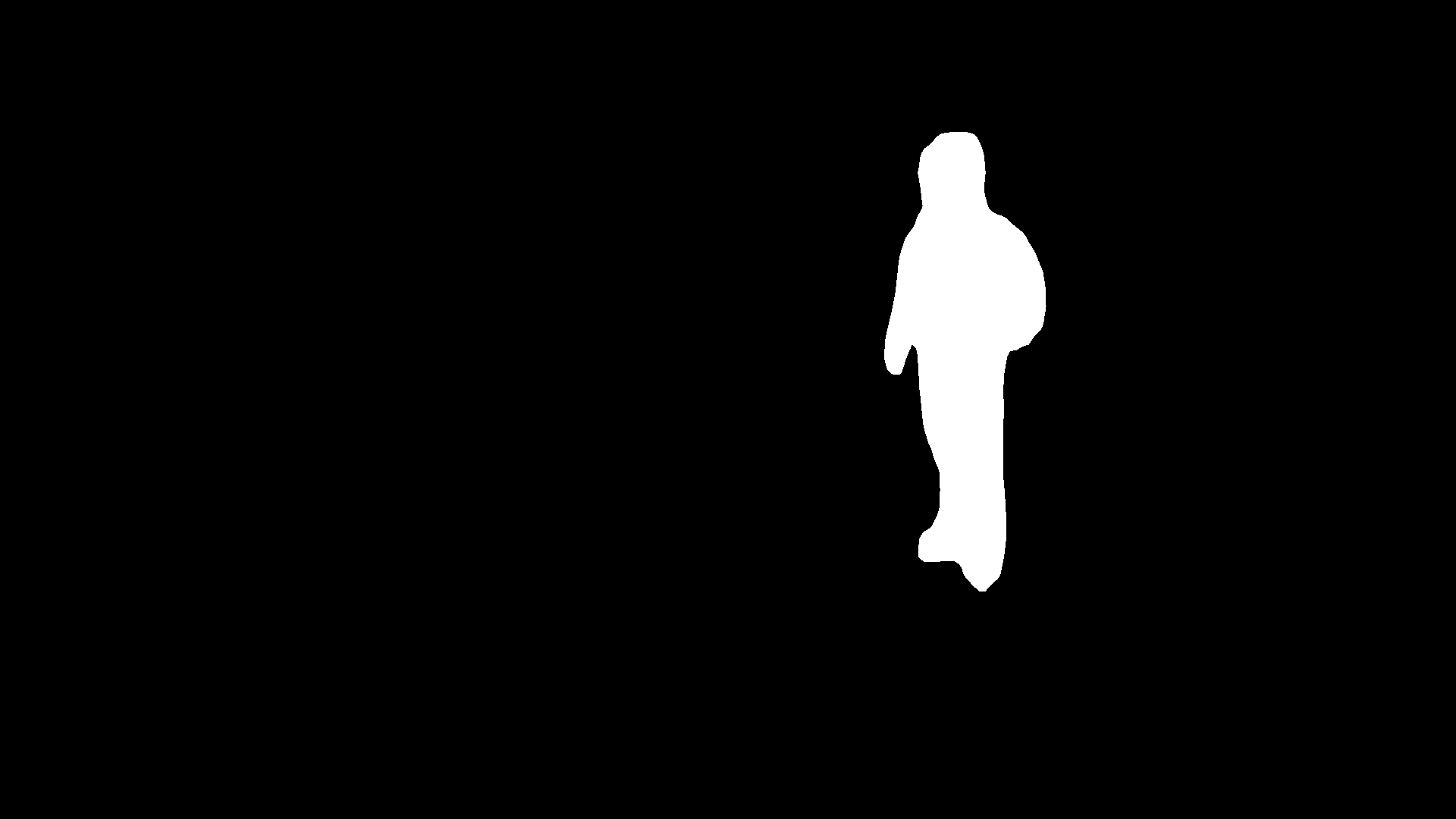} &
\includegraphics[width=1.68cm]{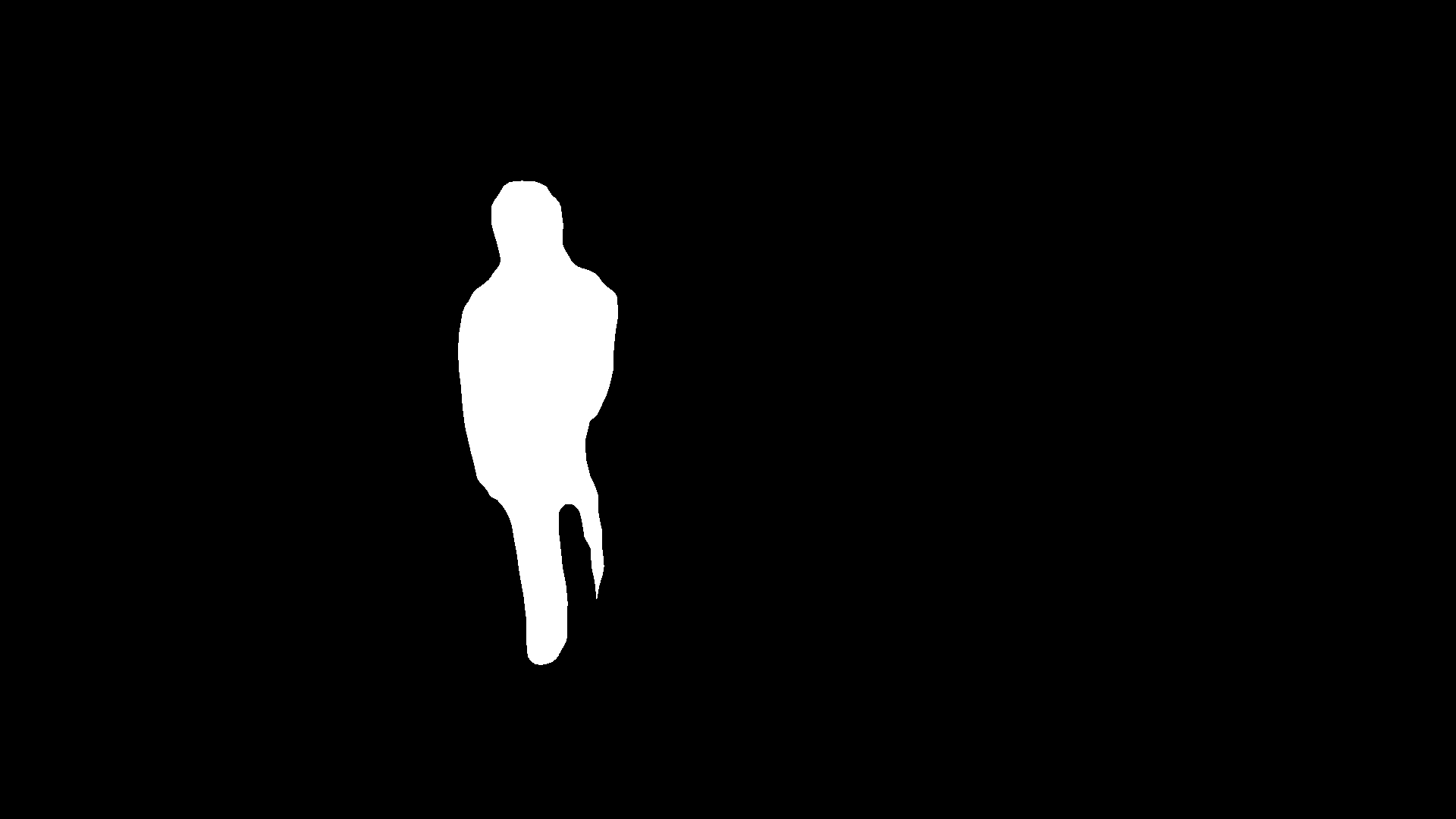} &
\includegraphics[width=1.68cm]{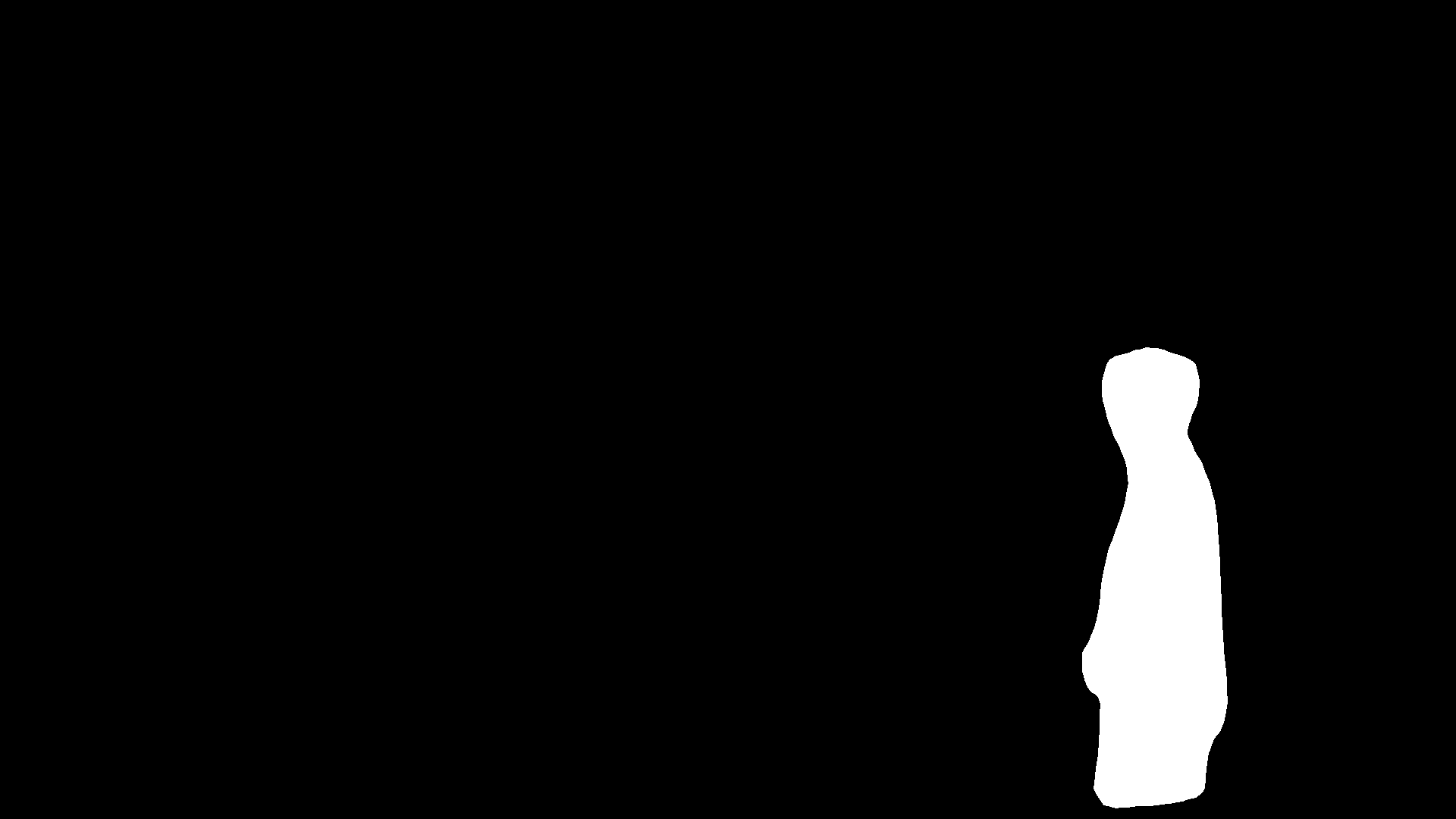} \\
\includegraphics[width=1.68cm]{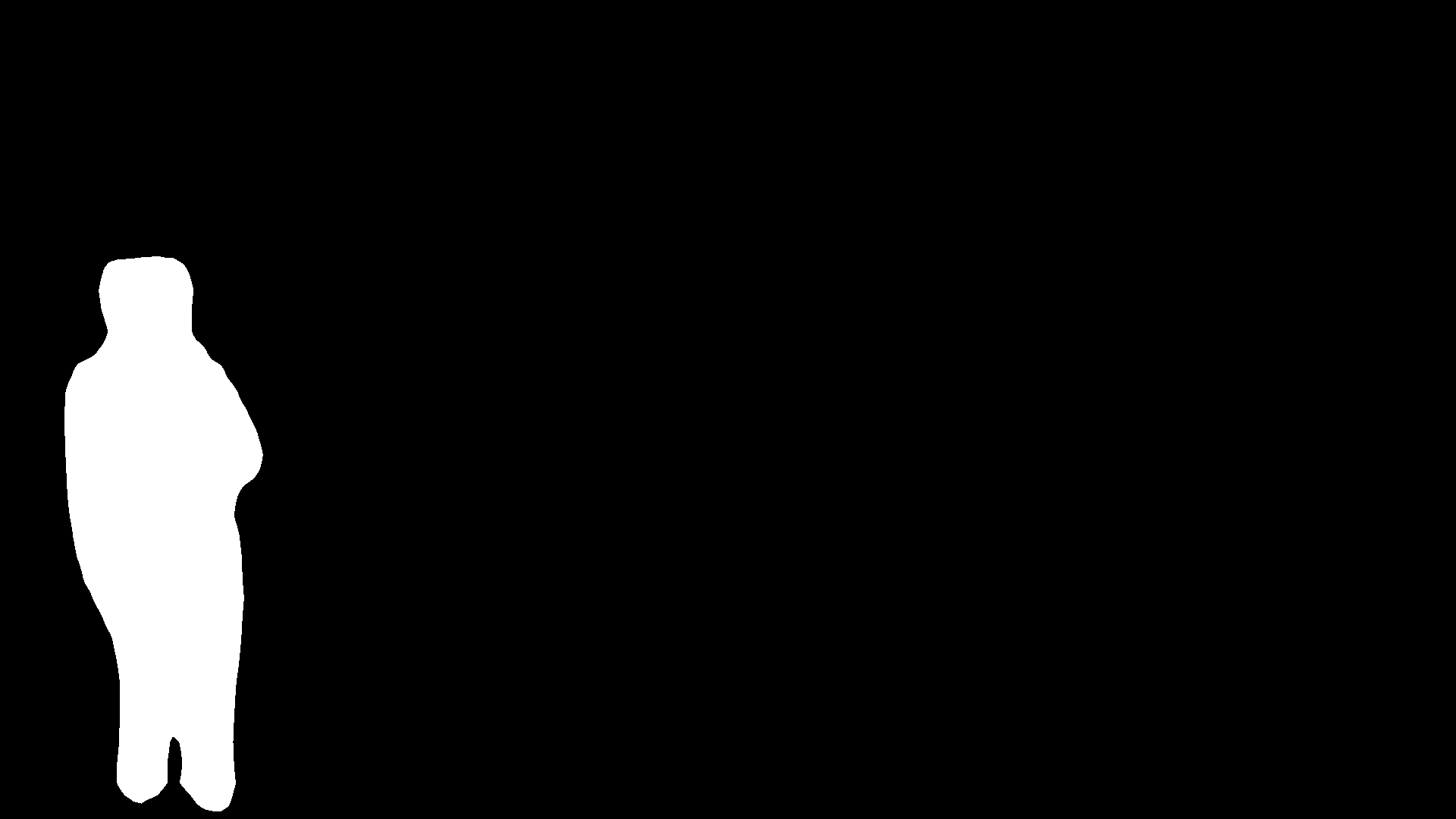} &
\includegraphics[width=1.68cm]{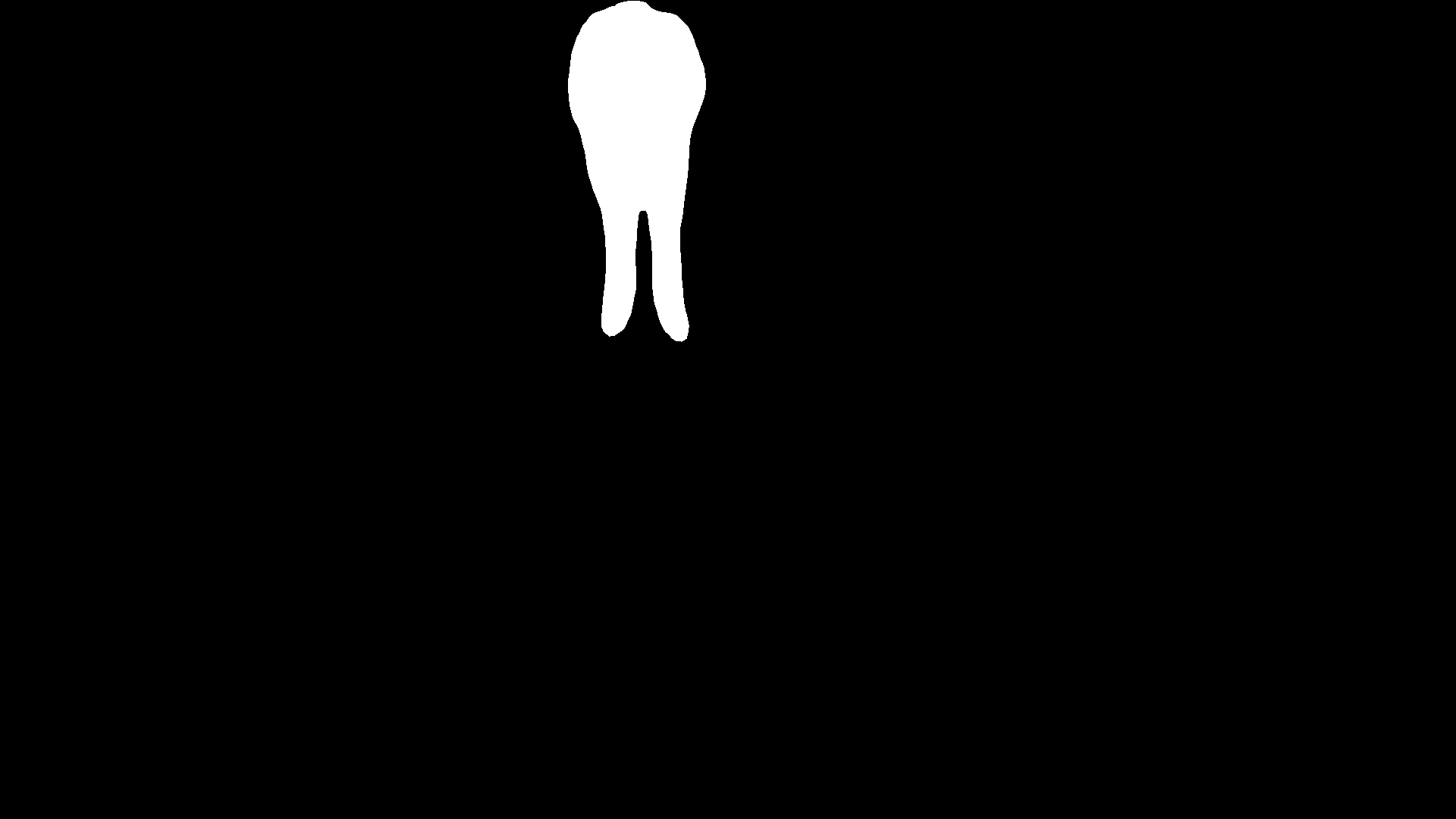} &
\includegraphics[width=1.68cm]{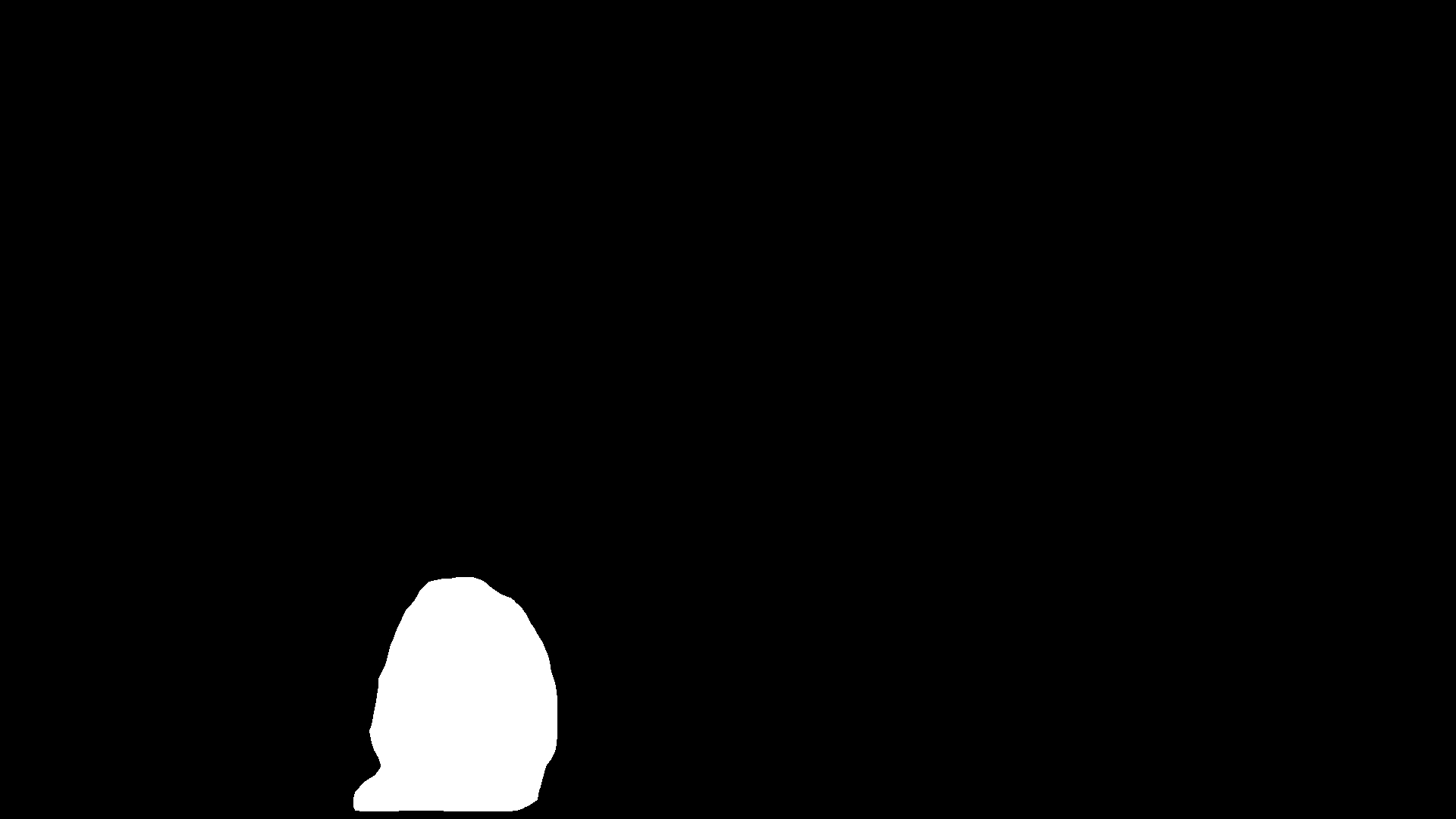} &
\includegraphics[width=1.68cm]{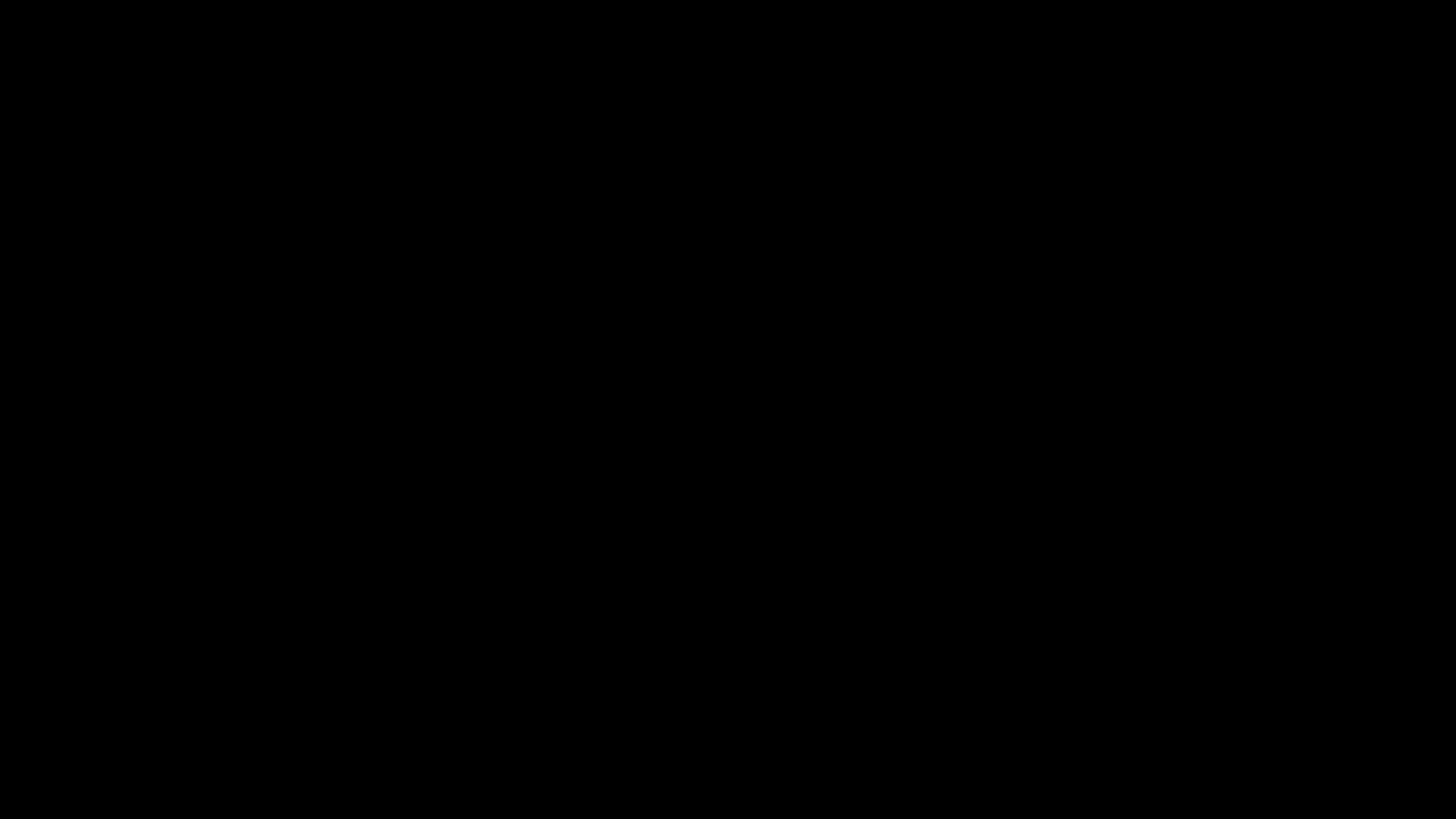} &
\includegraphics[width=1.68cm]{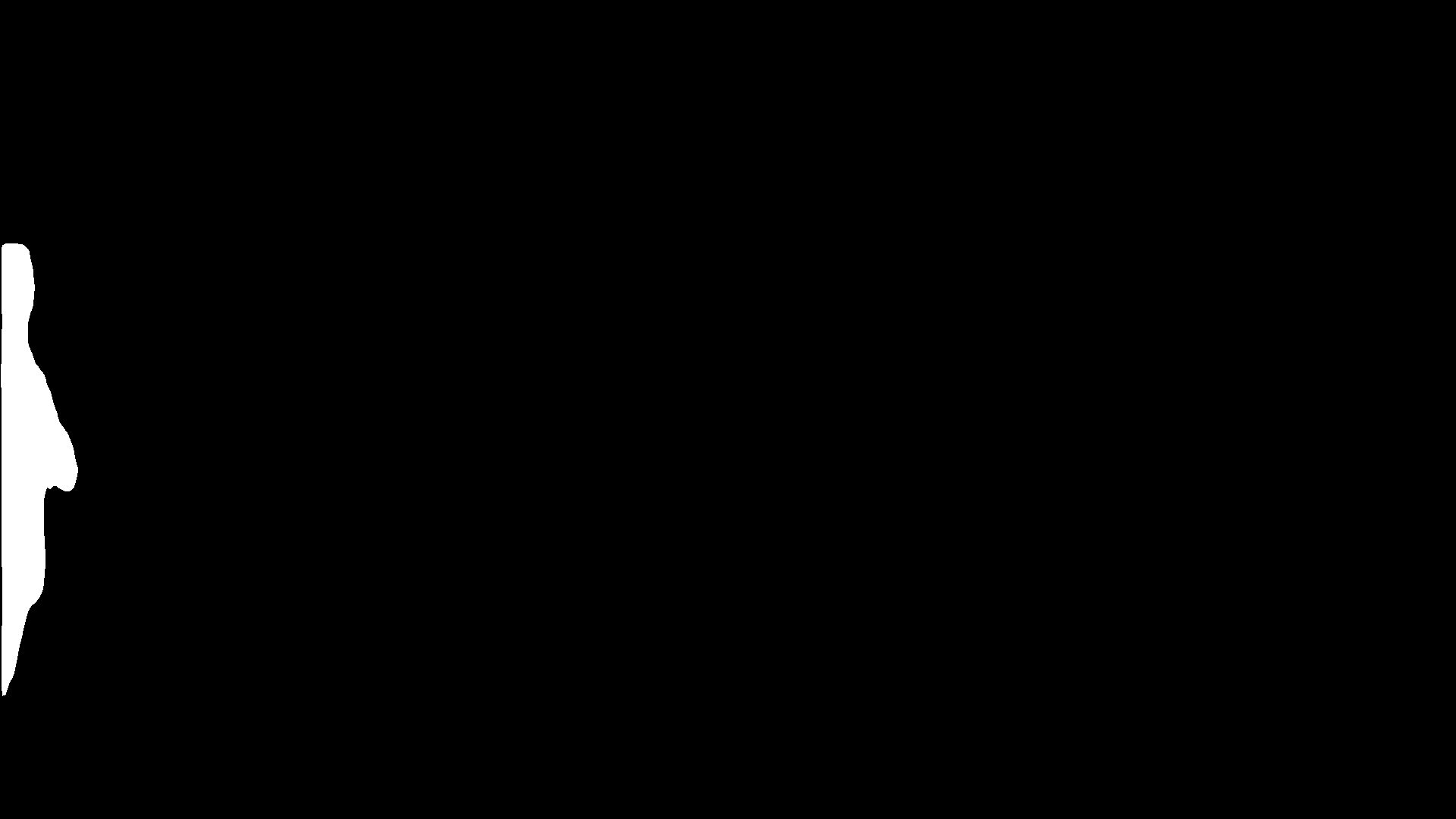} \\
\end{tabular}
\caption{The five training views and person segmentations used to produce results in Figure 4 of the main submission. }
\label{fig:seg} 
\end{figure}

\paragraph{Comparison to single human NeRF}
In Figure 4 in the main submission, we compared our work to the single human NeRF method ARAH~\cite{wang2022arah} on the CMU Panoptic dataset~\cite{Simon_2017_CVPR,Joo_2017_TPAMI}. Figure~\ref{fig:seg} shows the training images used in this experiment. It also shows the segmentation masks used for ARAH for 3 people in the scene, that we built using a state-of-the-art method. Figure~\ref{fig:seg} shows additional comparative results for reconstructed appearance and geometry. 


\begin{figure}[h]
\centering
\setlength{\tabcolsep}{0pt}
\renewcommand{\arraystretch}{0} 
\begin{tabular}{ccccc}
\includegraphics[width=1.68cm]{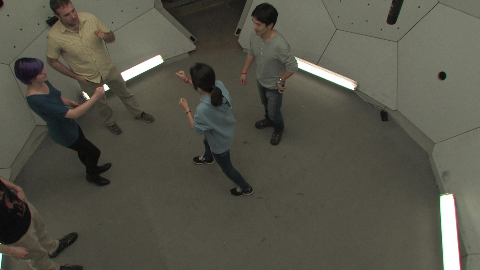} &
\includegraphics[width=1.68cm]{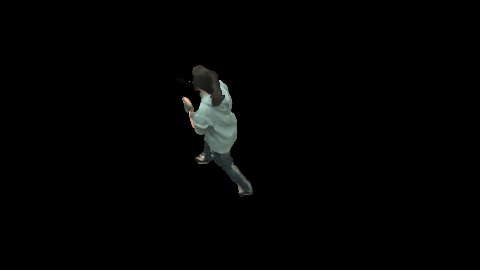} &
\includegraphics[width=1.68cm]{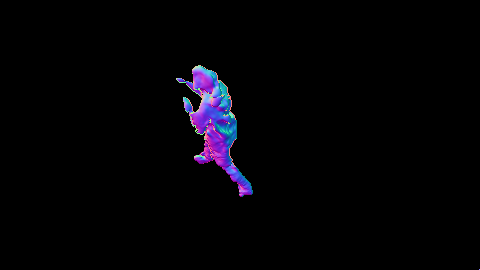}&
\includegraphics[width=1.68cm]{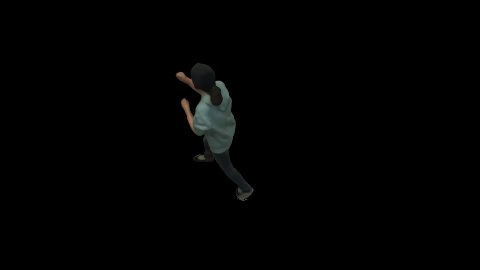} &
\includegraphics[width=1.68cm]{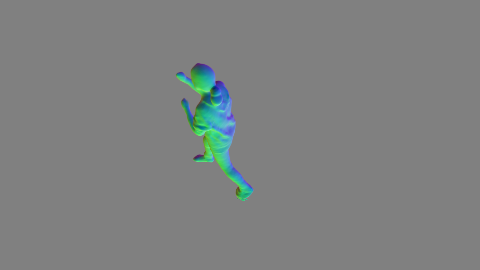} \\
\includegraphics[width=1.68cm]{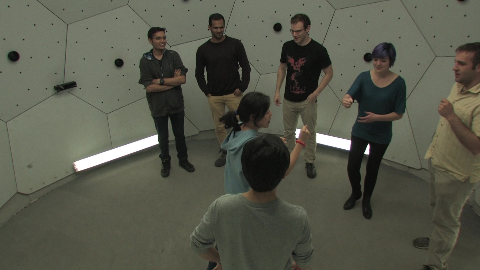} &
\includegraphics[width=1.68cm]{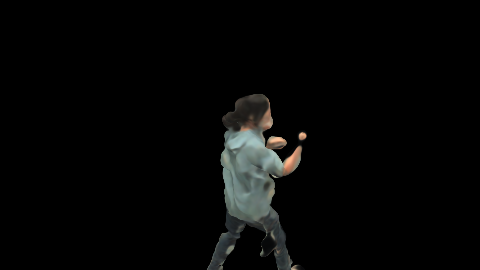} &
\includegraphics[width=1.68cm]{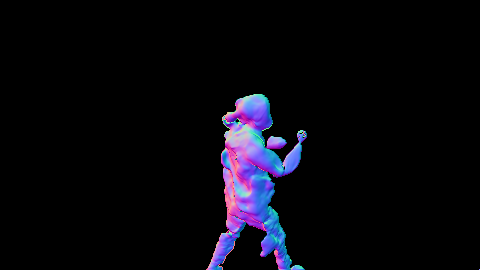}&
\includegraphics[width=1.68cm]{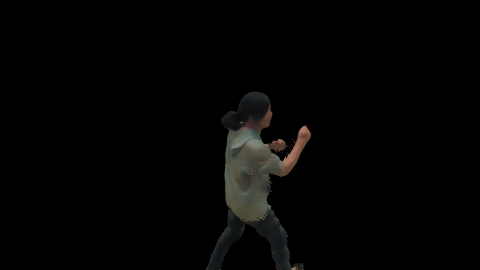} &
\includegraphics[width=1.68cm]{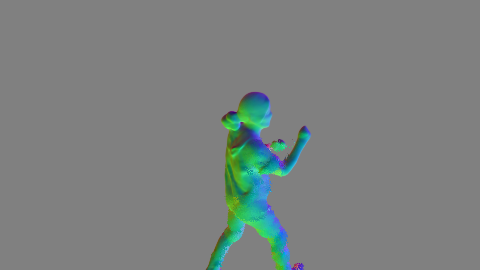} \\
\includegraphics[width=1.68cm]{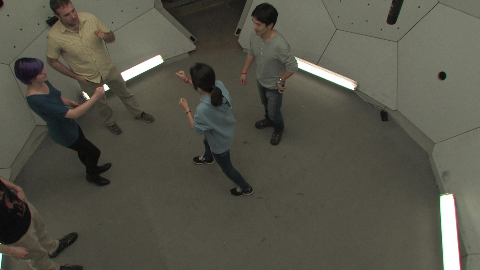} &
\includegraphics[width=1.68cm]{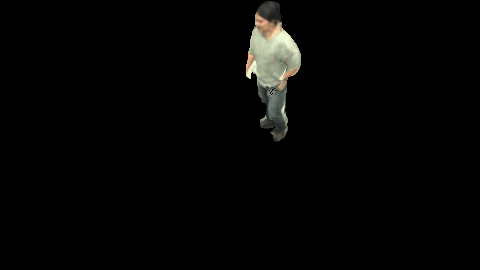} &
\includegraphics[width=1.68cm]{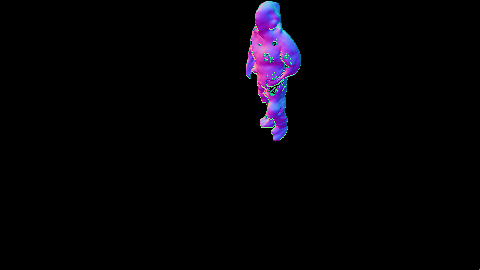}&
\includegraphics[width=1.68cm]{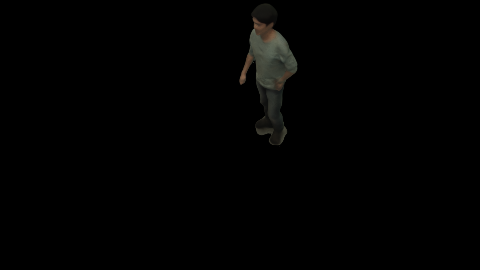} &
\includegraphics[width=1.68cm]{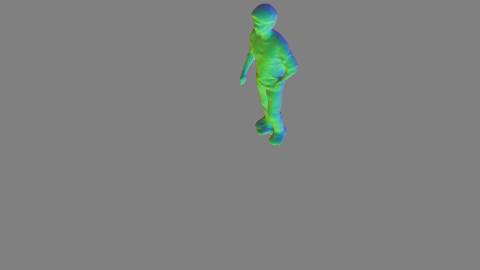} \\
\includegraphics[width=1.68cm]{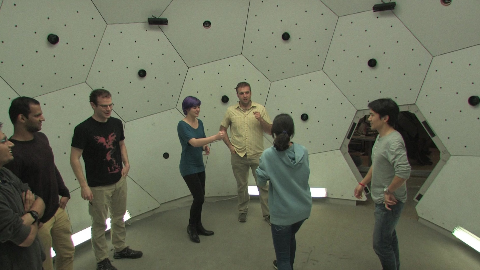} &
\includegraphics[width=1.68cm]{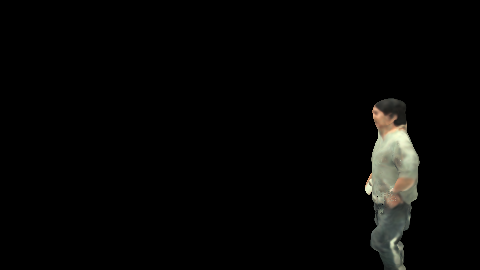} &
\includegraphics[width=1.68cm]{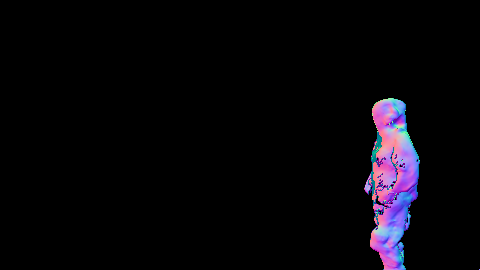}&
\includegraphics[width=1.68cm]{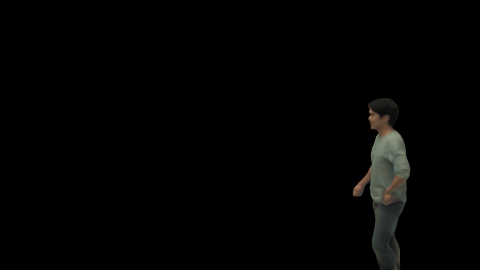} &
\includegraphics[width=1.68cm]{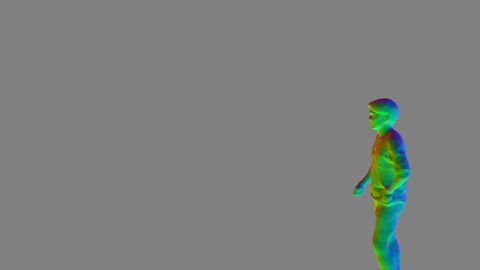} \\
\includegraphics[width=1.68cm]{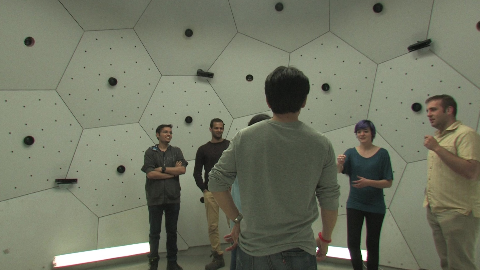} &
\includegraphics[width=1.68cm]{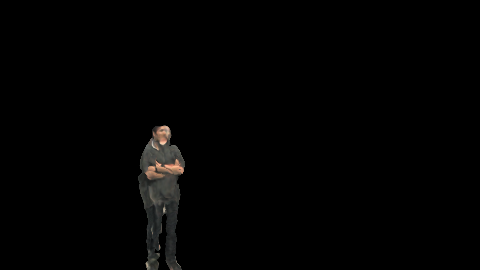} &
\includegraphics[width=1.68cm]{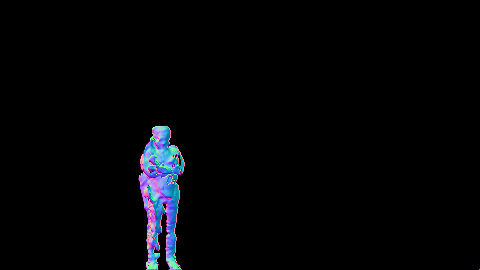}&
\includegraphics[width=1.68cm]{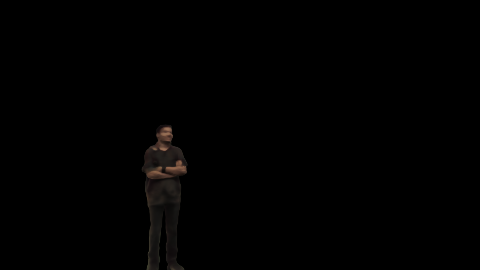} &
\includegraphics[width=1.68cm]{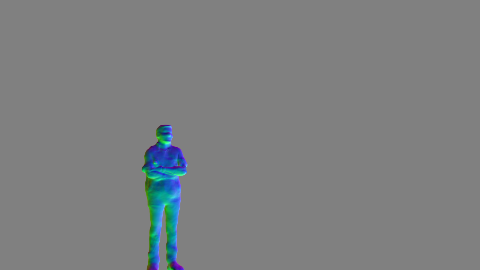} \\
\includegraphics[width=1.68cm]{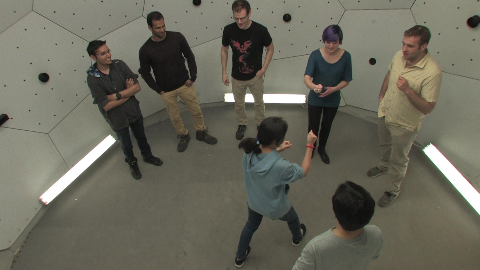} &

\includegraphics[width=1.68cm]{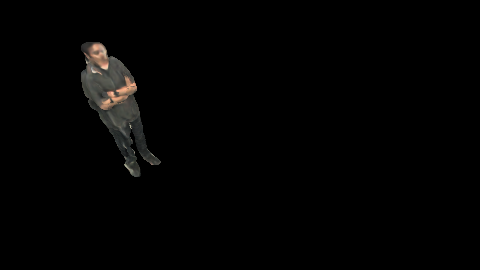} &
\includegraphics[width=1.68cm]{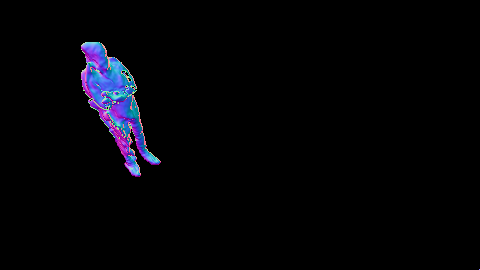}&
\includegraphics[width=1.68cm]{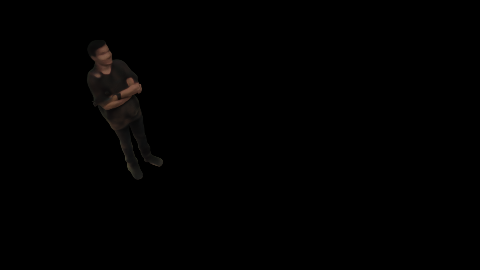} &
\includegraphics[width=1.68cm]{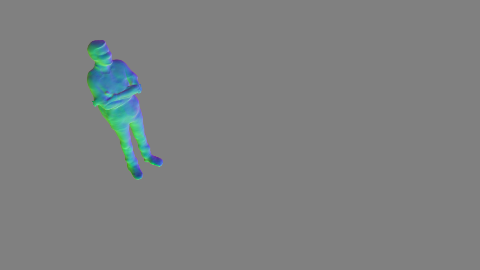} \\
GT& Arah  &Arah & Ours&  Ours\\
\end{tabular}
\caption{Comparison against single human method ARAH~\cite{wang2022arah} using 5 training views.The average PSNR in these examples is 24.11/\textbf{27.40} (ARAH/\textbf{Ours}).}
\label{fig:single} 
\vspace{-3mm}
\end{figure}

\section{Implementation Details}
\begin{figure}[h!]
\centering
\includegraphics[width=\linewidth]{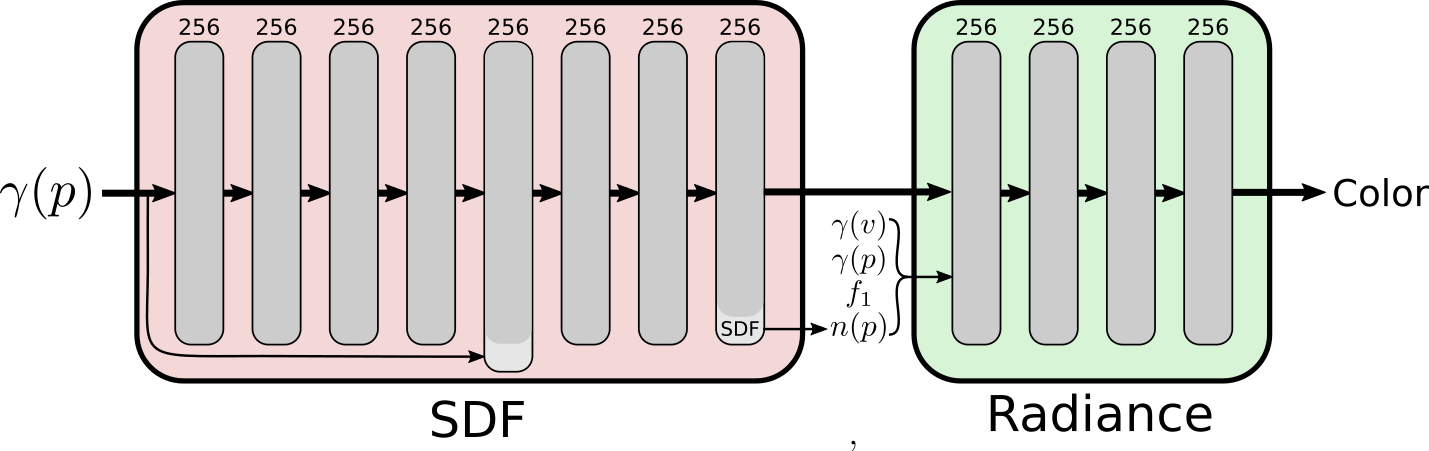}
  \caption{\textbf{Network architecture} (Section 3 of the main submission). 
$p$ is a sampled point along a ray.
$\gamma$ is the positional encoding \cite{tancik2020fourier,mildenhall2020nerf}. $n(p)$ is the gradient of predicted sdf w.r.t the input point $p$. $v$ is the direction of the ray, and $f_1$ the rasterized depth feature described in Section 3.2.}
    \label{fig:network}
\end{figure}

Fig. \ref{fig:network} shows the architecture of our network in more detail (Section 3 in the main submission).
The geometry MLP has 8 layers of width 256,  with a skip connection from the input to the 4th layer. 
The radiance MLP consists of additional 4 layers of width 256, and receives as input the positional encoding of the point $\gamma(p)$, positional encoding of the view direction $\gamma(v)$, rasterized depth feature $f_1$, 
and gradient of the SDF $n(p)$. All layers are linear with ReLU activation, except for the last layer which uses a sigmoid activation function. 
 During training we sample  512  rays  per  batch  and  follow  the  coarse  and fine sampling strategy of \cite{mildenhall2020nerf,wang2021neus}. For a fair comparison, we unified the number of sampled points on each ray for all methods, namely, each ray with $N=64$ coarsely sampled points and $N=64$ finely sampled points for the 
 foreground,
 and $N=32$ for the 
 background.
%
%

Fig. \ref{fig:pipe} illustrates the losses involved in the training of our method. Rays with available ground-truth pixels are supervised with pixel colors. Sub-pixel rays without available ground-truth are supervised using color and density pseudo-ground-truth from neighboring rays.

\begin{figure}[h!]
\centering
\includegraphics[width=\linewidth]{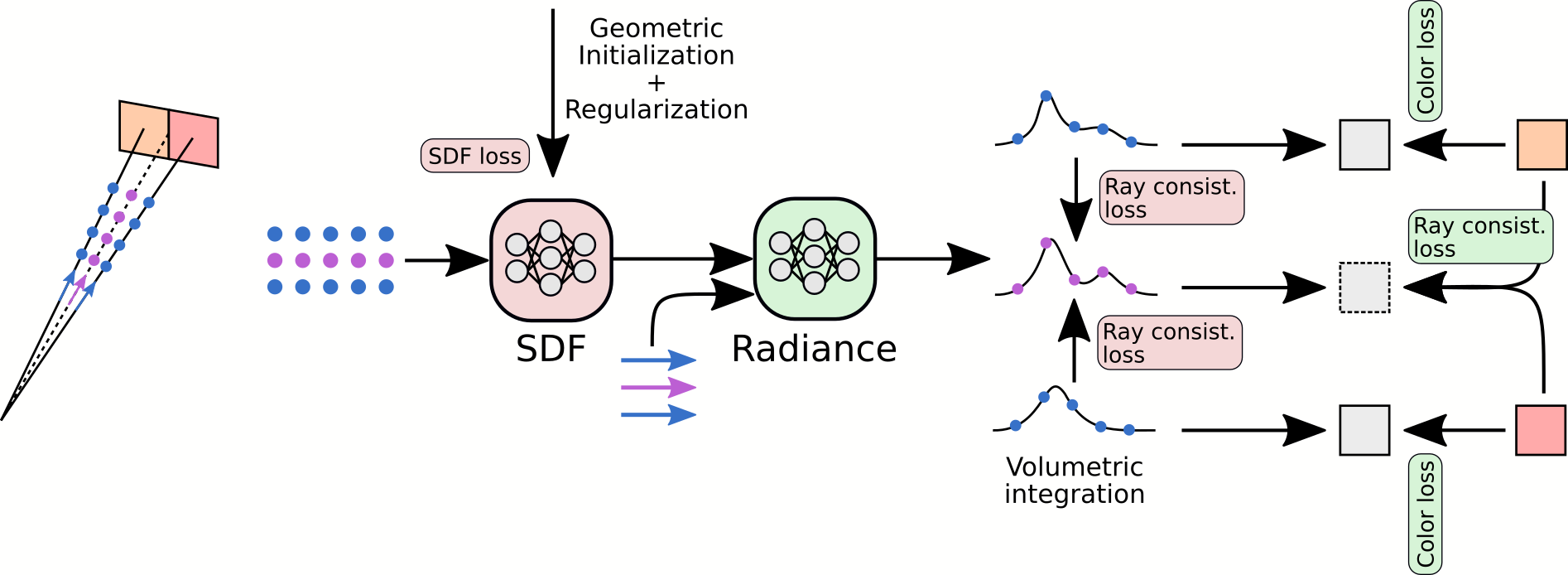}
\caption{Illustration of our losses. Rays without ground-truth are supervised using our ray consistency loss (Section 3.4). For rays corresponding to pixels in the training data, we supervise points using a combination of SDF losses and color losses (Section 3.4).}
\label{fig:pipe}
\end{figure}

\section{Datasets}
We provide here additional details on the evaluation datasets used in Section 4 from the main paper.

\paragraph{CMU Panoptic~\cite{Simon_2017_CVPR,Joo_2017_TPAMI}.}
Our experiments were performed on five different scenes from the CMU Panoptic dataset~\cite{Simon_2017_CVPR,Joo_2017_TPAMI}, where each scene includes originally 30 views located on a spherical spiral. 
The training views were randomly extracted from the HD sequences `Ultimatum' and `Haggling', and contain between 3 and 7 people. 
Specifically, we used frame 9200 from `Haggling', and frames 5500,7800,9200 and 22900 from 'Ultimatum'. 
We uniformly sampled 5, 10, 15 and 20 views as training and we used the remaining 25, 20, 15 and 10 views respectively as testing. The image resolution in training and testing is $1920 \times 1080$.

\paragraph{Synthetic Dataset from MultiHuman-Dataset \cite{tao2021function4d,zheng2021deepmulticap}}
Based on the MultiHuman-Dataset \cite{tao2021function4d,zheng2021deepmulticap}, we rendered a synthetic dataset with 29 cameras arranged in a sphere. There are three scenes in this dataset with similar backgrounds but different lighting conditions, camera locations and orientations. The scenes contain 1,5 and 10 people respectively. The image resolution is $1920 \times 1080$. We sample 5,10 and 15 views uniformly on each scene for training, and 14 views for testing.

\end{document}